\documentclass[preprint,aps,pre,letterpaper,onecolumn,superscriptaddress]{revtex4-2} %PRE numbers sections if you want that

\usepackage{comment}
\usepackage{algorithm}
\usepackage{algpseudocode}
\usepackage[margin=1in]{geometry}
\usepackage{graphicx}
\usepackage{amssymb}
\usepackage{amsmath}
\usepackage{mathrsfs}
\usepackage{placeins}
\usepackage{caption3}
\usepackage{wrapfig}
\usepackage{caption} 
\usepackage{subcaption}
\usepackage{xcolor}
\usepackage{contour}
\usepackage{color}
\usepackage{outlines}
\usepackage[bookmarksnumbered=true,breaklinks=true,colorlinks,citecolor=blue,linkcolor=blue]{hyperref}
\usepackage{soul} % added by Carlos
\usepackage[font=small,labelfont=bf,
justification=raggedright,
format=plain]{caption} % added by Carlos

\usepackage[normalem]{ulem}
\newcommand{\MDG}[1]{\textcolor{magenta}{*** #1 ***}}
\definecolor{ao(english)}{rgb}{0.0, 0.5, 0.0}

\newcommand{\IM}{\mathcal{M}}

\newcommand{\reals}{\mathbb{R}} %this requires amssymb if not using txfonts

\newcommand{\Rey}{\mathrm{Re}}

\newlength{\figwidth}
\newlength{\SCwidth}
\def\XXint#1#2#3{{\setbox0=\hbox{$#1{#2#3}{\int}$}
		\vcenter{\hbox{$#2#3$}}\kern-.5\wd0}}

\begin{document}

%\preprint{}

\title{Data-driven low-dimensional dynamic model of Kolmogorov flow}
%\title{Minimal Dimensional Evolution of the Kuramoto-Sivashinsky Equation with Neural ODEs} %MDG


\author{Carlos E. P\'{e}rez De Jes\'{u}s}
\affiliation{Department of Chemical and Biological Engineering, University of Wisconsin-Madison, Madison WI 53706, USA}

\author{Michael D. Graham}
\email{mdgraham@wisc.edu}
\affiliation{Department of Chemical and Biological Engineering, University of Wisconsin-Madison, Madison WI 53706, USA}


\date{\today}

\begin{abstract}
	
%The high dimensionality of the Navier-Stokes equations (NSE) makes analytical or model based flow control intractable which in turn motivates the need to find reduced order models (ROMs)
Reduced order models (ROMs) that capture flow dynamics are of interest for decreasing computational costs for simulation as well as for model-based control approaches.  This work presents a data-driven framework for minimal-dimensional models that effectively capture the dynamics and properties of the flow. We apply this to Kolmogorov flow in a regime consisting of chaotic and intermittent behavior, which is common in many flows processes and is challenging to model. The trajectory of the flow travels near relative periodic orbits (RPOs), interspersed with sporadic bursting events corresponding to excursions between the regions containing the RPOs. The first step in development of the models is use of an undercomplete autoencoder to map from the full state data down to a latent space of dramatically lower dimension. Then models of the discrete-time evolution of the dynamics in the latent space are developed. By analyzing the model performance as a function of latent space dimension we can estimate the minimum number of dimensions required to capture the  system dynamics.  To further reduce the dimension of the dynamical model, we factor out a phase variable in the direction of translational invariance for the flow, leading to separate evolution equations for the pattern and phase dynamics. At a model dimension of five for the pattern dynamics, as opposed to the full state dimension of 1024 (i.e.~a $32\times 32$ grid), accurate predictions are found  for individual trajectories out to about two Lyapunov times, as well as for long-time statistics. Further small improvements in the results occur as dimension is increased to nine, beyond which the statistics of the model and true system are in very good agreement. The nearly heteroclinic connections between the different RPOs, including the quiescent and bursting time scales, are well captured. We also capture key features of the phase dynamics. Finally, we use the low-dimensional representation to predict future bursting events, finding good success.

\newpage

%To estimate the dimension we trained an undercomplete autoencoder on weakly chaotic vorticity data from Kolmogorov flow simulations tracking the reconstruction error as a function of dimension together with a discrete time map that evolves the reduced order model with a nonlinear dense neural network. 
%This is mainly due to their high dimensionality which affects computational cost and controllability.  Reduced order models are needed to correctly capture key dynamics of flows. , however finding the minimal dimension needed to correctly capture the key dynamics is not a trivial task. 

\end{abstract}

%\keywords{}

\maketitle

%2

%\section*{results -- to dos}
%\begin{itemize}
%	\item phase evolution
%	\item bursting and hibernation durations, fractions of time spent
%	\item transition predictions starting from hibernation vs burst
%	\item can we predict bursting from state in latent space?
%	\item time-series and state space projections in latent variables (connected to previous point)
%	\item differences between pdfs -- $L_2$ difference isn't a good measure. Talk to Alec about different things he's been looking at. 
%\end{itemize}

\begin{comment}
\section*{Carlos notes (to address)}

\begin{itemize}
	\item talk about reservoir papers in the introduction
	\item  in the trajectory difference part want to include 50/50 autoencoders, have to also show Power diss and Re Im plot in some section, also explain why I'm using these, what to they show us, should consider changing terms from hibernating to quiescent. 
	\item  in the hibernating and bursting time section I think we might/should also be able to take the initial peak out, by setting a threshold of the data we want to include in this PDF
	\item maybe talk about neural ODEs in the conclusion
\end{itemize}
\end{comment}
\section{Introduction} \label{sec:Intro}

%Singular value decomposition (SVD) 

%By doing singular value decomposition (SVD) on a snapshot matrix $X$ that contains the velocity variables of interest, and projecting onto the leading basis vectors $d_h$ of $U$ which come from $X=U \Sigma V^T$

%Hence a low-dimensional model can be constructed by truncating singular value decomposition (SVD) on a snapshot matrix $X$, and projecting onto the leading basis vectors $d_h$ of $U$ which come from $X=U \Sigma V^T$. 

%of data projected onto a lower dimensional basis set with respect to the true data. 

%The basis vectors come from doing singular value decomposition (SVD) on a snapshot matrix $X$, and projecting onto the leading basis vectors $d_h$ of $U$ which come from $X=U \Sigma V^T$. $U$ is organized in such a way that most of the energy is captured by the first basis vector.

%This basis can come from doing singular value decomposition (SVD) on a snapshot matrix $X$

%, and projecting onto the leading vectors of $U$ which come from $X=U \Sigma V^T$. Here $X$ is a collection of the snapshots of dimension $N$. 

%\hl{"Lit rev, motivation, citations. Talk about trajectory in Fourier space, stability, symmetry groups, weakly chaotic behaviour" at the end of intro intoduce the next sections, reduced models for control} 
 
 % focus on high dim, higly nonlinear, time scales

Development of reduced order dynamical models for complex flows is an issue of long-standing interest, with applications in improved understanding, as well as control, of flow phenomena. The classical approach for dimension reduction of these systems consists of extracting dominant modes from data via principal component analysis (PCA), also known as proper orthogonal decomposition (POD) and Karhunen-Lo\'{e}ve decomposition \cite{holmes2012turbulence}. PCA determines a set of basis vectors ordered by their contribution to the total variance (fluctuating kinetic energy) of the flow. Given $N_s$ data vectors (``snapshots") $x_i\in \reals^N$, one can obtain these basis vectors by performing singular value decomposition (SVD) on the data matrix $X=[ x_1,x_2,\cdots ] \in \mathbb{R}^{N \times N_s}$ such that $X=U \Sigma V^T$. Projecting the data onto the first $d_h$ basis vectors (columns of $U$) then gives a low-dimensional representation -- a projection onto a linear subspace of the full state space. To find a reduced order model (ROM), a Galerkin approximation of the Navier-Stokes Equations (NSE) using this basis can be implemented; these have shown some success in capturing the dynamics of coherent structures \cite{noack1994low, aubry1988dynamics}. Previous research has also used POD as well as a filtered version thereof  \cite{sieber2016spectral}, which are linear reduction techniques, to reduce dimensions and learn a time evolution map from data with the use of neural networks (NNs) \cite{lui2019construction}.

\begin{figure}%[h]
	%\vspace{-8mm}
	\centering
	\includegraphics[width=0.5\linewidth]{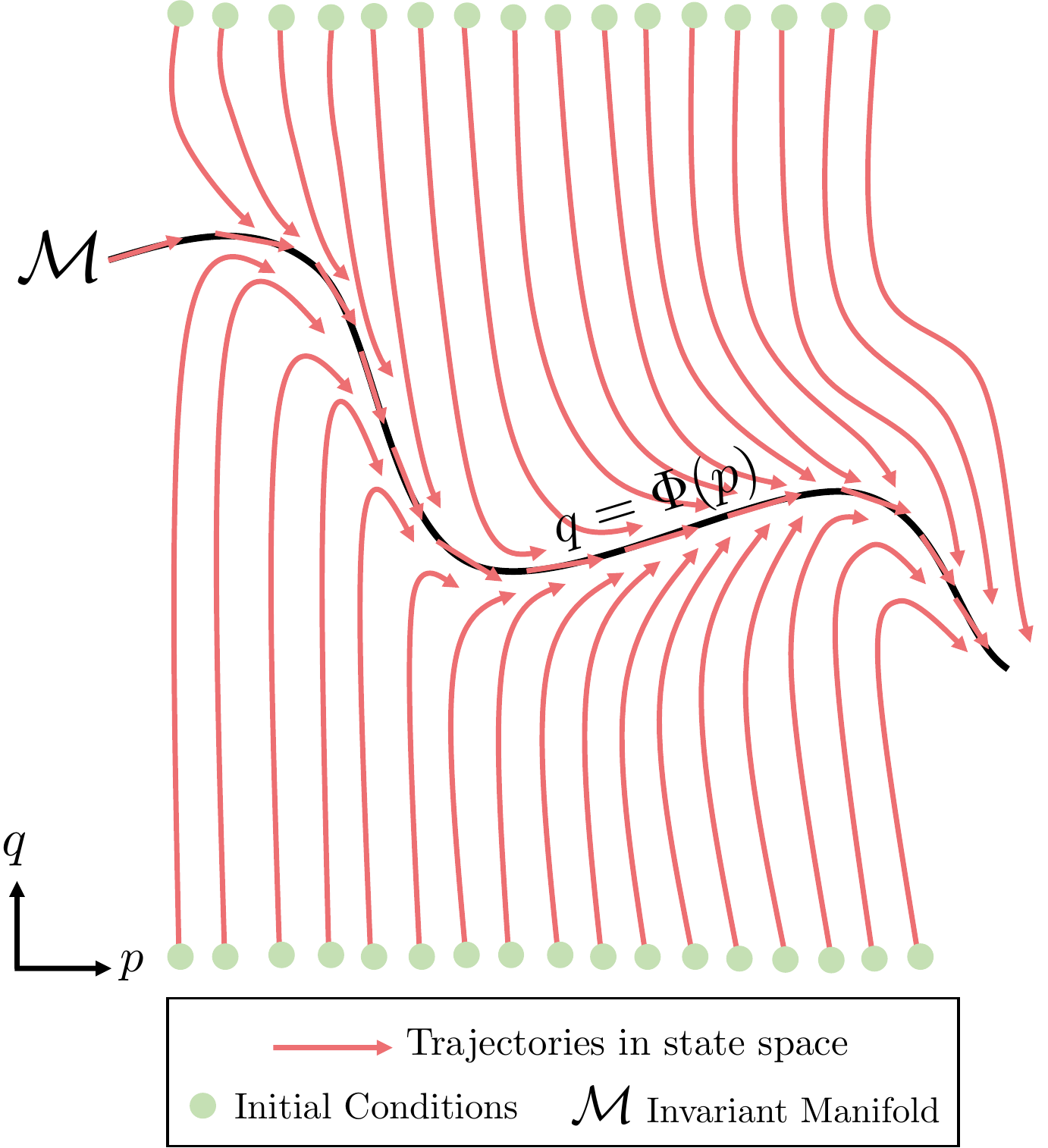}
	\caption{Schematic of state space with initial conditions collapsing onto an invariant manifold where the long time dynamics occur.}
	\label{fig:IM_picture}
\end{figure}

Although PCA provides the best linear representation of a data set in $d_h$ dimensions, in general the long-time dynamics of a general nonlinear dynamical systems are not expected to lie on a linear subspace of the state space. For a primer and more details on data-driven dimension reduction methods for dynamical systems refer to Linot \& Graham \cite{linot2022data}. For dissipative systems, such as the NSE, it is expected that the long-time dynamics will lie on an invariant manifold $\IM$, which can be represented \emph{locally} with Cartesian coordinates, but may have a complex global topology \cite{Hopf1948a}. In fluid mechanics, this manifold is often called an \emph{inertial manifold} \cite{foias1988modelling, temam1989inertial, zelik2022attractors}. 
Figure \ref{fig:IM_picture} schematically illustrates a simple example of this idea. Consider a dynamical system $\dot{x}=F(x)$ for state variable $x\in \reals^N$. As time proceeds, general initial conditions in this space evolve toward an invariant manifold $\IM$ of dimension $d_\IM$, which in this example can be described by the equation $q=\Phi(p)$ where $x=p+q$, $p\in \reals^{d_\IM},q\in \reals^{N-d_\IM}$. Furthermore, if we write the dynamics in terms of $p$ and $q$ as $\dot{p}=f(p,q), \dot{q}=g(p,q)$, then trajectories on $\IM$ evolve according to $\dot{p}=f(p,\Phi(p))$: i.e.~the long time dynamics are given by a set of ordinary differential equations in $d_\IM$ dimensions,  rather than the $N$ dimensions of the original system. More generally, since $\IM$ is invariant under the dynamics, the vector field on $\IM$ is always tangent to $\IM$, and the dynamics on $\IM$ are determined by this vector field. In the present work we do not require that the manifold be represented in this simple form, but rather a more general form $G(x)=0$. In this example, $G(x)=q-\Phi(p)$.
% \CEPrevise{As described by Temam \cite{temam1989inertial}, given the dissipative part of the NSE by $A$, and assuming that $A^{-1}$ is positive self-adjoint and compact we decompose the state $u$ into $p$ and $q$ where $p=P_mu$ and $q=(I-P_m)u=Q_mu$. Here $P_m$ and $Q_m$ are projections into the leading $m$ and trailing eigenvectors respectively that come from $A w_{j}=\lambda_{j} w_{j}$. By projecting the NSE onto $P_m$ and $Q_m$ we get the coupled system: $dp/dt=f_p(p,q)$, $dq/dt=f_q(p,q)$. For an attracting invariant manifold $q=\Phi (p)$. Hence we get, $dp/dt=f_p(p,\Phi (p))$. We include a visual depiction of this in Figure \ref{IM_picture}.} Hence to find a high-fidelity low-dimensional model one desires to find $\mathcal{M}$ and the dynamical system on it. The present work will consider discrete-time models, though differential equation models could be found as well \cite{linot2022data}. 

In general one can think of breaking up $\mathcal{M}$ into overlapping regions that cover the domain, to find a local representation. These are called charts and are equipped with a coordinate domain and a coordinate map \cite{lee2013smooth}. The strong Whitney's embedding theorem states that any smooth manifold of dimension $d_\mathcal{M}$ can be embedded into a Euclidean space of so-called \emph{embedding} dimension $2d_\mathcal{M}$ \cite{lee2013smooth, whitney1944self}. This means that in the worst case we can expect in principle to be able to find a $2d_\mathcal{M}$-dimensional Euclidean space in which the dynamics lie. To find a $d_\mathcal{M}$-dimensional Euclidean space one would in general need to develop overlapping local representations and evolution equations -- this avenue is not pursued in the present work but has been done elsewhere  \cite{floryan2021charts}.  In this work we aim to find a high-fidelity low-dimensional dynamical model using data from simulations of two-dimensional Kolmogorov flow. In this work, the governing Navier-Stokes Equations will only be used to generate the data -- the models will only use this data, not the equations that generated it.  Neural networks (NNs) will be used to map between the full state space and the manifold, as well as for the dynamical system model on the manifold. 
%We will use the performance of models with various numbers of dimensions to estimate the embedding dimension, 
%In this work we aim to find a high-fidelity low-dimensional data-driven model model of to estimate the embedding dimension and predict dynamics on it. Our focus is based on data-driven methods using NNs where we learn a nonlinear low-dimensional projection and a nonlinear time map.

A number of previous studies have focused on finding {data-driven} models for fluid flow problems with the use of NNs. Srinivasan \textit{et al.} \cite{srinivasan2019predictions} developed NN models to attempt to predict the time evolution of the Moehlis-Faisst-Eckhardt (MFE) model \cite{moehlis2004low}, which is a nine-dimensional  model for turbulent shear flows. They used two approaches to finding discrete-time dynamical systems. The first is to simply use a neural network as a discrete-time map, yielding a Markovian representation of the time evolution. The second is to use a long short-term memory (LSTM) network, which yields a non-Markovian evolution equation. Despite the fact that the dynamics are in fact Markovian, the LSTM approach worked better, yielding reasonable agreement with the Reynolds stress profiles. Page \textit{et al.} used deep convolutional autoencoders (CAEs) to learn low-dimensional representations for two-dimensional (in physical space) Kolmogorov flow, showing that these networks retain a wide spectrum of lengthscales and capture meaningful patterns related to the embedded invariant solutions \cite{page2021revealing}. They considered the case where bursting dynamics is obtained at a Reynolds number of $\text{Re}=40$ and $n=4$ wavelengths in the periodic domain. Nakamura \textit{et al.} used CAEs for dimension reduction combined with LSTMs and applied it to minimal turbulent channel flow for $\text{Re}_\tau=110$ where they showed to capture velocity and Reynolds stress statistics \cite{nakamura2021convolutional}. They studied various degrees of dimension reduction, showing good performance in terms of capturing the statistics; however for drastic dimension reduction they showed how only large vortical structures were captured. Hence, the selection of the minimal dimension to accurately represent the state becomes a challenging task. Reservoir networks  have also shown great potential in learning nonlinear models for time evolution. For example, Doan \textit{et al.} trained what they call an Auto-Encoded Reservoir-Computing (AE-RC) framework where the latent space is fed into an Echo State Network (ESN) to model evolution in discrete time \cite{doan2021auto}. By considering the two-dimensional Kolmogorov flow for $\text{Re}=30$ and $n=4$ good performance was obtained when comparing the kinetic energy and dissipation evolution in time. They also showed how the model captures the velocity statistics. However, the nature of the reservoir in the ESN stores past history, making the model non-Markovian.

 Although previous research has found data-driven ROMs for fluid flow problems, the focus on these has not been to find the minimal dimension required to capture the data manifold and dynamics. Linot \& Graham have addressed this issue for the Kuramoto-Sivashinsky equation (KSE) \cite{linot2020deep, linot2022data}. They showed that the mean squared error (MSE) of the reconstruction of the snapshots using an AE for the domain size of $L=22$ exhibited an orders-of-magnitude drop when the dimension of the inertial manifold is reached. Furthermore, modeling the dynamics with a dense NN at this dimension either with a discrete time map \cite{linot2020deep} or a system of ordinary differential equations (ODE) \cite{linot2022data} yields excellent trajectory predictions and long-time statistics. Increasing domain size to $L=44$ and $L=66$, which makes the system more chaotic, affects the drops of MSE significantly. However a drop is still seen, and when obtaining the dynamics and calculating long time statistics, good agreement with the true data is obtained. This work, denoted ``Data-driven manifold dynamics" (DManD) has been extended to incorporate reinforcement learning control for reduction of dissipation in the KSE, yielding a very effective control policy \cite{Zeng.2022}.

We aim to extend this approach to the NSE, specifically to the two-dimensional Kolmogorov flow, where an external forcing drives the dynamics. As $\text{Re}$ increases, the trivial state becomes unstable, giving rise to periodic orbits (POs), relative periodic orbits (RPOs) and eventually chaos. Relative periodic orbits correspond to periodic orbits in in a moving reference frame, such that in a fixed frame, the pattern at time $t+T$ is a phase-shifted replica of the pattern at time $t$. The nature of the weakly turbulent dynamics at a Reynolds number of $\operatorname{Re} = 14.4$, and connections with  RPO solutions are the focus of this study. Due to the symmetries of the system the chaotic dynamics travels between unstable RPOs \cite{crowley2022turbulence} through bursting events \cite{armbruster1996symmetries} that shadow heteroclinic orbits connecting the RPOs. A past study \cite{armbruster1992phase} shows that low-dimensional representations can be found with PCA for  two-dimensional Kolmogorov flow where in the case of weakly turbulent data, the first two PCA basis in the streamfunction formulation  capture most of the energetic content when filtering out the bursting events before the analysis, and including a third basis function captures the bursting information. This point hints at the low-dimensional nature of this system, where a low number of PCA basis functions can energetically represent the data. However, even though the energy can be contained in a low number of basis functions, this does not imply that these will properly capture the dynamics  \cite{rowley2017model}. In \cite{armbruster1992phase}, development of a model of time-evolution was not considered.

Returning to the aims of the present work, our focus is twofold. We aim to learn a minimal-dimensional high fidelity data-driven model for the long-time dynamics of two-dimensional Kolmogorov flow with the use of an autoencoder (AE), and a discrete-time map, in the form of a dense NN, of the dynamics on the invariant manifold. In this map, the future time prediction only depends on the present state (on the manifold), in keeping with the Markovian nature of the dynamics on the manifold. This approach contrasts with models that use an RNN such as an LSTM, which carry a memory of past states so are not Markovian. It is important to note, however, that the dimension of the invariant manifold is not known a priori, and if we map the data onto a manifold of too low a dimension, then the dynamics on that manifold will not be Markovian. Accordingly, in this work we will carefully assess the performance of our Markovian models as a function of manifold dimension. For our results, the model predictions will be evaluated as a function of dimension, considering short-time trajectories, long-time statistics, quiescent and bursting time distributions, and predictions of bursting events. This paper is structured as follows: in Section \ref{sec:Framework} we present the governing equations together with the symmetries of the system. We also present the dynamics at the two values of $\operatorname{Re}$ considered and the connections of the RPOs with the chaotic regime. In Section \ref{sec:AEs} we show the methodology for data-driven dimension reduction and dynamic modeling, which includes the AE architecture and the time map NN. Section \ref{sec:Results} shows the results, and concluding remarks are given in Section \ref{sec:Conclusion}.

\newpage

\section{Kolmogorov flow formulation and dynamics} \label{sec:Framework}

The two-dimensional Navier-Stokes equations (NSE) with Kolmogorov forcing are
\begin{gather}
\frac{\partial \boldsymbol{u}}{\partial t}+\boldsymbol{u} \cdot \nabla \boldsymbol{u}+\nabla p=\frac{1}{\operatorname{Re}} \nabla^{2} \boldsymbol{u}+\sin (n y) \hat{\boldsymbol{x}}  \\
\nabla \cdot \boldsymbol{u}=0
\end{gather}
where $\boldsymbol{u}=[u,v]$ is the velocity vector, $p$ is the pressure, $n$ is the wavenumber of the forcing, and $\hat{\boldsymbol{x}}$ is the unit vector in the $x$ direction. Here $\operatorname{Re}=\frac{\sqrt{\chi}}{v}\left(\frac{L_{y}}{2 \pi}\right)^{3 / 2}$ where $\chi$ is the dimensional forcing amplitude, $\nu$ is the kinematic viscosity, and $L_y$ is the size of the domain in the $y$ direction. We consider the periodic domain $[0,2 \pi / \alpha] \times[0,2 \pi]$ with $\alpha=1$. Vorticity is defined as $\omega = \nabla \times \boldsymbol{u}$. The equations are invariant under several symmetry operations \cite{chandler2013invariant}, namely a shift (in $y$)-reflect (in $x$), a rotation through $\pi$, and a continuous translation in $x$:
\begin{gather}
\mathscr{S}:[u, v, \omega](x, y) \rightarrow[-u, v,-\omega]\left(-x, y+\frac{\pi}{n}\right), \\
\mathscr{R}:[u, v, \omega](x, y) \rightarrow[-u,-v, \omega](-x,-y), \\
\mathscr{T}_{l}:[u, v, \omega](x, y) \rightarrow[u, v, \omega](x+l, y) \quad \text { for } 0 \leqslant l<\frac{2 \pi}{\alpha}.
\end{gather}
%\begin{equation}
%\mathscr{S}:[u, v, \omega](x, y) \rightarrow[-u, v,-\omega]\left(-x, y+\frac{\pi}{n}\right), 
%\end{equation}
%\begin{equation}
%\mathscr{R}:[u, v, \omega](x, y) \rightarrow[-u,-v, \omega](-x,-y),
%\end{equation}
%\begin{equation}
%\mathscr{T}_{l}:[u, v, \omega](x, y) \rightarrow[u, v, \omega](x+l, y) \quad \text { for } 0 \leqslant l<\frac{2 \pi}{\alpha}.
%\end{equation}
%\hl{should have defined Lx and Ly previously in the text also introduce symmetries}. 
%\begin{equation}
%\frac{\partial \boldsymbol{u}}{\partial t}+\boldsymbol{u} \cdot \nabla \boldsymbol{u}+\frac{1}{\rho} \nabla p=v \nabla^{2} \boldsymbol{u}+\chi \sin \left(2 \pi n y / L_{y}\right) \hat{\boldsymbol{x}} 
%\end{equation}
%\begin{equation}
%\nabla \cdot u=0
%\end{equation}
%where 
%we then introduce the length scale
\begin{figure} % Here I had [H] before, looks better without for now
	%\vspace{-8mm}
	\centering
	\begin{subfigure}{.5\textwidth}
		\centering
		\includegraphics[width=1\linewidth]{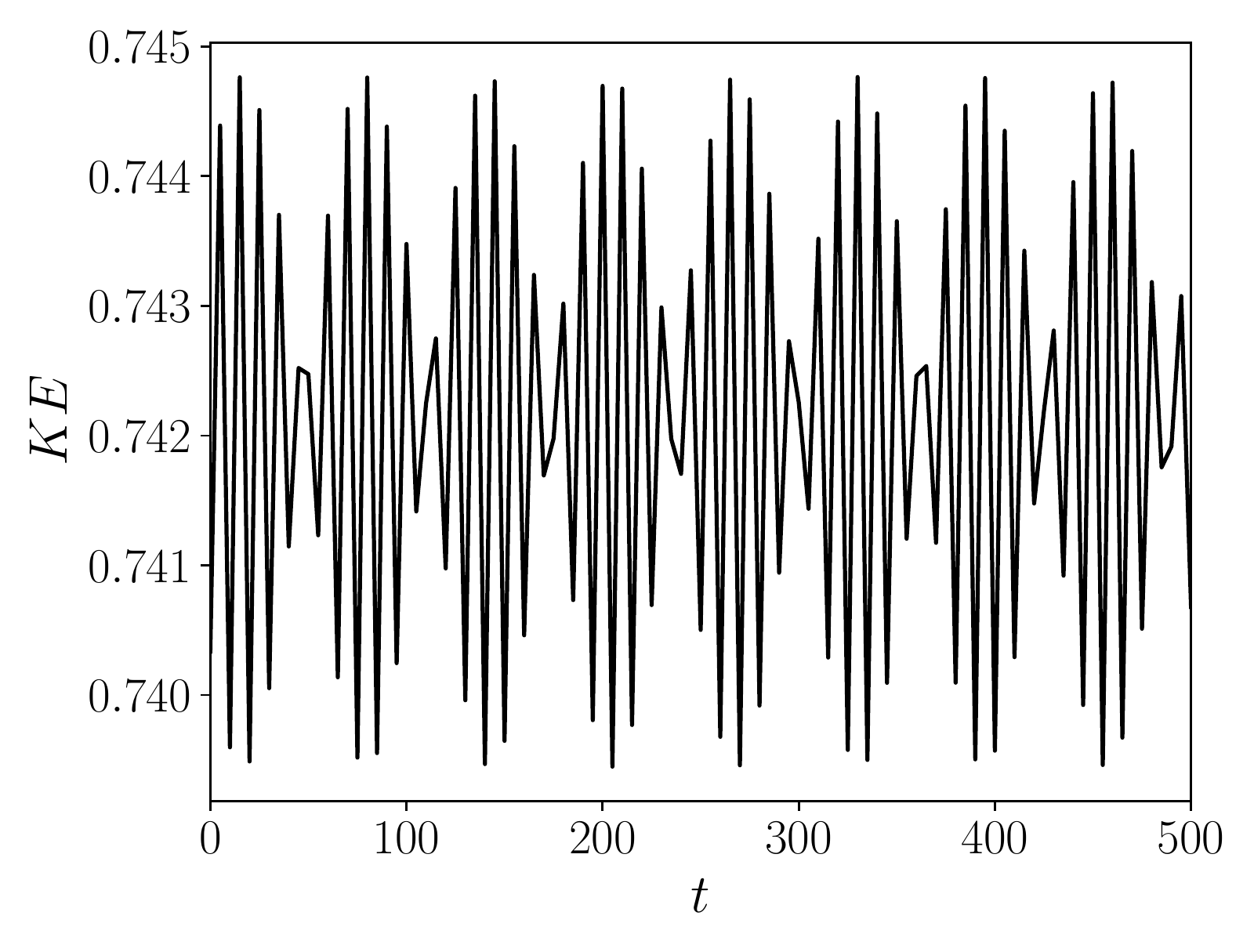}
		\caption{}
		\label{fig:sub1}
	\end{subfigure}%
	\begin{subfigure}{.5\textwidth}
		\centering
		\includegraphics[width=1\linewidth]{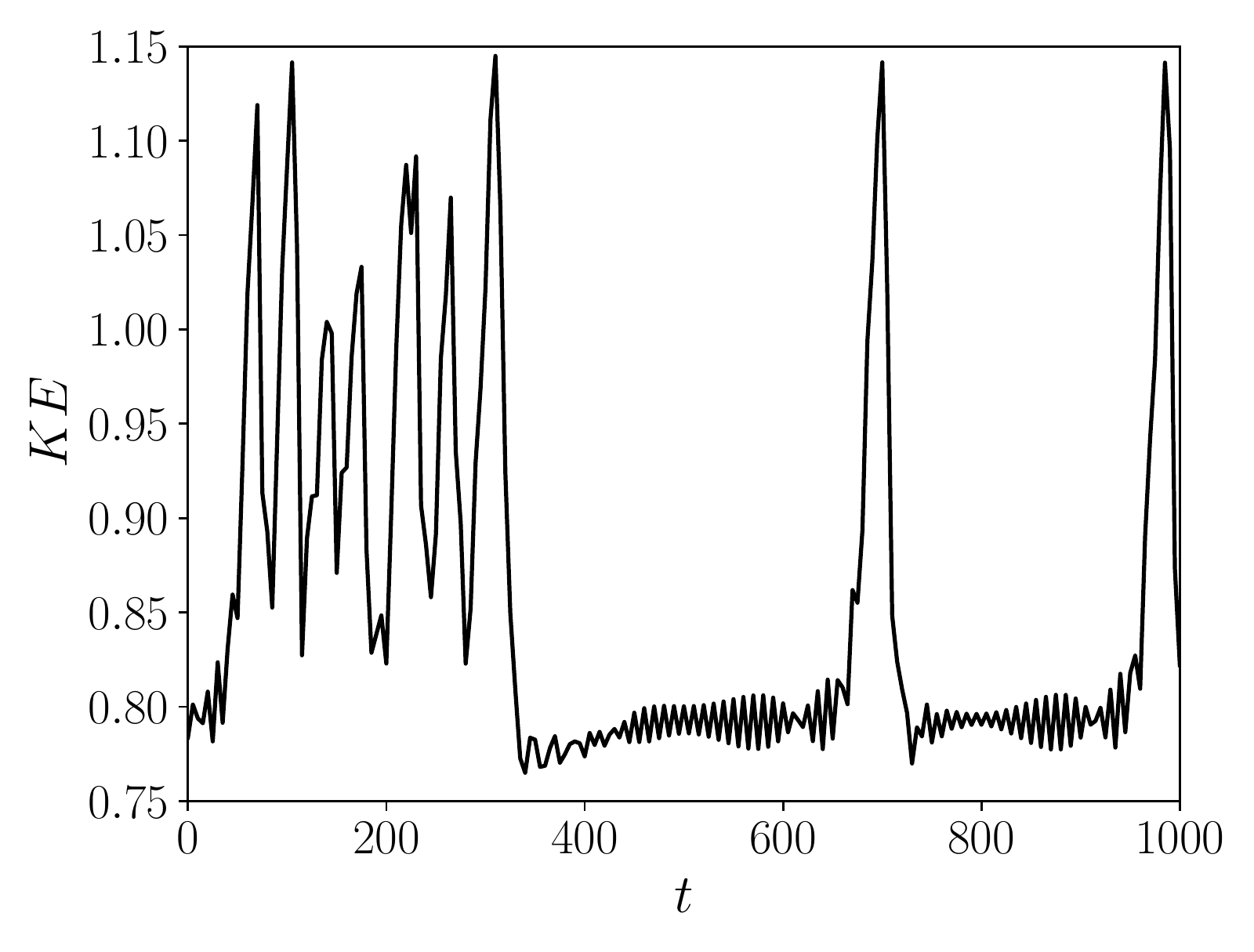}
		\caption{}
		\label{fig:sub2}
	\end{subfigure}
	\begin{subfigure}{.5\textwidth}
		\centering
		\includegraphics[width=1\linewidth]{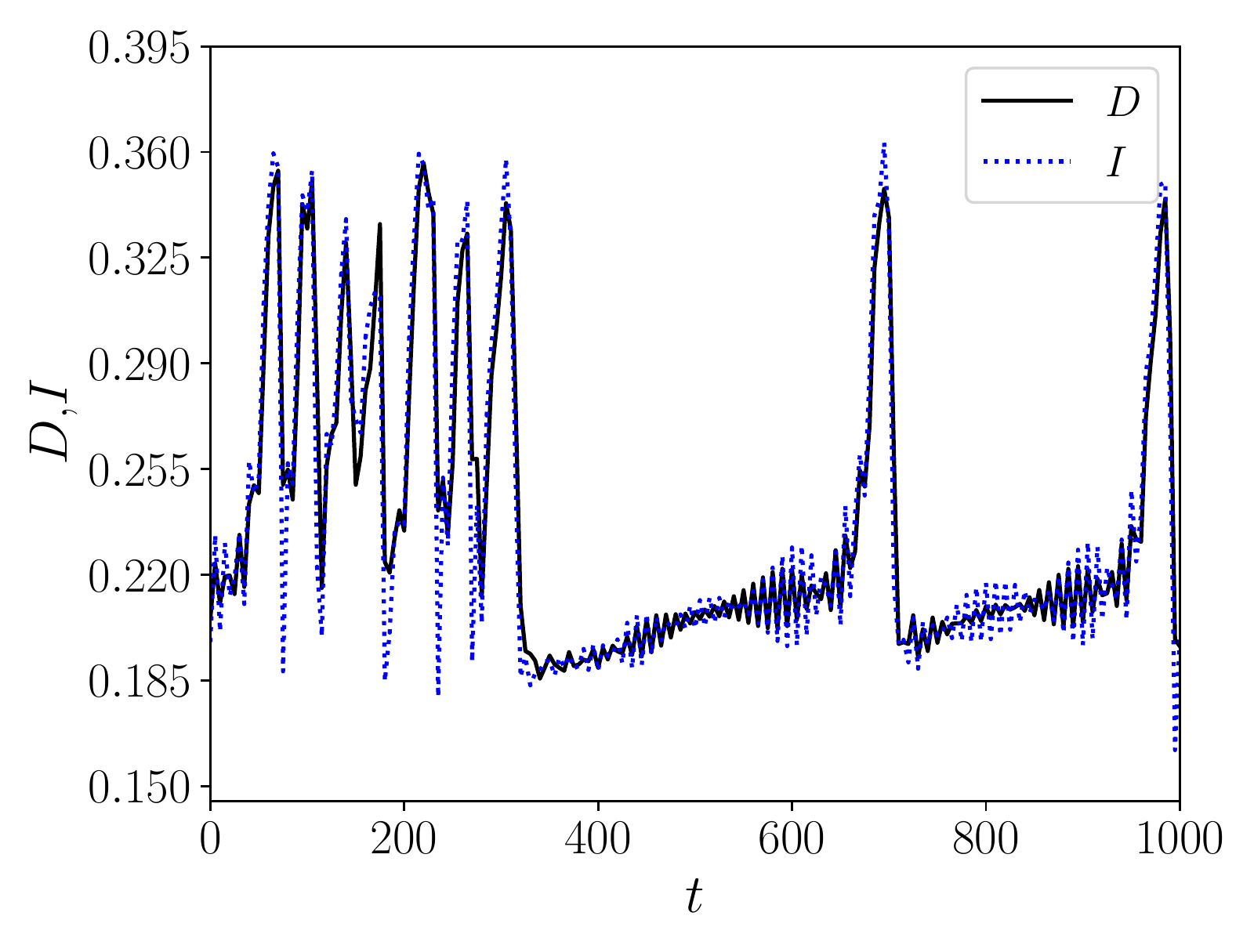} %this figure is in testHL3trial3/load.py
		\caption{}
		\label{fig:subID}
	\end{subfigure}
	\caption{(a) Time evolution of $KE$ at $\operatorname{Re}=13.5$. (b) Time evolution of $KE$ at $\operatorname{Re}=14.4$. (c) Time evolution of $D$ and $I$ at $\operatorname{Re}=14.4$.}
	\label{fig:test}
\end{figure}
The total kinetic energy for this system ($KE$), dissipation rate ($D$) and power input ($I$) are 
%\begin{equation}
%KE =\frac{1}{2}\left\langle\boldsymbol{u}^{2}\right\rangle_{V}
%\end{equation}
\begin{equation}
KE =\frac{1}{2}\left\langle\boldsymbol{u}^{2}\right\rangle_{V}, D=\frac{1}{\operatorname{Re}}\left\langle|\nabla \boldsymbol{u}|^{2}\right\rangle_{V}, \quad I=\langle u \sin (n y)\rangle_{V}
\end{equation}
where subscript $V$ corresponds to the average taken over the domain. For the case of $n=1$ the trivial solution is linearly stable at all $\text{Re}$ \cite{iudovich1965example}. It is not until $n=2$ that the laminar state becomes unstable, with a critical value of $\operatorname{Re}_c=n^{3/2}2^{1/4}$\cite{meshalkin1961investigation, green1974two, thess1992instabilities}.

%It is not until $n=2$ that chaotic dynamics can be obtained with increasing $\operatorname{Re}$.
%Specifically this happens at a critical value of $\operatorname{Re}_c=n^{3/2}2^{1/4}$ \cite{meshalkin1961investigation, green1974two, thess1992instabilities}.

\begin{figure}%[H]
	%\vspace{-8mm}
	\centering
	\includegraphics[width=.6\linewidth]{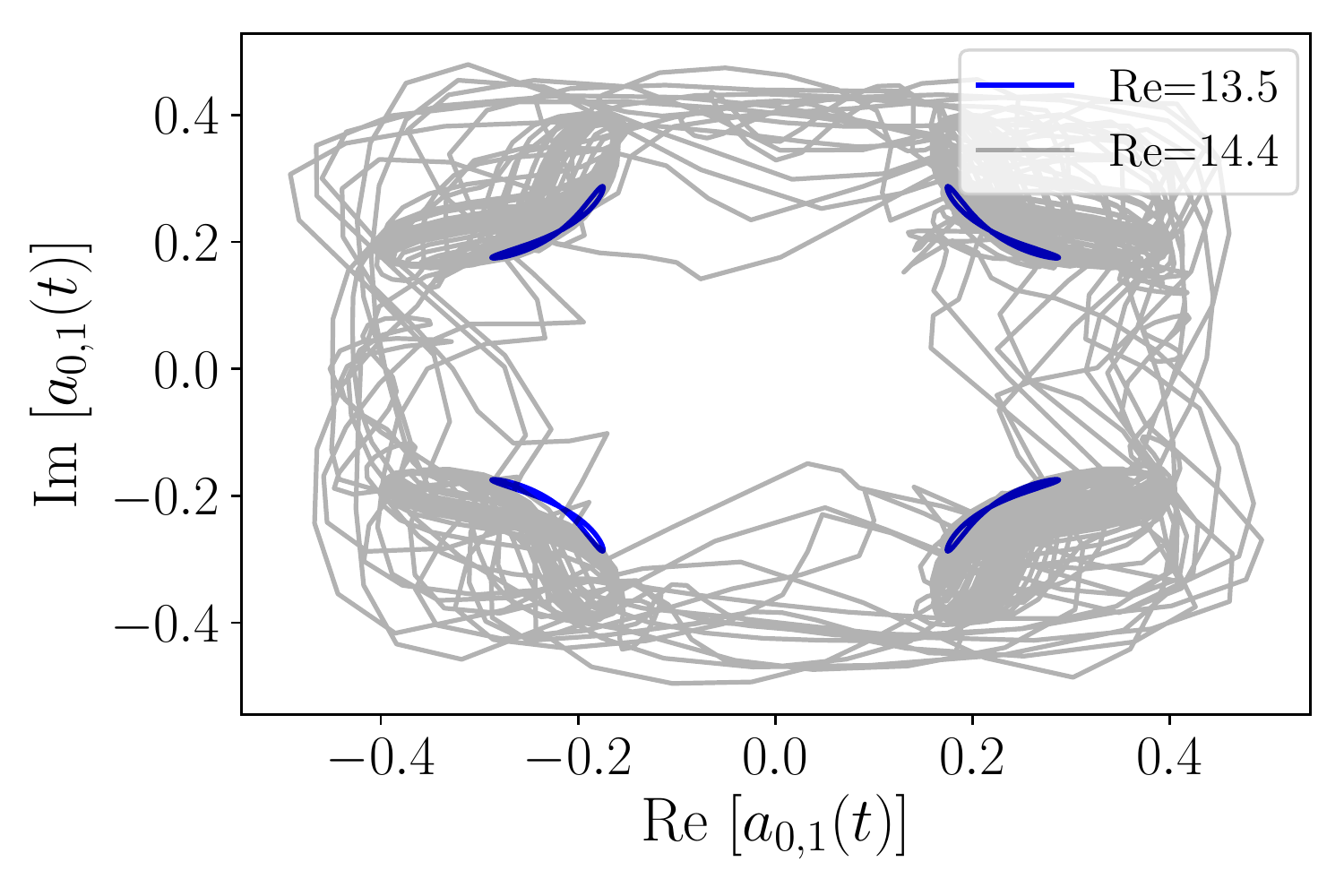}
	\caption{Evolution of the real and imaginary components corresponding to the  $a_{0,1}(t)$ Fourier mode for $\operatorname{Re}=13.5$ and $\operatorname{Re}=14.4$.}
	\label{re14d4re13d5}
\end{figure}

The NSE are evolved numerically in time in the vorticity representation on a $[d_x \times d_y]=[32 \times 32]$ grid following the pseudo-spectral scheme given by Chandler \& Kerswell \cite{chandler2013invariant}, which is based on the code by Bartello \& Warn \cite{bartello1996self}. We show here time series results for the two dynamical regimes considered in this work, an RPO regime at $\Rey=13.5$ and a chaotic regime at $\Rey=14.4$. Figure \ref{fig:sub1} shows the $KE$ evolution for an RPO obtained at $\operatorname{Re} = 13.5$.  Due to the discrete symmetries of the system, there are several RPOs \cite{armbruster1996symmetries}, as we further discuss below. Figure \ref{fig:sub2} shows the $KE$ evolution for a trajectory at  $\operatorname{Re} = 14.4$. The dynamics are characterized by quiescent intervals where the trajectories are close to RPOs (which are now unstable), punctuated by heteroclinic-like excursions between the RPOs, which are indicated by the intermittent increases of the $KE$. The RPOs are all related by the symmetries $\mathscr{S}$ and $\mathscr{R}$ \cite{armbruster1996symmetries, platt1991investigation, nicolaenko1990symmetry}. This behavior can also be seen in Figure \ref{fig:subID}, where the black curve corresponds to the time evolution of $D$ and the blue curve to the time evolution of $I$. Figure \ref{re14d4re13d5}, shows a state-space projection of a trajectory onto the plane $\operatorname{Re}\left[a_{0,1}(t)\right]-\operatorname{Im}\left[a_{0,1}(t)\right]$ where $a(k_x,k_y,t) = a_{k_x,k_y}(t)=\mathcal{F}\{\omega(x,y,t)\}$ is the discrete Fourier transform in $x$ and $y$. The grey curve corresponds to $\operatorname{Re} = 14.4$ and the different blue curves show four different RPOs related by the shift-reflect symmetry $\mathscr{S}$ at $\operatorname{Re} = 13.5$.

%"Talk about hybrid autoencoder, time stepper, data used, normalization of data, loss function, method of slices, Shift reflect symmetry correction, rotation correction"

%Notice that for $n=2$ two SRs correspond to a shift in $\pi$.

\section{Data-driven dimension reduction and dynamic modeling} \label{sec:AEs}

\subsection{Dimension reduction with autoencoders}\label{sec:aesub}

%We post process the data by subtracting the mean and dividing by the standard deviation. By doing this we ensure that all features of the flow lie in the same range.

%, hence $\omega(t) \in \mathbb{R}^N$
To learn a minimal-dimensional model for the two-dimensional Kolmogorov flow we first have to find a low-dimensional nonlinear mapping from the full state to the reduced representation. For this purpose we consider a common machine learning architecture known as an undercomplete autoencoder (AE), whose purpose is to learn a reduced representation of the state such that the reconstruction error with respect to the true data is minimized. The AE consists of an encoder, $\mathcal{E}(\cdot)$, that maps from the full space $\mathbb{R}^{N}$ to the lower dimensional latent space $h(t) \in \mathbb{R}^{d_h}$ (i.e., coordinates on the manifold $\IM$), and a decoder, $\mathcal{D}(\cdot)$, that maps back to the full space. Flattened versions of $\omega (x,y,t)$ are used, which we refer from this point on as $\omega(t)$, so $N = 32 \times 32 = 1024$. We shall see that the latent space dimension $d_h$ will be much smaller than the dimension $N$ of the full spatially-resolved state. The encoder $\mathcal{E}(\omega(t))$ is a coordinate mapping from $\mathbb{R}^{N}$ to $\IM$, and the decoder $\mathcal{D}(h(t))$ is the mapping back from $\IM$ to $\mathbb{R}^{N}$.

%\MDG{The next sentence and some of the text below need a rewrite.  First of all, as we discussed, you are using the symbol $\omega$ both to denote the vorticity field $\omega(x,y,t)$ and the discretized and flattened version of it $\omega(t)$. Second of all, you write that `From this point on $N$ will correspond to a flattened version\ldots". What you mean to say is that from this point on, the full state will be taken as the flattened version of the $32\times 32$ discretized vorticity field, so $N=32\times32=1024$. Finally, what actually gets fed into the autoencoder is the phase-aligned state and symmetry operations on it.}  

%From this point on $N$ will correspond to a flattened version of the $\omega(x,y,t)$ snapshots where $N=32^2=1024$

We train the AEs with $\omega(t)$ obtained from the evolution of NSE for the original data as well as accounting for the discrete and continuous symmetries. By accounting for the symmetries it is expected that the networks will perform better, by not having to learn the symmetries in the latent space mapping. We account for the continuous symmetry in $x$, $\mathscr{T}_{l}$, with the method of slices \cite{budanur2015periodic, budanur2015reduction}. The $k_x=1, k_y=0$ Fourier mode is used to find the spatial phase: $\phi_x(t)=\operatorname{atan} 2\left\{\operatorname{Im}\left[a_{1,0}(t)\right], \operatorname{Re}\left[a_{1,0}(t)\right]\right\}$. This can then be used to phase-align the vorticity snapshots such that this mode is a pure cosine: $\hat{\omega}(x,y,t)=\mathcal{F}^{-1}\left\{\mathcal{F}\{\omega(x,y,t)\} e^{-i k \phi_x(t)}\right\}$. Doing this ensures that the snapshots lie in a reference frame were no translation happens in the $x$ direction. We will learn evolution equations for both $\hat{\omega}(t)$ and $\phi_x(t)$, which we will denote as the pattern dynamics and phase dynamics, respectively. We also consider the shift-reflect (SR) symmetry, $\mathscr{S}$, as well as the rotation through $\pi$, $\mathscr{R}$. To account for the SR symmetry the goal is to collapse the phase-aligned snapshots to the same common state. We can define two indicator functions such that the SR subspace is specified. The first one, $I_{Even}=\operatorname{sgn}(\phi_y)$, where $\phi_y(t)=\operatorname{atan} 2\left\{\operatorname{Im}\left[a_{0,1}(t)\right], \operatorname{Re}\left[a_{0,1}(t)\right]\right\}$ is the spatial phase in $y$. The second indicator function is $I_{odd}=\operatorname{sgn}(\operatorname{Re}[a_{2,0} (t)])$, the sign of the real part of the second Fourier mode in $x$. We can then map the vorticity snapshots in such a way that $I_{Even},I_{Odd}>0$ by applying SR operations to the state. The rotation symmetry is accounted for, on top of the SR symmetry, by minimizing the $l^2$-norm of the data with respect to a template snapshot. This is done by applying the discrete operation that rotates and shift-reflects the vorticity snapshots and selecting the snapshot that minimizes the norm. We note that we take a different approach for reducing the symmetries compared to previous research on symmetry-aware AEs \cite{kneer2021symmetry}.

%This method is different from previous research 

%\MDG{Rewrite this paragraph.  You are implying that we are truncating the representation with PCA} 

Previous work \cite{linot2020deep} has shown that training a NN to learn the difference between the data and the projection onto the leading PCA basis vectors improved reconstruction performance compared to learning a latent space directly from the full data. To present the framework, we will use the phase-aligned and flattened vorticity $\hat{\omega} (t)$, since that is what we use for the time-evolution. Below, however, we will present some results where other versions of the data are used -- e.g.~the data with phase-shifting. The autoencoder aspect of the analysis is identical.  

\begin{figure}%[h]
	%\vspace{-8mm}
	\centering
	\includegraphics[width=0.6\linewidth]{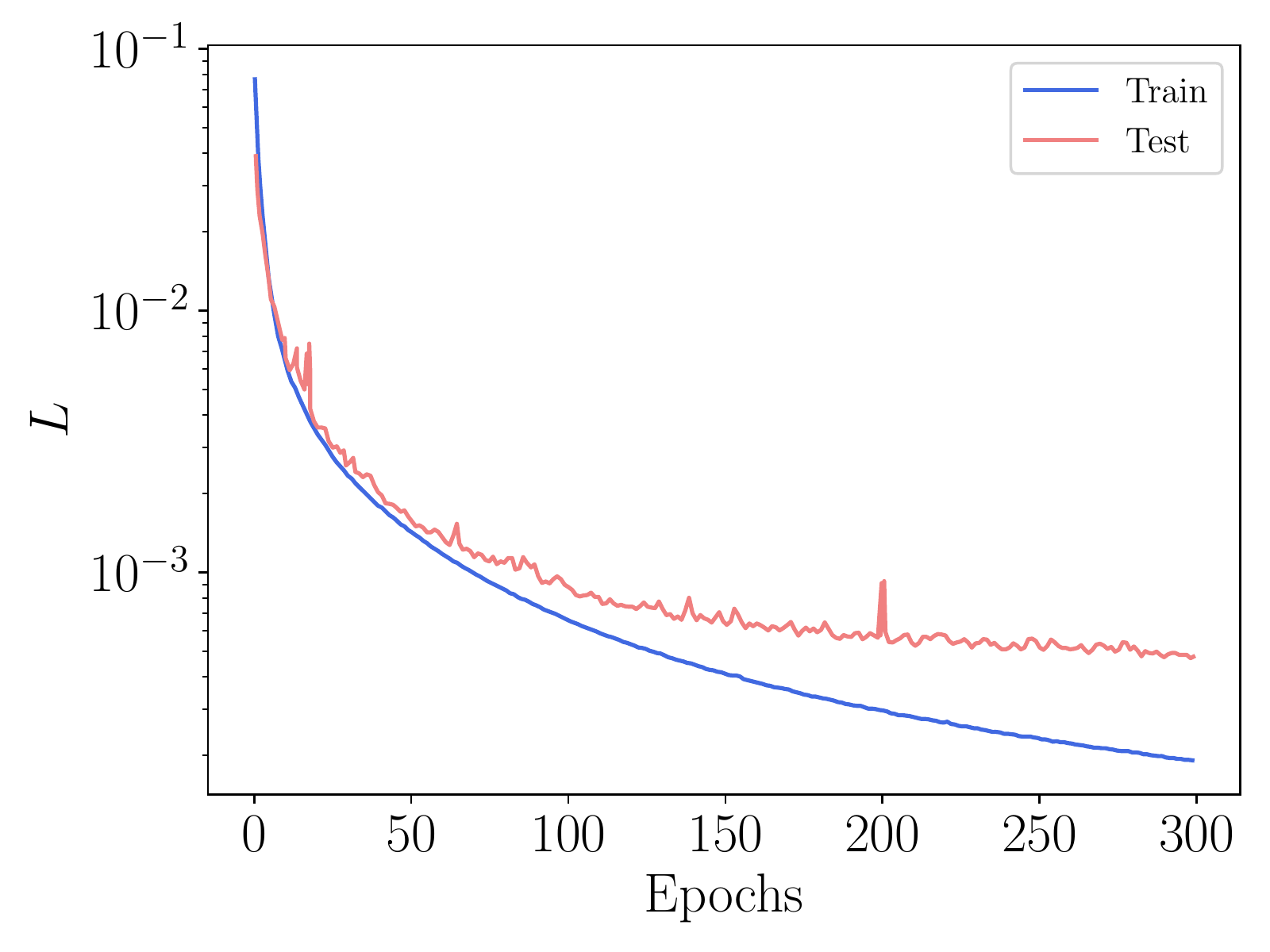}
	\caption{Autoencoder loss versus epochs over training and test data sets corresponding to a trial from the case $\operatorname{Re}=14.4$, $d_h=9$.}
	\label{Loss}
\end{figure}

We begin the process by computing the projection of the data onto the first $d_h$ basis vectors, $P_{d_{h}} U^{T}\hat{\omega}(t)$. We then seek to learn a $d_h$-dimensional correction to that projection, $E\left(U^{T} \hat{\omega}(t)\right)$ -- the sum of these is the latent-space representation $h(t)$. In other words, the encoding step learns the deviation from PCA
\begin{equation}
E\left(U^{T} \hat{\omega}(t)\right)=h(t)-P_{d_{h}} U^{T} \hat{\omega}(t).
\label{encoder}
\end{equation}
We emphasize that this step \emph{is not} simply a projection onto a linear subspace defined by $d_h$ PCA modes-- rather it is an approach that learns the deviation of the data from that projection.
Similarly the decoding section learns the difference 
\begin{equation}
D(h(t))=U^{T} \tilde{\hat{\omega}}(t)-\left[\begin{array}{c}h(t) \\ 0\end{array}\right],
\label{decoder}
\end{equation}
where $\tilde{\hat{\omega}}(t)$ corresponds to the reconstruction of $\hat{\omega} (t)$. Inserting Equation \ref{encoder} into Equation \ref{decoder} and noting that by definition $\tilde{\hat{\omega}}(t) = U[P_{d_{h}} U^{T} \hat{\omega}(t), P_{d-d_{h}} U^{T} \hat{\omega}(t)]^{T}$  we get that the exact solution satisfies $E\left(U^{T} \hat{\omega}(t)\right)+D_{d_{h}}((h(t))=0$. To satisfy this constraint we add it to the loss function as a penalty to obtain
\begin{equation}
L=\|\hat{\omega}(t)-\tilde{\hat{\omega}}(t)\|^{2}+\alpha_L \left\|E(U^{T}\hat{\omega}(t))+D_{d_{h}}(h(t))\right\|^{2}
\end{equation}
where $\| \cdot \|$ is the $l^2$-norm and we select $\alpha_L=1$.
% \CEPrevise{Here the penalty controlled by $\alpha_L$ does not affect the size of $h(t)$ of dimension $d_h \ll N$, and only affects the loss whenever this sum is greater than zero.}
We can now train the AEs by minimizing $L$ via stochastic gradient descent. We train 4 AEs at each of several values of $d_h$ to study the MSE of the reconstruction of $\hat{\omega} (t)$. All models were trained for 300 epochs with an Adam optimizer using Keras. After 300 epochs no further improvement over the test data was observed; see Figure \ref{Loss}. The training data consists of long time series from the direct simulations, with initial transients removed. We use a total of $10^5$ snapshots separated by $\tau=5$ time units for $\text{Re}=14.4$, and $10^4$ snapshots separated by $\tau=5$ for $\text{Re}=13.5$. We do an $80\%/20\%$ split for training and testing respectively.  Figure \ref{framework1}\textcolor{blue}{a} shows a summary of the AE and Table \ref{tablenn} gives information on the layer dimensions, and activations used in each layer of the encoder and decoder. At each value of $d_h$, the model with the smallest MSE over a test data set from the phase-aligned data is then selected for the discrete time map. We will show in  Section \ref{sec:autoencoders} that factoring out the phase dramatically increases AE performance.

%The model with the smallest MSE over a test data set from the phase-aligned data is then selected for the discrete time map NN as will be shown in the next section. We will show in the AE results how accounting for the phase increases performance drastically.

%Figure \ref{framework1} shows a summary of the AE. The model with the smallest MSE over a test data set from the phase-aligned data is then selected for the discrete time map. We will show in the AE results how accounting for the phase increases performance drastically.

%Figure \ref{framework1} shows a summary of the AE, as well as the discrete time map and how we reconstruct from the predicted data in time. The model with the smallest MSE over a test data set from the phase-aligned data is then selected for the discrete time map. We will show in the AE results how accounting for the phase increases performance drastically.

\subsection{Time evolution via a dense NN}\label{sec:aetime}

% For the time evolution results we will consider the phase-aligned data. We will see in the results section how phase-a

\begin{figure}%[h]
	%\vspace{-8mm}
	\centering
	\includegraphics[width=1\linewidth]{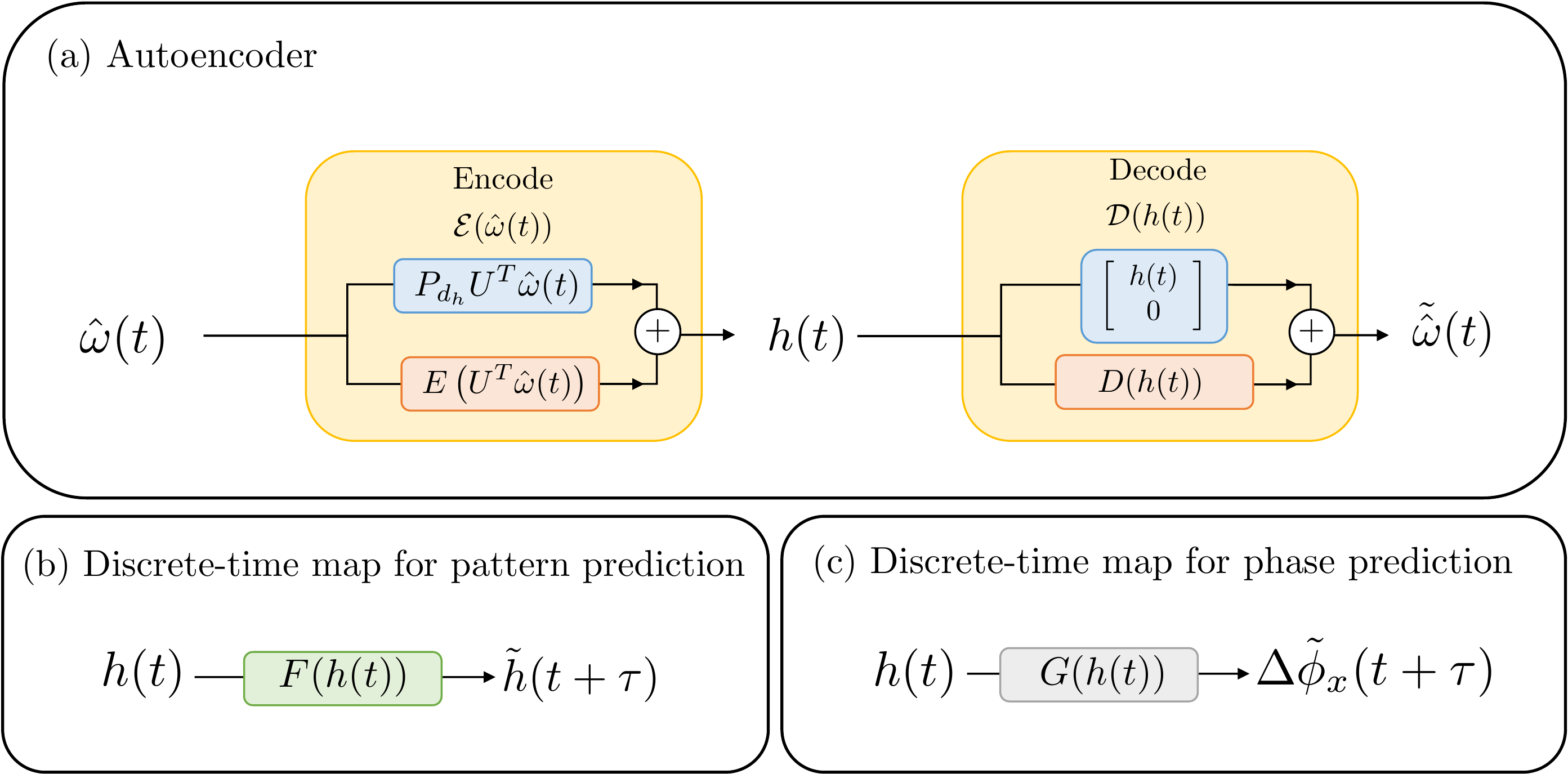}
	\caption{Neural network frameworks for (a) autoencoder (b) discrete-time map for pattern prediction and (c) discrete-time for phase prediction.}
	\label{framework1}
\end{figure}

After finding $h(t)$ from the AEs, we seek a discrete-time map
\begin{equation}
	h(t+\tau)=F(h(t))
\end{equation}
that evolves $h(t)$ from time $t$ to $t+\tau$. We fix $\tau=5$. The function $F$ is also expressed as a dense NN. Here we train 5 NNs for the different $d_h$ cases with the following loss
\begin{equation}
L_t=\|\tilde{h}(t+\tau)-h(t+\tau)\|^{2},
\end{equation}
where $h(t+\tau)$ comes from true data and $\tilde{h}(t+\tau) = F(h(t))$ from the prediction, and select the one with the best performance. For the discrete time map we trained for 600 epochs with the use of a learning rate scheduler. In this case we noticed an increase in performance when dropping the learning rate hyperparameter by an order of magnitude after 300 epochs. Figure \ref{framework1}\textcolor{blue}{b}  shows a summary the framework just described, and Table \ref{tablenn} gives information on the layer dimensions and activations used in each layer.

%Table \ref{tablenn} shows the architectures for the different NNs considered including the encoder and decoder for the HNN-AE and the discrete time map for $h(t)$.

%\subsection{Phase prediction from pattern}\label{sec:aephase}

As discussed previously, the time evolution is done in the phase-aligned space. To complete the dynamical picture we seek a discrete-time map for the phase evolution 
\begin{equation}
	\Delta \tilde{\phi}_x(t + \tau) = G(h(t)),\label{eq:phaseevolution}
\end{equation}
where $\Delta \phi_x(t + \tau)=\phi_x(t + \tau)-\phi_x(t)$.  Because of translation equivariance, the actual phase is only unique to within a constant. We train 5 NNs for the the different $d_h$ cases with the following loss
\begin{equation}
L_p=\|\Delta \tilde{\phi}_x(t + \tau)-\Delta \phi_x(t + \tau)\|^{2},
\end{equation}
such that $\Delta \tilde{\phi}_x(t + \tau) = G(h(t))$. Figure \ref{framework1}\textcolor{blue}{c}  shows a summary of the framework we have described, and Table \ref{tablenn} gives information on the layer dimensions and activations used in each layer.

\begin{table}[h]
\caption{Neural network layer dimensions and activations used in each layer. Sigmoid function are denoted 'S'.}
$$
\begin{array}{lccc}\hline \hline & \text { Function } & \text { Shape } & \text { Activation } \\ \hline \text { Encoder } & E & 1024: 5000:1000:d_{h} & \text { S:S:S } \\ \text { Decoder } & D & d_{h}: 1000: 5000: 1024 & \text { S:S:linear } \\ \text { Evolution } & F & d_{h}: 500: 500: d_{h} & \text { S:S:linear } \\ \text { Phase Prediction } & G & d_{h}: 500: 500:500: 1 & \text { S:S:S:linear } \\ \hline \hline\end{array}
$$
\label{tablenn}
\end{table}

%The encoding and decoding portions

\section{Results} \label{sec:Results}
We present results as follows. First we will show the AE performance for the various $d_h$ and symmetries considered. We then report results for time evolution models, again studying performance as a function of the number of dimensions. Both evolution of the pattern and phase dynamics are considered. {We wrap up the results by predicting bursting events based on  the low-dimensional representation.}

%\vspace{-80pt}
\subsection{Dimension reduction with autoencoders}\label{sec:autoencoders}

\begin{figure}%[H]
	%\vspace{-8mm}
	%\centering
	\begin{subfigure}{0.5\textwidth}
		%\centering
		\includegraphics[width=1\linewidth]{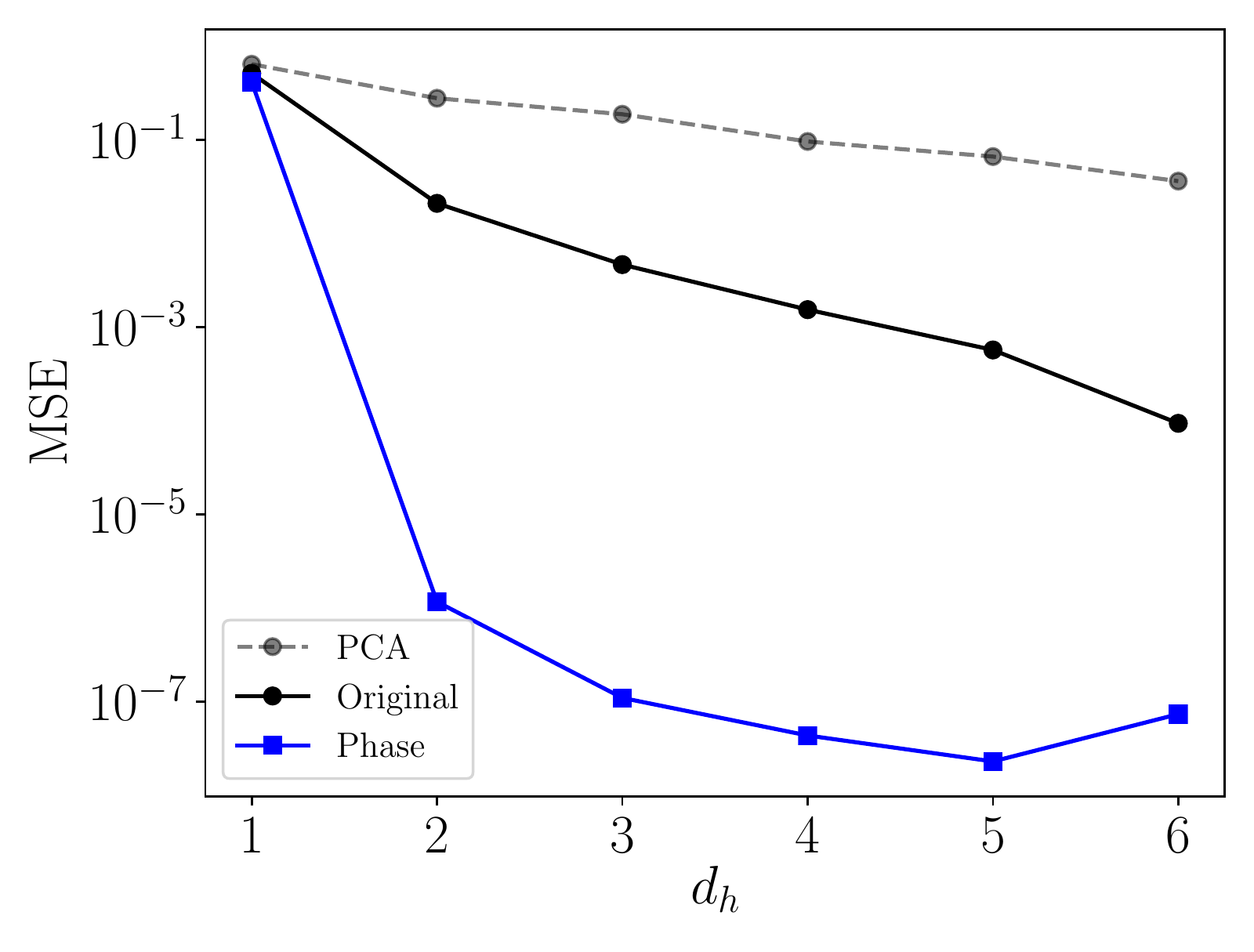}
		\caption{}
		\label{fig:subMSEre13d5}
	\end{subfigure}%
	\begin{subfigure}{0.5\textwidth}
		%\centering
		\includegraphics[width=1\linewidth]{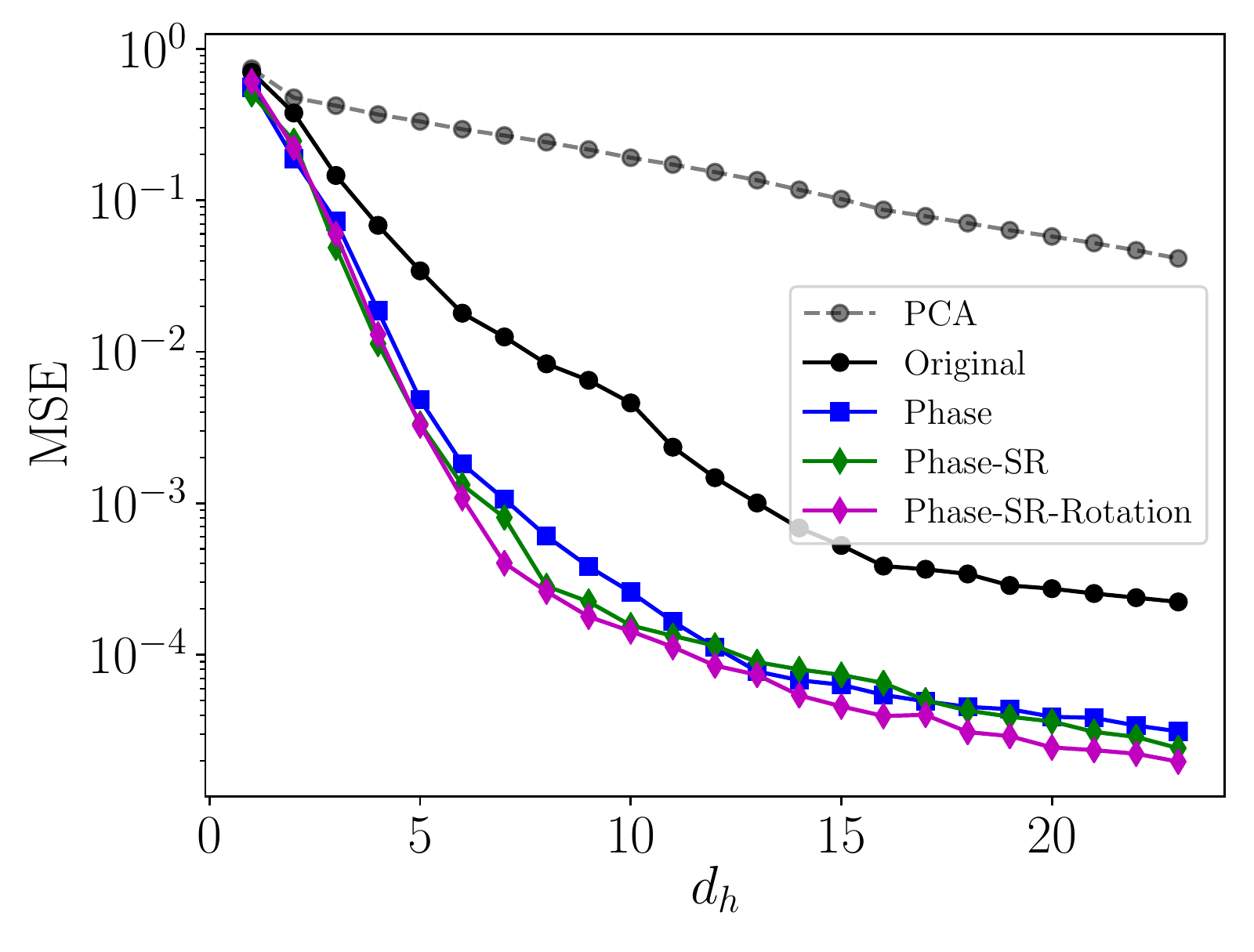}
		\caption{}
		\label{fig:subMSEre14d4}
	\end{subfigure}
%\begin{comment}
	%\begin{subfigure}{0.5\textwidth}
	%\centering
	%\includegraphics[width=.9\linewidth]{figures/MSE_Re20.pdf}
	%\caption{}
	%\label{fig:subMSEre20}
%\end{subfigure}
%\end{comment}
	\caption{MSE versus dimension $d_h$ over the test data corresponding to (a) $\operatorname{Re}=13.5$ and (b) $\operatorname{Re}=14.4$. The PCA curve corresponds to the MSE of the reconstruction for the test data set with respect to the true data $\omega (t)$, with no symmetries factored out, using the truncated $U$ into $d_h$ dimensions such that $\tilde{\omega}(t)=U_{d_h}U^T_{d_h} \omega (t)$ ; the `Original', `Phase', `Phase-SR', and `Phase-SR-Rotation' curves correspond to the MSEs of the reconstruction for the test data set with respect to the true data using AEs. In the curve labeled `Original', no symmetries are factored out and in the other curves the corresponding symmetries in the labels are factored out. }
	\label{re13d5_14d4_auto}	

\end{figure}

We begin by showing results for $\operatorname{Re} = 13.5$. In Figure \ref{fig:subMSEre13d5} we see the MSE versus $d_h$ trend where the grey curve corresponds to the PCA reconstruction for the original data ($\tilde{\omega}(t)=U_{d_h}U^T_{d_h} \omega (t)$), the black curve to the AE with the original data, and the blue curve to the AE with the phase factored out before training. The MSE is calculated over the test data set. Notice that, as expected, the AEs perform better than PCA. This is because of the nonlinearities that are added to the linear optimal latent space found in PCA in combination with the nonlinear decoder. The blue curve exhibits a sharp drop in the MSE at a dimension of $d_h=2$, which is the correct embedding dimension for a limit cycle. This happens because the phase is accounted for; the dynamics of the system in the phase-aligned reference frame corresponds to a PO and the autoencoder does not have to learn all the possible phases due to the continuous translation in $x$. The overall embedding dimension is $d_h+1 = 3$, where 1 corresponds to the phase. Hence we are able to estimate the dimension for this system by looking at the drop in the MSE curve. 
%\MDG{incorrect as written -- rewrite or eliminate} We obtain a drop at $d_h=2$ because the PO is represented with one cartesian coordinate, so we should expect that dividing the space into charts would give the $d_h=1$ as was shown by Floryan \& Graham \cite{floryan2021charts}.

%Notice that the black curve does a poor job compared with AE, Phase\MDG{is the reader going to know what you mean by ``AE, Phase"? This reads awkwardly}. This is because of the added complexity of the AE having to learn the translation in $x$. The Phase-SR correction shows to help slightly on the MSE reconstruction but it is still in the same order of magnitude as in the Phase case.

%, which agrees with the real dimension of the system. 

%Notice that the dynamical representation 

%This case represents an RPO as discussed previouslyThis case represents an RPO as discussed previously

%Notice that

%We expect to see a decrease on $MSE$ that as $d_h$ increases. 

%Need to comment that in next section we will predict in latent space dimension

\begin{comment}
\begin{figure}%[H]
	%\vspace{-8mm}
	\centering
	\includegraphics[width=.7\linewidth]{figures/MSE_Re13d5.pdf}
	\caption{MSE for test data corresponding to $\operatorname{Re}=13.5$}
	\label{re13d5intro}
\end{figure}
\end{comment}

We now consider the $\operatorname{Re} = 14.4$ case, where the dynamics are chaotic, moving between the regions near the now unstable RPOs. In Figure \ref{fig:subMSEre14d4} we show the same curves as in Figure \ref{fig:subMSEre13d5} but we also include the green and magenta curves, which in addition factor out the SR and the SR-Rotation symmetries respectively before training the AEs. These are included due to the added complexity of $\operatorname{Re}=14.4$, where the chaotic trajectory travels in the vicinity of the RPOs related by the symmetry groups previously discussed. A monotonic decrease in MSE can be seen for the different symmetries considered in the blue, green, and magenta curves, but no sharp drop is apparent. Instead we notice that the MSE drops at different rates in different regions. For example, in the blue curve corresponding to the phase aligned data, we see a sharp drop from $d_h=1-6$ followed by a more gradual drop from $d_h=6-13$. In the following sections we couple the dimension-reduction analysis with models for prediction of time evolution for the phase aligned data. We expect that this combination will help us determine how many dimensions are needed to correctly represent the state.

 %\hl{comment about other curves}. this can be conducive to the dim

%After approximately $d_h=13$ we see no significant improvement and most of the trends plateau.
\begin{comment}
\begin{figure}%[H]
	%\vspace{-8mm}
	\centering
	\includegraphics[width=.7\linewidth]{figures/MSE_Re14d4.pdf}
	\caption{MSE for test data corresponding to $\operatorname{Re}=14.4$}
	\label{re14d4auto}
\end{figure}
\end{comment}

\subsection{Time evolution as a function of dimension  - Short time predictions}\label{sec:dimredevshort}

\begin{figure}%[H]
	%\vspace{-8mm}
	%\centering
	\begin{subfigure}{0.5\textwidth}
		%\centering
		\includegraphics[width=0.9\linewidth]{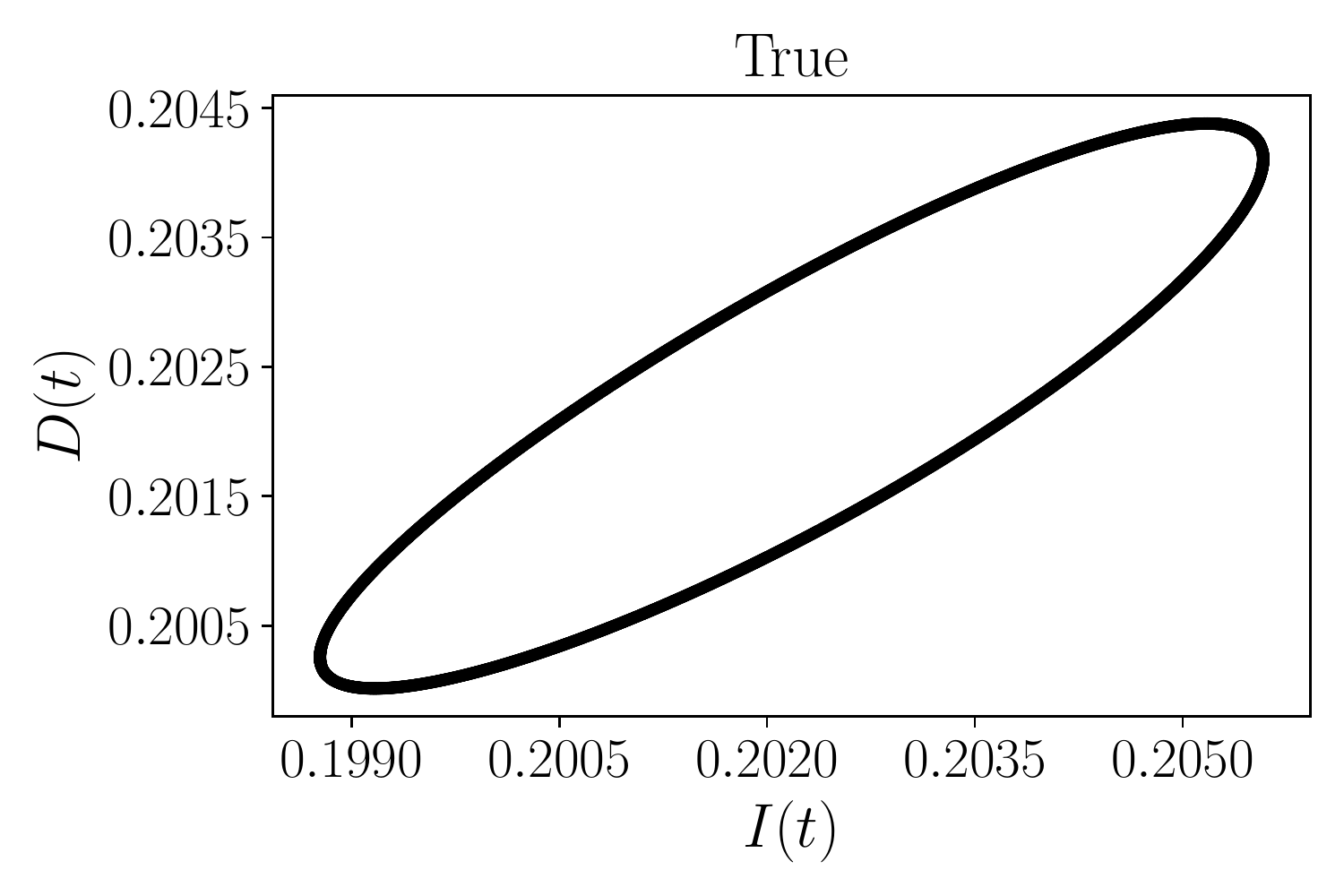}
		\caption{}
		\label{fig:subipdtruere13d5}
	\end{subfigure}%
	\begin{subfigure}{0.5\textwidth}
		%\centering
		\includegraphics[width=.9\linewidth]{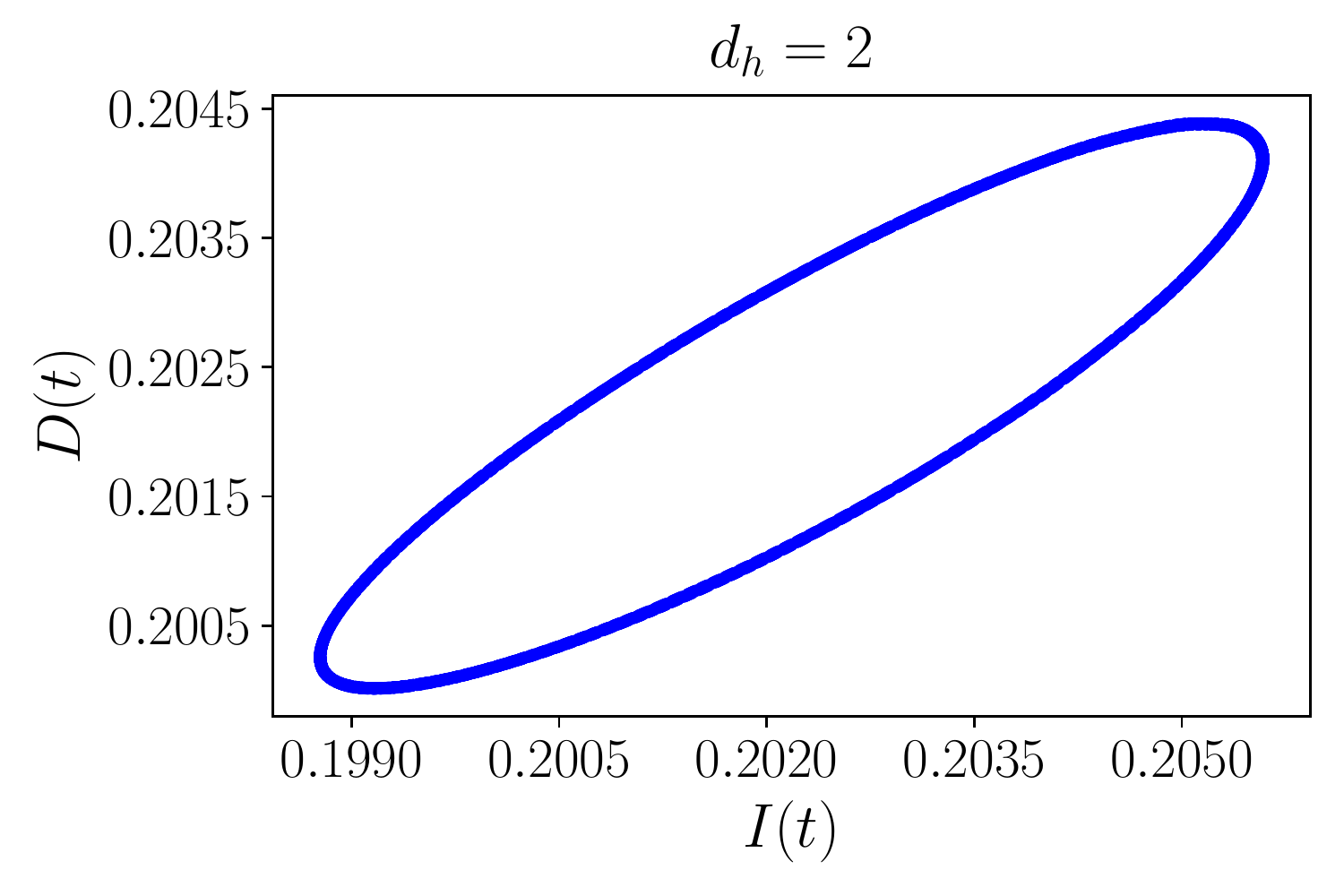}
		\caption{}
		\label{fig:subipd2_re13d5}
	\end{subfigure}
	\caption{Trajectory of $I(t)$ vs $D(t)$ corresponding to $\operatorname{Re}=13.5$ for (a) true and (b) predicted data  corresponding to dimensions $d_h=2$.}
	\label{re13d5PD}	
\end{figure}

The focus of this work is the chaotic dynamics at $\Rey=14.4$.  Before considering  that case, for completeness we briefly present results for $\operatorname{Re}=13.5$. In Figure \ref{re13d5PD} we see $D(t)$ versus $I(t)$ for the true and predicted dynamics at $d_h=2$; they are indistinguishable. At $d_h=1$, which is not shown, the model fails and the dynamics can not be captured. The reason for this is simple -- the embedding dimension for a limit cycle is two.

Now we return to the case of $\Rey=14.4$, focusing first on short-time trajectory predictions. The Lyapunov time $t_L$ for this system is approximately $t_L \approx 20$ \cite{inubushi2012covariant}, hence $t_L \approx 4\tau$. We take initial conditions $h(t) \in \mathbb{R}^{d_{h}}$ to evolve recurrently with the discrete time map $F(\cdot)$, such that $\tilde{h}(t+\tau) = F(h(t))$, $\tilde{h}(t+2\tau) = F(\tilde{h}(t+\tau))$, $\tilde{h}(t+3\tau) = F(\tilde{h}(t+2\tau))$ and so on. After evolving in time the data is then decoded to get $\tilde{\hat{\omega}}_h (t)$ and compared with $\hat{\omega} (t)$. We consider trajectories with ICs starting in the quiescent as well as in the bursting regions. The nature of the intermittency of the data makes it challenging to assign either bursting or quiescent labels. We consider a window of past and future snapshots and a criterion on $ \| \hat{\omega} (t) \| $ to make this decision, using the algorithm described in Algorithm \ref{alg:bursting}. 

\begin{figure}%[H]
	%\vspace{-8mm}
	\centering
	\includegraphics[width=.55\linewidth]{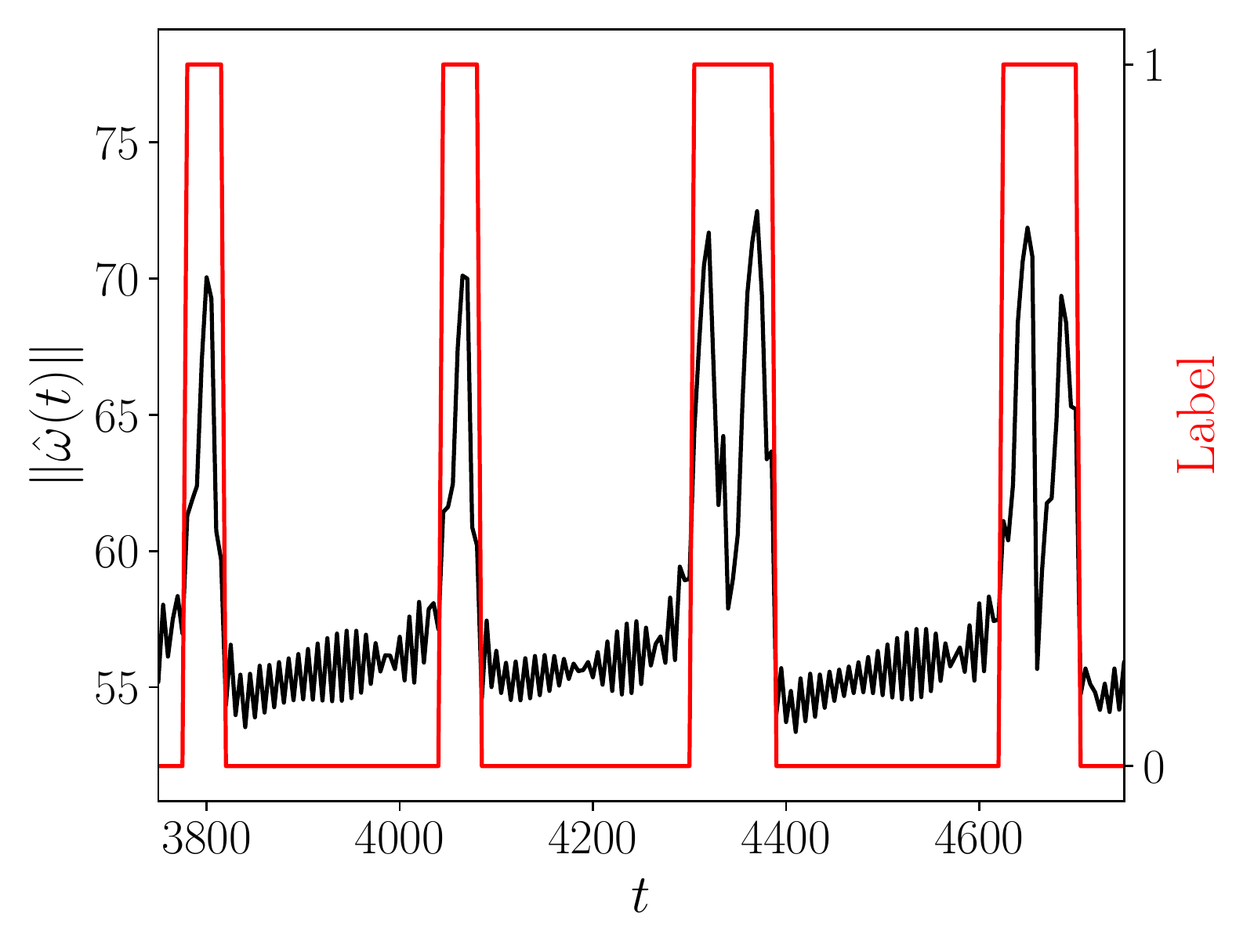} %This figure is in file BurstingPrediction/Convert_changetoKE.py
	\caption{Labeling of $\hat{\omega} (t)$ snapshots in a short time series where 1 corresponds to bursting and 0 to quiescent.}
	\label{re14d4labels}
\end{figure}

\noindent Doing this we ensure that snapshots that are contained in the bursting events and have a value of $ \| \hat{\omega} (t) \| $ similar to quiescent snapshots are correctly classified. We use a threshold on $ \| \hat{\omega} (t) \| $ to determine if a check is needed. For the classification strategy any snapshot above a threshold of 60 is classified as bursting with a label of 1, below 60 we enter a loop as shown in Algorithm \ref{alg:cap} to determine if it should be classified as bursting or quiescent, where quiescent corresponds to a label of 0. This check is needed to correctly label snapshots that have comparable $ \| \hat{\omega} (t) \| $ but are still in the bursting regime. Figure \ref{re14d4labels} shows a short time trajectory where the black line corresponds to $ \| \hat{\omega} (t) \| $ and the red to the 0/1 labels. Notice that, as shown in Algorithm \ref{alg:cap}, some of the data at the beginning and at the end of the time series will not be labeled, there are no past or future snapshots to compare to, and can be removed.

\begin{figure}%[H]
	%\vspace{-8mm}
	%\centering
	\begin{subfigure}{0.5\textwidth}
		%\centering
		\includegraphics[width=1\linewidth]{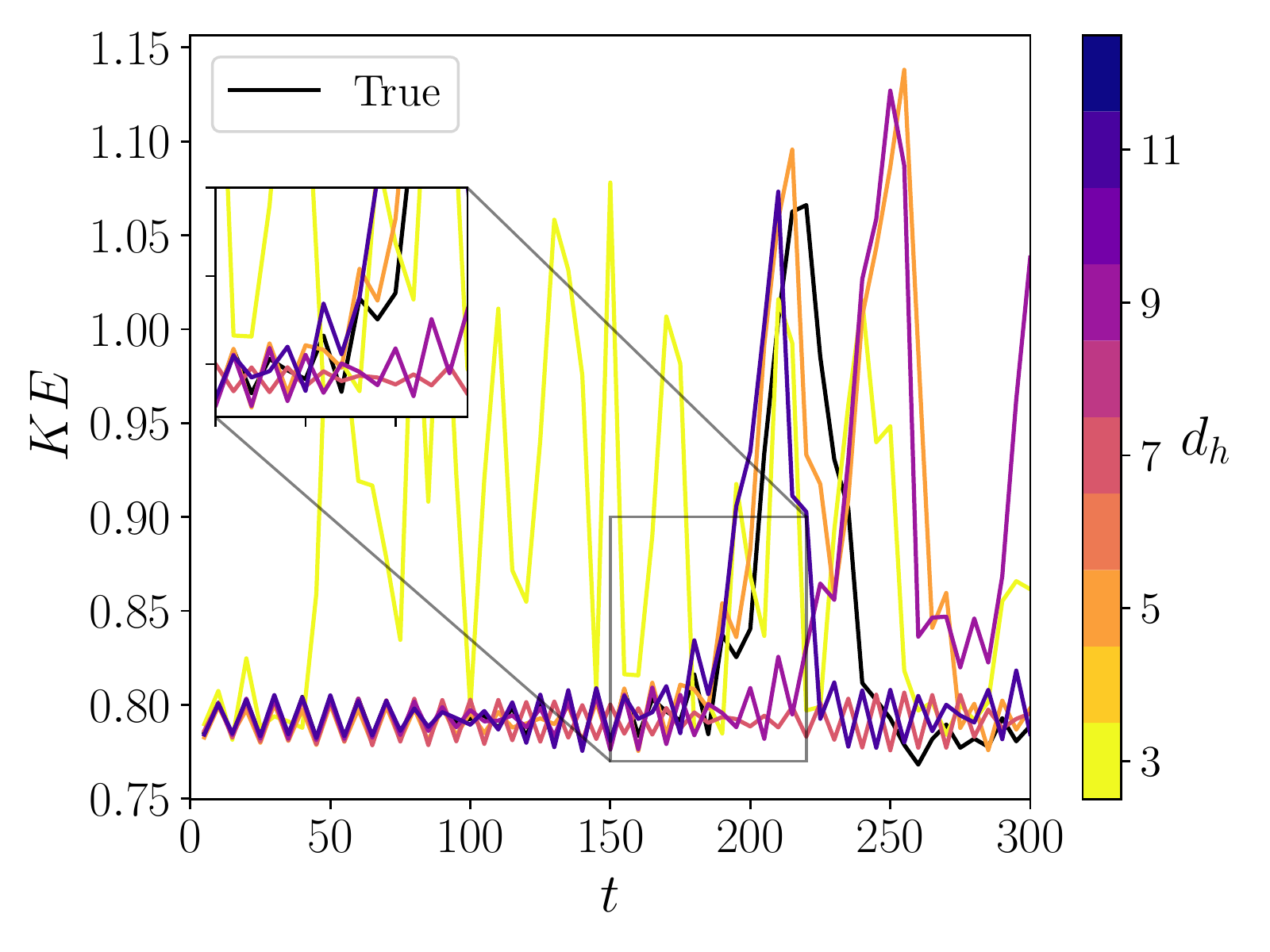} % This figure is in folder and file plotKEhibburplot/KEhibburplot.py
		\caption{}
		\label{KEhib}
	\end{subfigure}%
	\begin{subfigure}{0.5\textwidth}
		%\centering
		\includegraphics[width=1\linewidth]{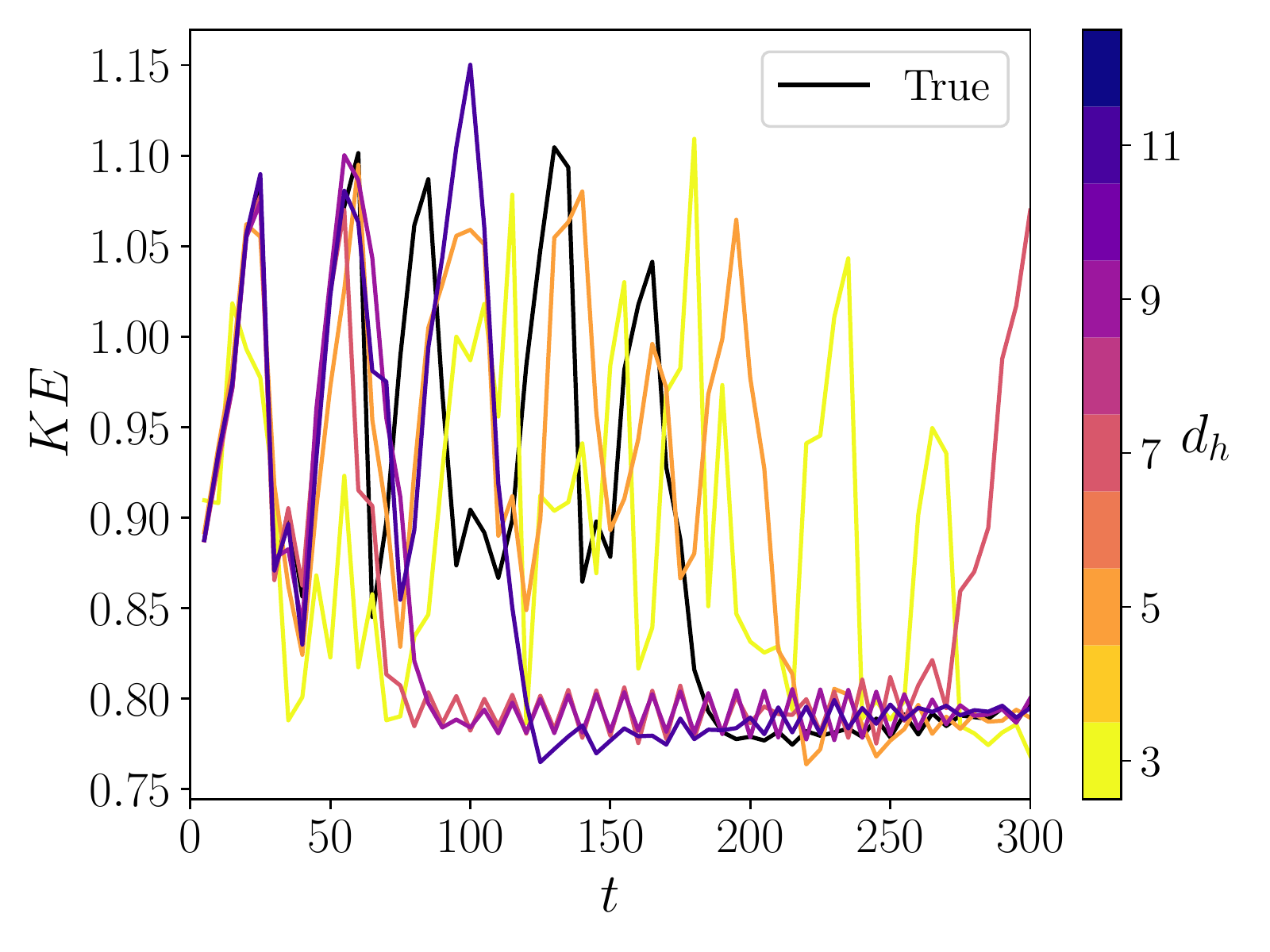}
		\caption{}
		\label{KEbur}
	\end{subfigure}
	
	\caption{Example trajectories of $KE$ at different $d_h$ for (a) a quiescent initial condition and (b) a bursting initial condition, for dimensions $d_h=3,5,7,9,$ and $11$.}
	\label{KEtots}
	
\end{figure}

After labeling the data as quiescent or bursting, we then consider the time evolution from ICs of $h(t)$ using the models of various dimensions. We will first show sample trajectories from ICs starting in the two regions, then show the ensemble-averaged prediction error as a function of time. Figure \ref{KEhib} shows the KE evolution for an IC starting in the quiescent region. The black curve corresponds to the true data and the colored curves to the different $d_h$ models. At a dimension of $d_h=3$ the predicted $KE$ diverges quickly with respect to the true $KE$. In the case of $d_h=5$ we see that the bursting event is correctly captured, but with a slight lag. However $d_h=7$ does not capture the bursting in this time frame considered. For $d_h=9$ the bursting event happens with a significant lag with respect to the true data and $d_h=11$ captures the event similar to $d_h=5$. Figure \ref{KEbur} shows the KE evolution for an IC starting in the bursting region. The black curve corresponds to the true data and the colored curves to the different $d_h$ models. At a dimension of $d_h=3$ the $KE$ stays bursting and does not show agreement with the true $KE$. However $d_h=5$ shows better agreement and is also capable of closely predicting the end of the bursting event. In the case of $d_h=7,9,$ and $11$ these agree closely with the $KE$ evolution before traveling to the quiescent region.

\begin{algorithm}[H]
	\caption{Quiescent/Bursting labeling of vorticity snapshots}\label{alg:cap}
	\begin{algorithmic}
		\State $W \gets [\hat{\omega}(t_1),\hat{\omega}(t_2) \cdots]$ \Comment{Matrix with $N_s$ vorticity snapshots, $W \in \mathbb{R}^{N \times N_s}$}
		\State $S$ \Comment{Initialize label array $S$}
		\State $W_{l2} \gets \| W \|$ \Comment{Calculate $l^2$-norm of snapshots, $W_{l2} \in \mathbb{R}^{ N_s}$ }
		\State $b \gets 10$ \Comment{Number of past snapshots in time to consider }
		\State $f \gets 10$  \Comment{Number of future snapshots in time to consider }
		\For {$i=b$, $b+1,\ldots N_s-f$} \Comment{$i$ is snapshot I.D.}
		\If{$W_{l2}[i]<60$ }
		\State $d_p \gets \operatorname{abs}(W_{l2}[i-b:i]-W_{l2}[i])$ \Comment{Difference between current and past snapshots}
		\State $b_p \gets \operatorname{sum}(d_p>5)$ \Comment{Sums values that exceed a threshold of 5 (user defined)}
		\State $d_f \gets \operatorname{abs}(W_{l2}[i:i+f]-W_{l2}[i])$ \Comment{Difference between current and future snapshots}
		\State $b_f \gets \operatorname{sum}(d_f>5)$ \Comment{Sums values that exceed a threshold of 5 (user defined)}
		\If{$b_p = 0$ or $b_f = 0$ }
		\State $S[i-b] \gets 0$
		\Else 
		\State $S[i-b] \gets 1$
		\EndIf
		\Else
		\State $S[i-b] \gets 1$
		\EndIf
		\EndFor
	\end{algorithmic}\label{alg:bursting}
\end{algorithm}

%In the rest of the paper we will keep showing results for this range of $d_h$ \MDG{be more specific -- do you mean 5-7?}. %As we will see, increasing dimension \MDG{beyond 7} does not drastically improve performance of time series data.

\begin{figure}%[H]
	%\vspace{-8mm}
	%\centering
	\begin{subfigure}{0.5\textwidth}
		%\centering
		\includegraphics[width=0.9\linewidth]{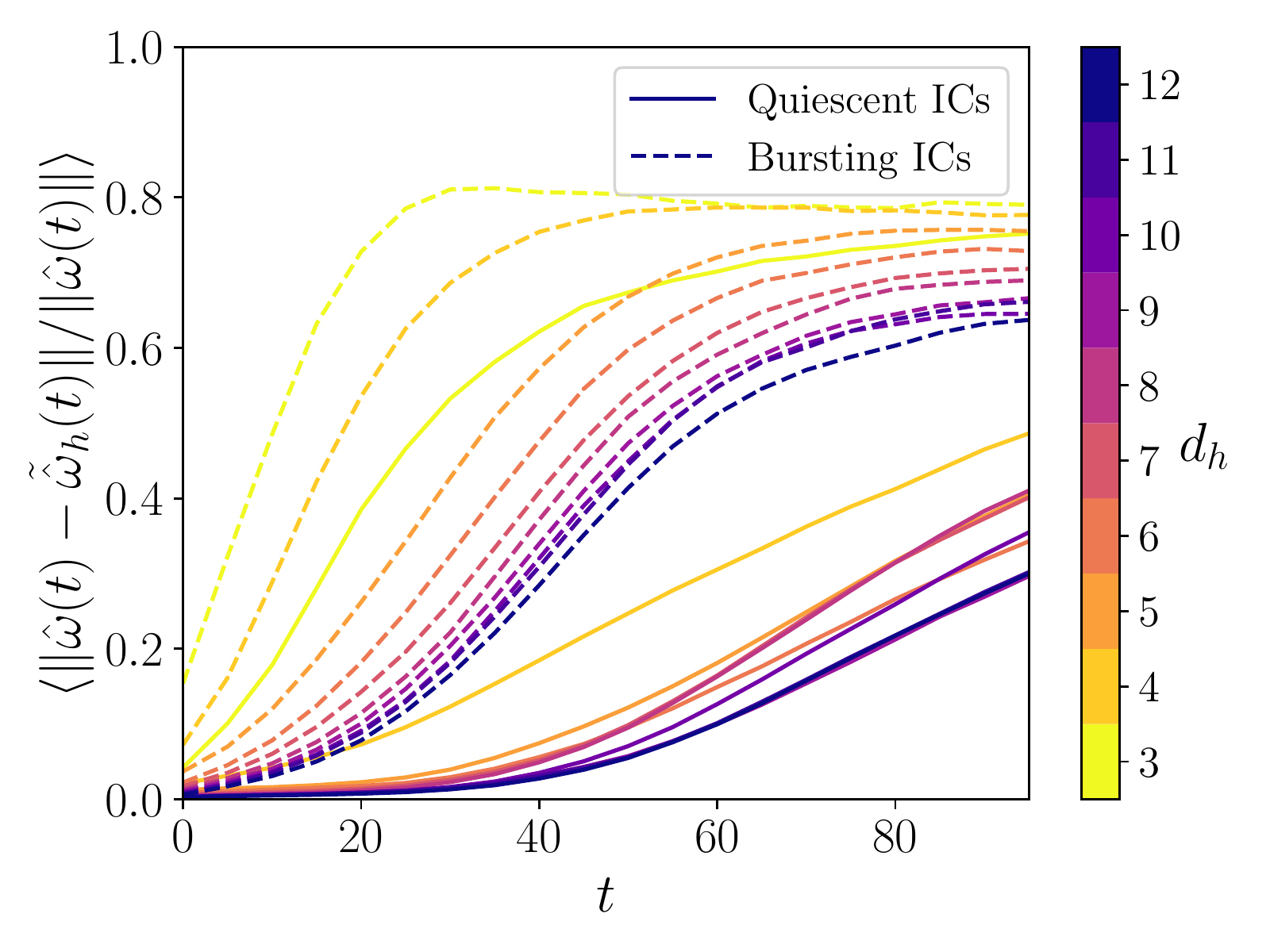} % This was in file getplotslatexfixed.py, new version is in file true_v_truedec in folder TrajectoryDiffCalc
		\caption{}
		\label{re14d4ICs}
	\end{subfigure}%
	\begin{subfigure}{0.5\textwidth}
		%\centering
		\includegraphics[width=.9\linewidth]{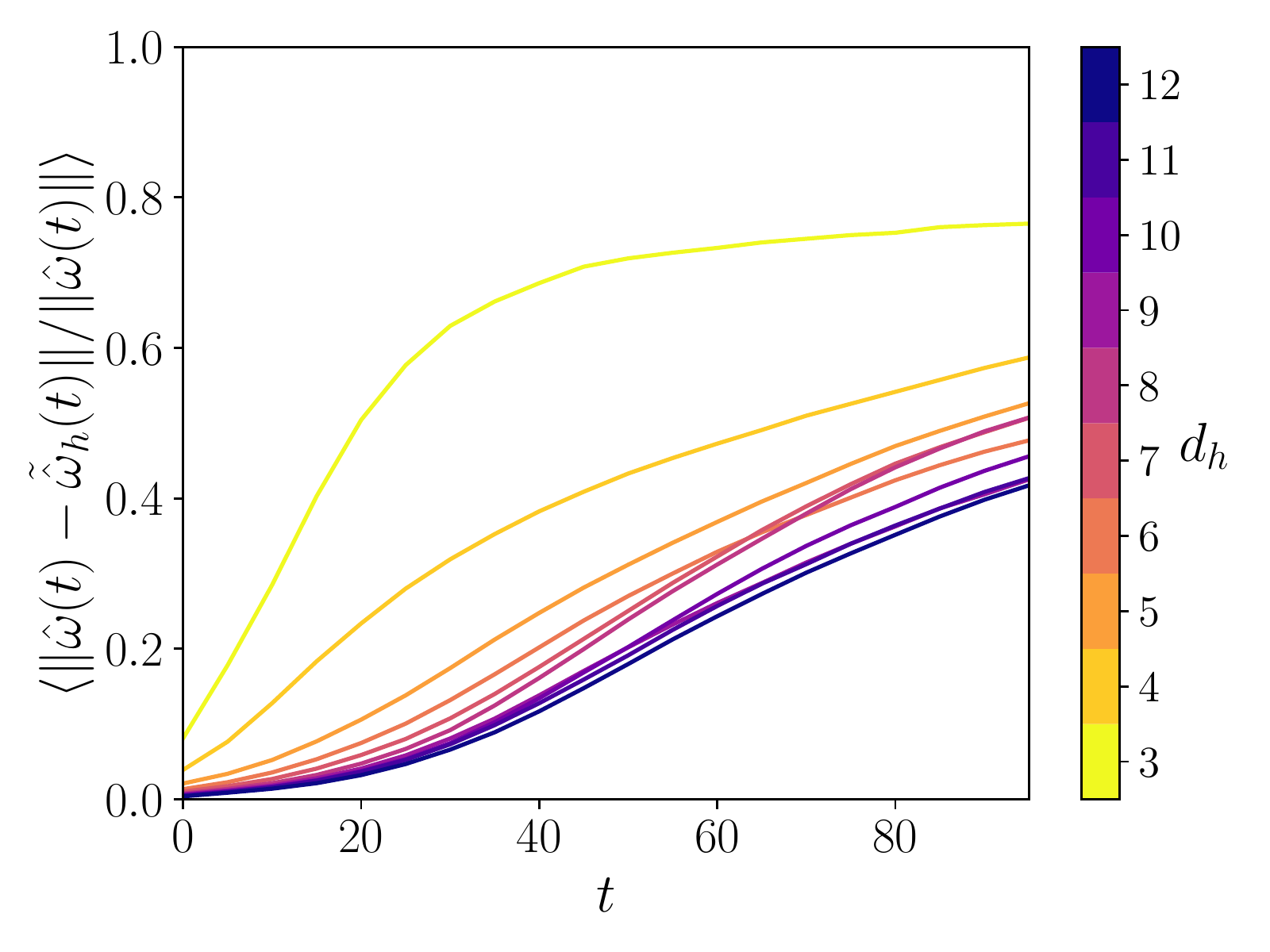}
		\caption{}
		\label{re14d4ICstotal}
	\end{subfigure}

        \begin{subfigure}{0.5\textwidth}
		%\centering
		\includegraphics[width=.9\linewidth]{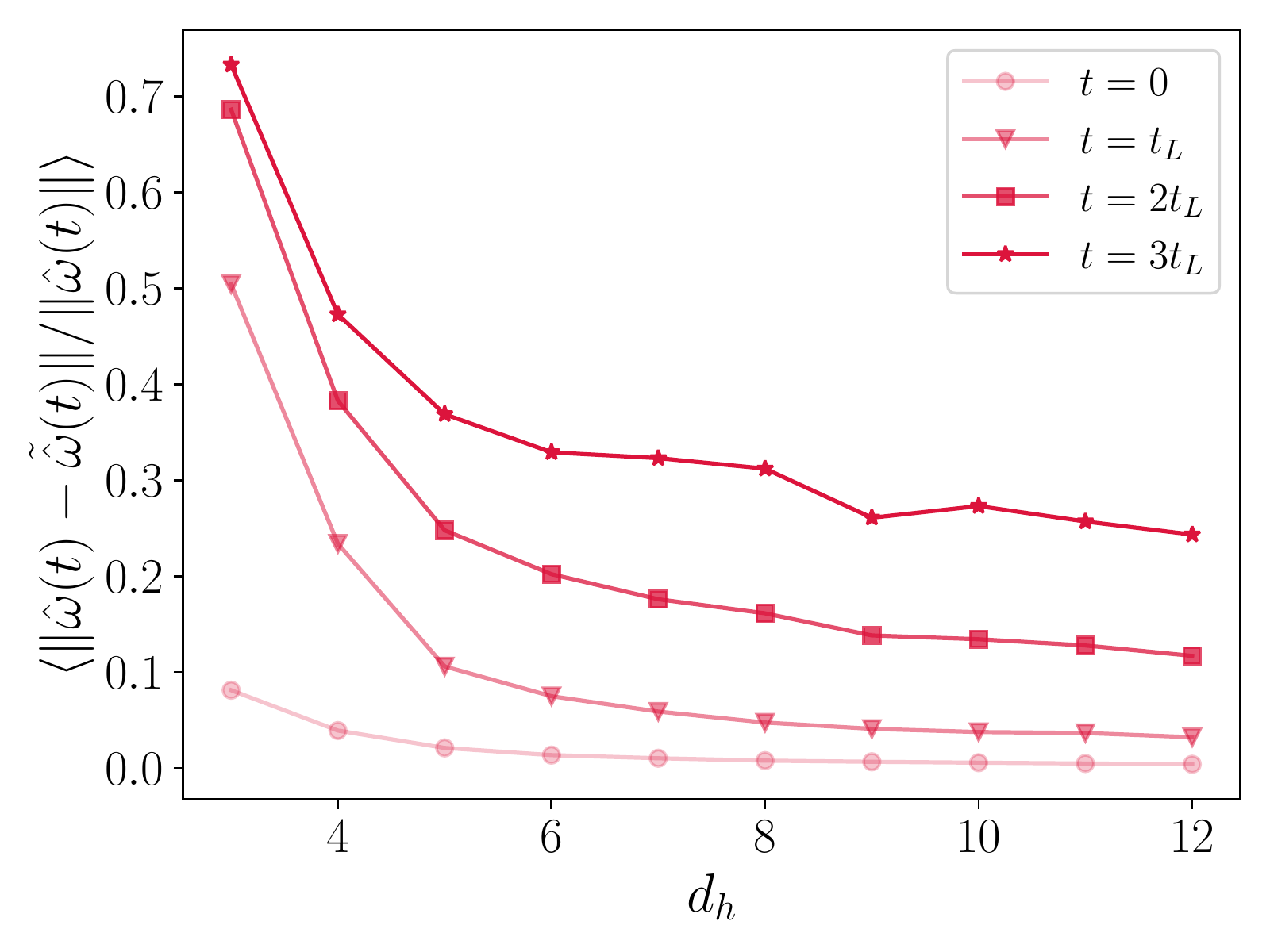}
		\caption{}
		\label{re14d4ICstotal_vsdh}
	\end{subfigure}
	
	\caption{Difference between true vorticity evolution and vorticity evolution obtained from the time map $F$ from $h(t)$ where (a) correspond to averages taken over bursting  and quiescent ICs and (b) averages over all the data. (c) Difference between true vorticity evolution and vorticity evolution obtained from the time map $F$ from $h(t)$ with varying $d_h$ for increasing $t_L$. This corresponds to averages over all the data.}
	\label{totalsplitICs}
	
\end{figure}

Turning from examples of individual trajectories to ensemble averages, Figure \ref{re14d4ICs} shows ensemble averages of the difference between the true and predicted trajectories, separately considering ICs in the bursting and quiescent regions. Solid curves correspond to quiescent ICs and dashed curves to bursting ICs. Starting from $d_h=3$ we increase up to $d_h=12$. We selected $10^4$ ICs in total where approximately 1/3 of the ICs correspond to bursting. As expected, predictions at $d_h=3$ diverge quickly from the true dynamics in both quiescent and bursting IC scenarios. With increasing $d_h$, trajectories track better for both types of ICs. We can also notice that the two darkest curves, corresponding to $d_h=11,12$, fall on top of each other in the case of quiescent ICs and the trajectories for the quiescent ICs track almost perfectly for approximately two Lyapunov times for dimensions $d_h=5$ and higher. In Figure \ref{re14d4ICstotal} we show ensemble averages of the difference between the true and predicted dynamics based on all ICs. The same trend is obtained as discussed for Figure \ref{re14d4ICs} with dimensions of $d_h=9$ and higher in similar agreement, and as expected the errors increase for all of the curves due to the divergence of the bursting ICs. We can conclude that models of dimensions $d_h=5$ and higher are very good at capturing trajectories in the quiescent regions, which happens through the accurate prediction of the oscillatory behavior of the unstable RPO right before a bursting occurs. Prediction from bursting ICs is harder, due to the complex dynamics involved in this region. We also consider , in Figure \ref{re14d4ICstotal_vsdh} the ensemble averages of the difference between the true and predicted trajectories versus $d_h$ for all ICs with at time instants $t=0,t_L, 2t_L, 3t_L$. As expected, with increasing $t$ the trajectories deviate from the true data. However we notice that for all of the curves the error decreases with increasing $d_h$ and after $d_h=9$ little to no improved performance is observed.

%With increasing $d_h$ trajectories track better for the bursting ICs, however 

\begin{comment}
\begin{figure}%[H]
%\vspace{-8mm}
\centering
\includegraphics[width=.6\linewidth]{figures/hibbur_ICs.pdf}
\caption{Difference between predicted and real vorticity evolution. Green curves correspond to averages taken over bursting ICs and black over hibernating ICs. Lightest curve corresponds to $d_h=3$, $d_h$ is increased until reaching the darkest curve corresponding to $d_h=7$}
\label{re14d4ICs}
\end{figure}
\end{comment}

%, if so a window of future and past snapshots is 

%Based on $KE$ we can split data as hibernating and bursting as seen in Figure ?. To classify as hibernating or bursting we select a cutoff based on $KE$ such that a clear division can be seen between the two different regions. Notice that because of the nature of the intermittency of the problem some bursting events can be classified as hibernating if we only consider the $KE$ cutoff. To circumvent this we also consider the near horizon 

%\hl{use this for the busting prediction section }To do this we label the snapshots as 1 if they're in the bursting region and 0 if hibernating. A cut-off is selected where there's a clear difference between 

\subsection{Time evolution as a function of dimension  - Long time predictions}\label{sec:dimredev}

In this section we present long time statistics for the models and  true data at $\Rey=14.4$.  From ICs on the attractor, we evolve for $2 \times 10^5$ time units, yielding to get $4 \times 10^4$ snapshots of data. This duration is sufficient to densely sample the quiescent and bursting regions.  We note that long time statistics did not change if the IC was in a bursting or quiescent region.

% dont like the transition to thi part

%Recall that the HNN results pointed us towards the dimension of the manifold $\mathcal{M}$ for $\operatorname{Re} = 13.5$, and the time map NN validated this result by obtaining near perfect prediction of the flow statistics when latent space is evolved in time(\hl{maybe include pdf difference plot}). 
%By calculating the $I(t)-D(t)$ joint PDF we notice that at a dimension of $d_h=4$, Figure \ref{fig:subipd4}, the PDF takes a shape that looks similar to the true data. 

\begin{figure}%[H]
	%\vspace{-8mm}
	%\centering
	\begin{subfigure}{0.45\textwidth}
	%\centering
	\includegraphics[width=0.85\linewidth]{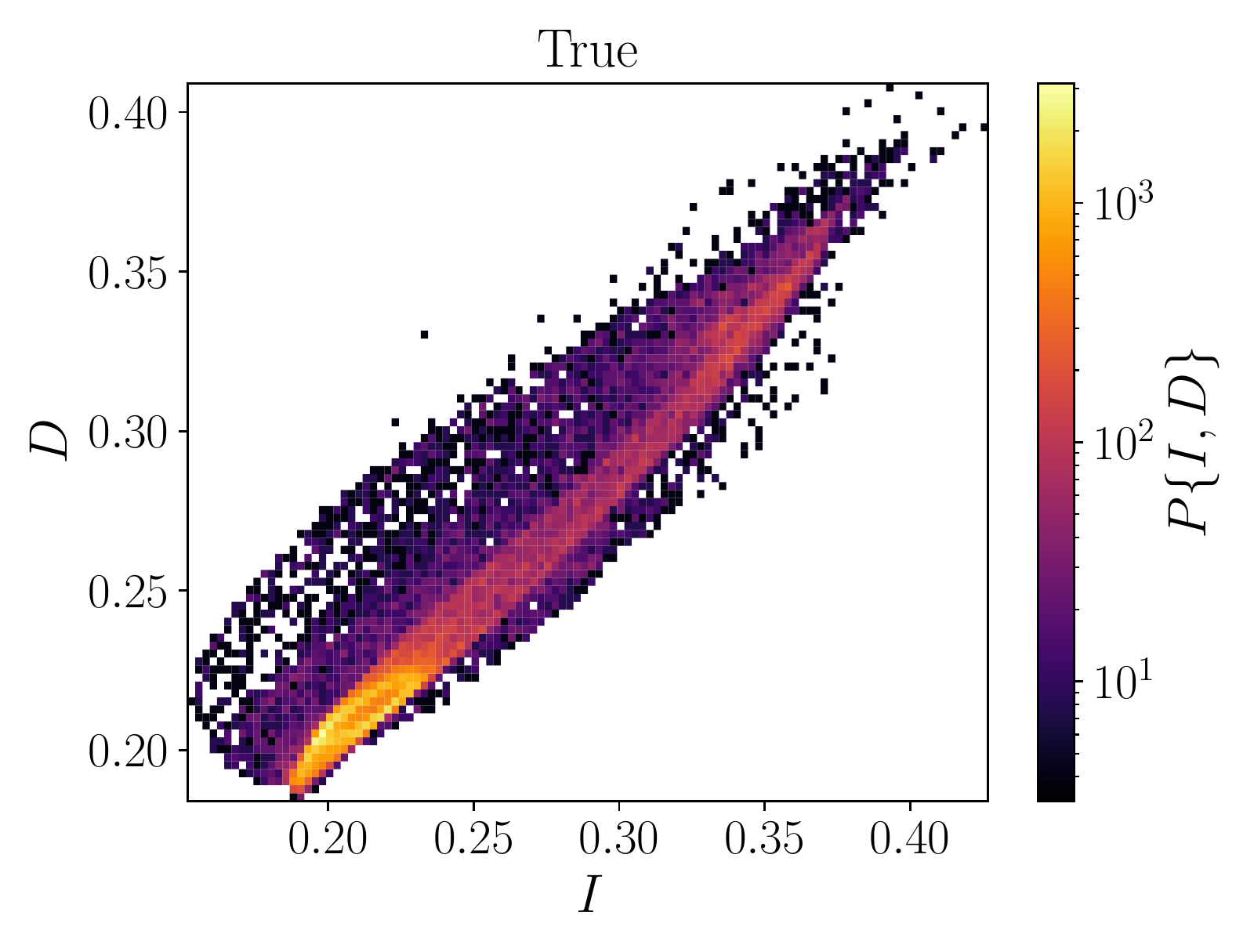}
	\caption{}
	\label{fig:subipdtrue}
	\end{subfigure}%
	\begin{subfigure}{0.45\textwidth}
	%\centering
	\includegraphics[width=.85\linewidth]{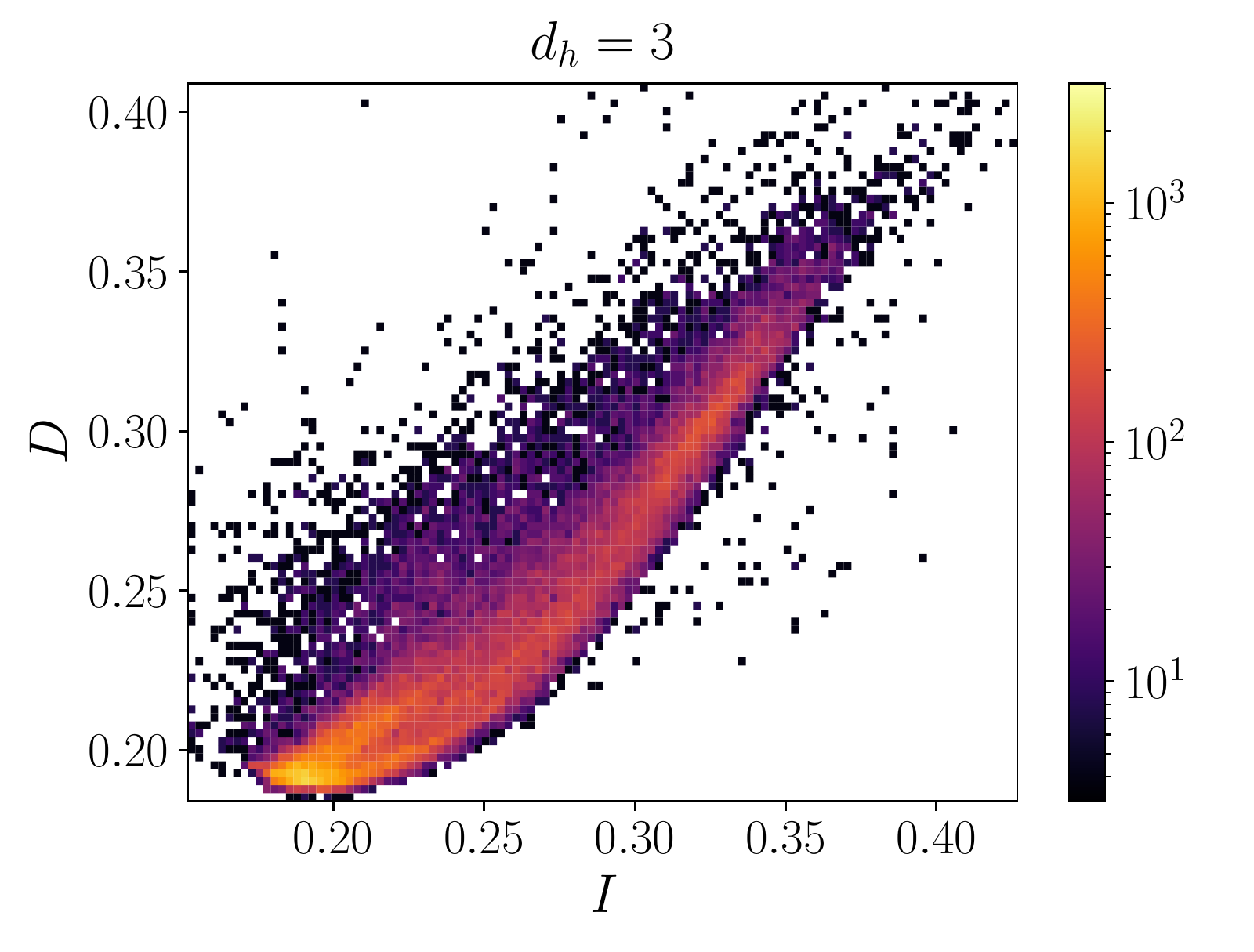}
	\caption{}
	\label{fig:subipd3}
	\end{subfigure}
	\begin{subfigure}{0.45\textwidth}
	%\centering
	\includegraphics[width=.85\linewidth]{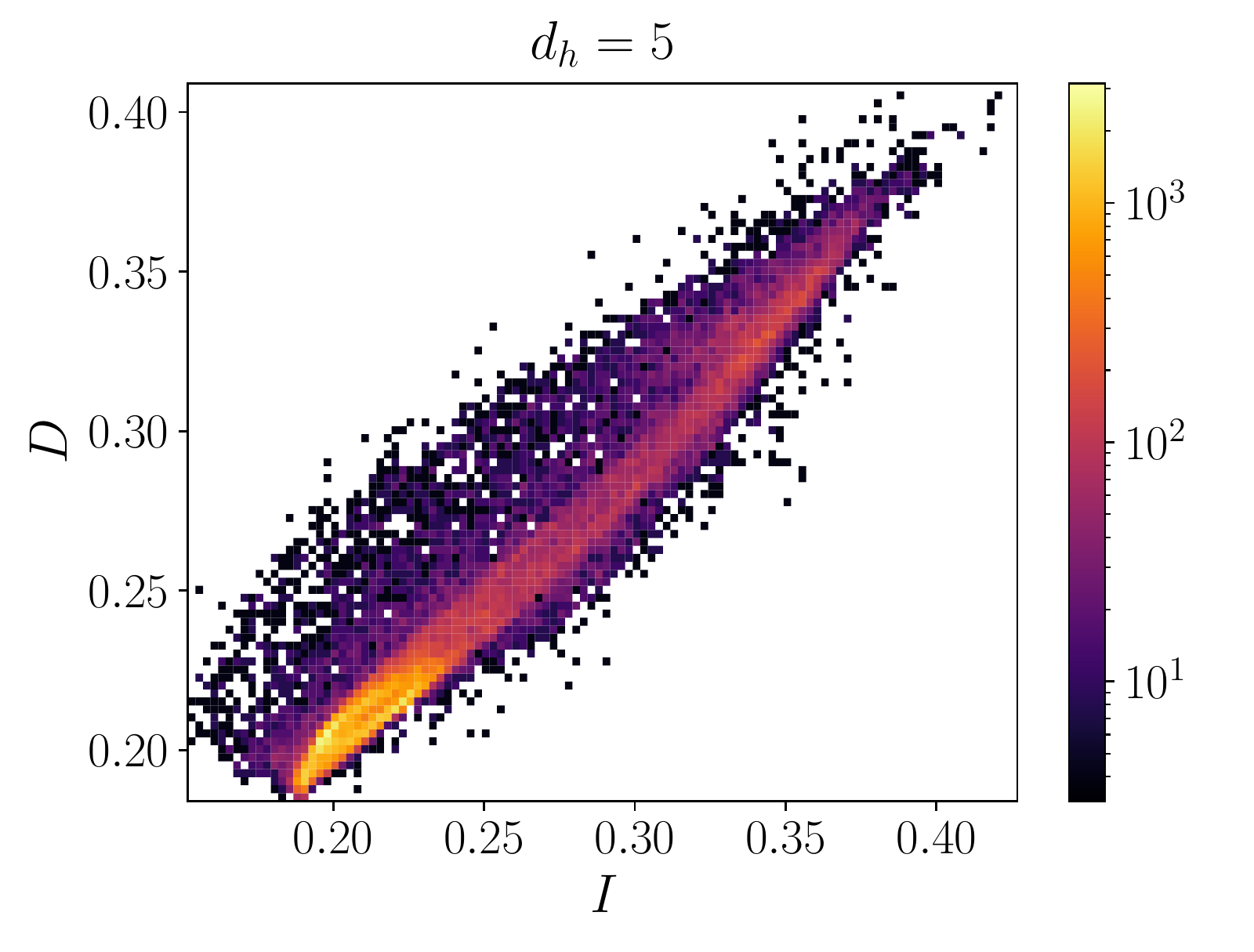}
	\caption{}
	\label{fig:subipd4}
	\end{subfigure}
	\begin{subfigure}{0.45\textwidth}
	%\centering
	\includegraphics[width=.85\linewidth]{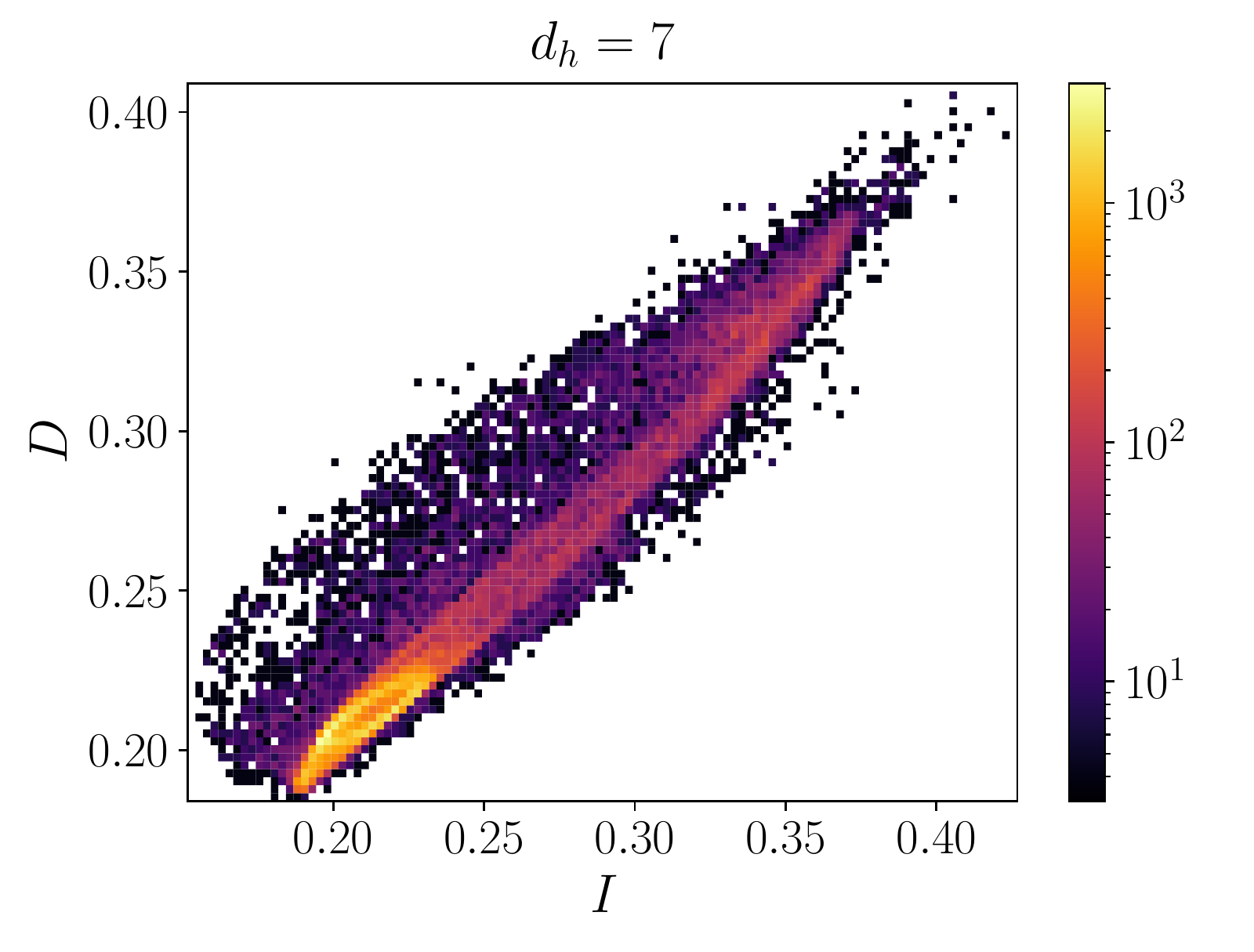}
	\caption{}
	\label{fig:subipd5}
	\end{subfigure}
	\begin{subfigure}{0.45\textwidth}
	%\centering
	\includegraphics[width=.85\linewidth]{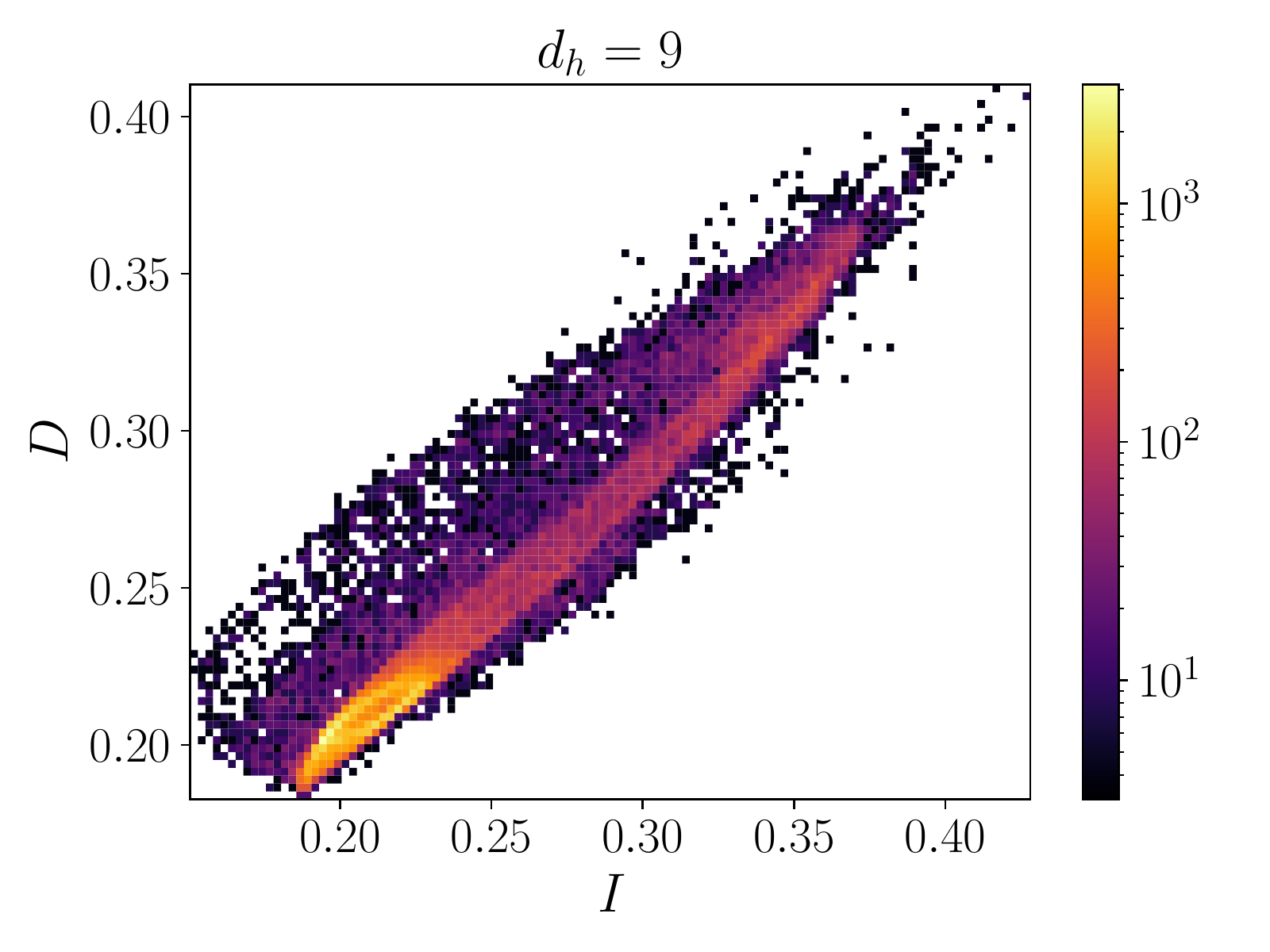}
	\caption{}
	\label{fig:subipd6}
\end{subfigure}
	\begin{subfigure}{0.45\textwidth}
	%\centering
	\includegraphics[width=.85\linewidth]{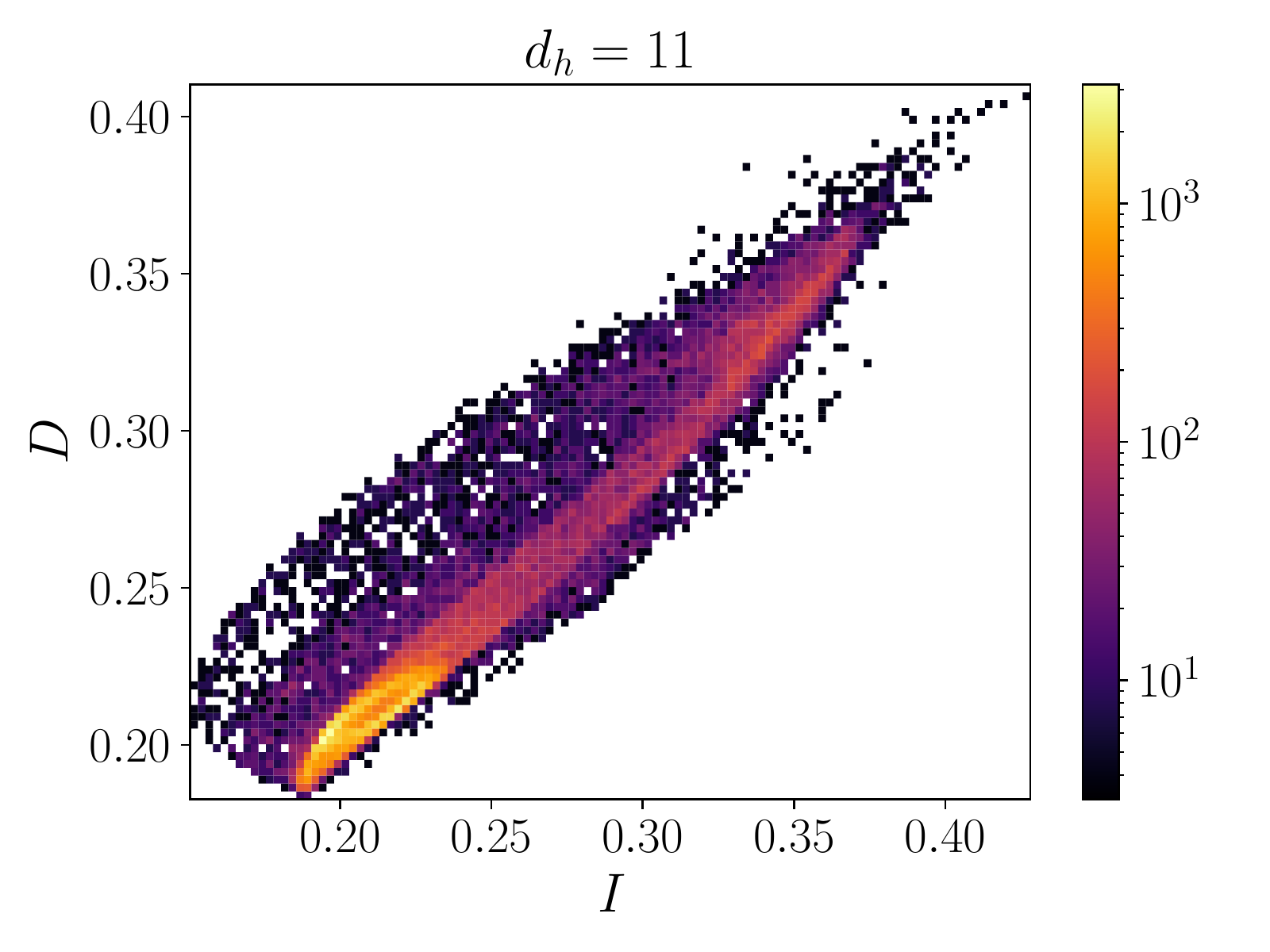}
	\caption{}
	\label{fig:subipd7}
\end{subfigure}
	\caption{$\operatorname{Re}=14.4$: Joint PDFs of $I$-$D$ corresponding to $\operatorname{Re}=14.4$ for (a) true and  (b)-(f) predicted data  corresponding to dimensions $d_h=3,5,7,9,$ and $11$.}
	\label{re14d4PD}	
\end{figure}

\begin{figure}%[H]
	%\vspace{-8mm}
	%\centering
	\begin{subfigure}{0.45\textwidth}
		%\centering
		\includegraphics[width=0.85\linewidth]{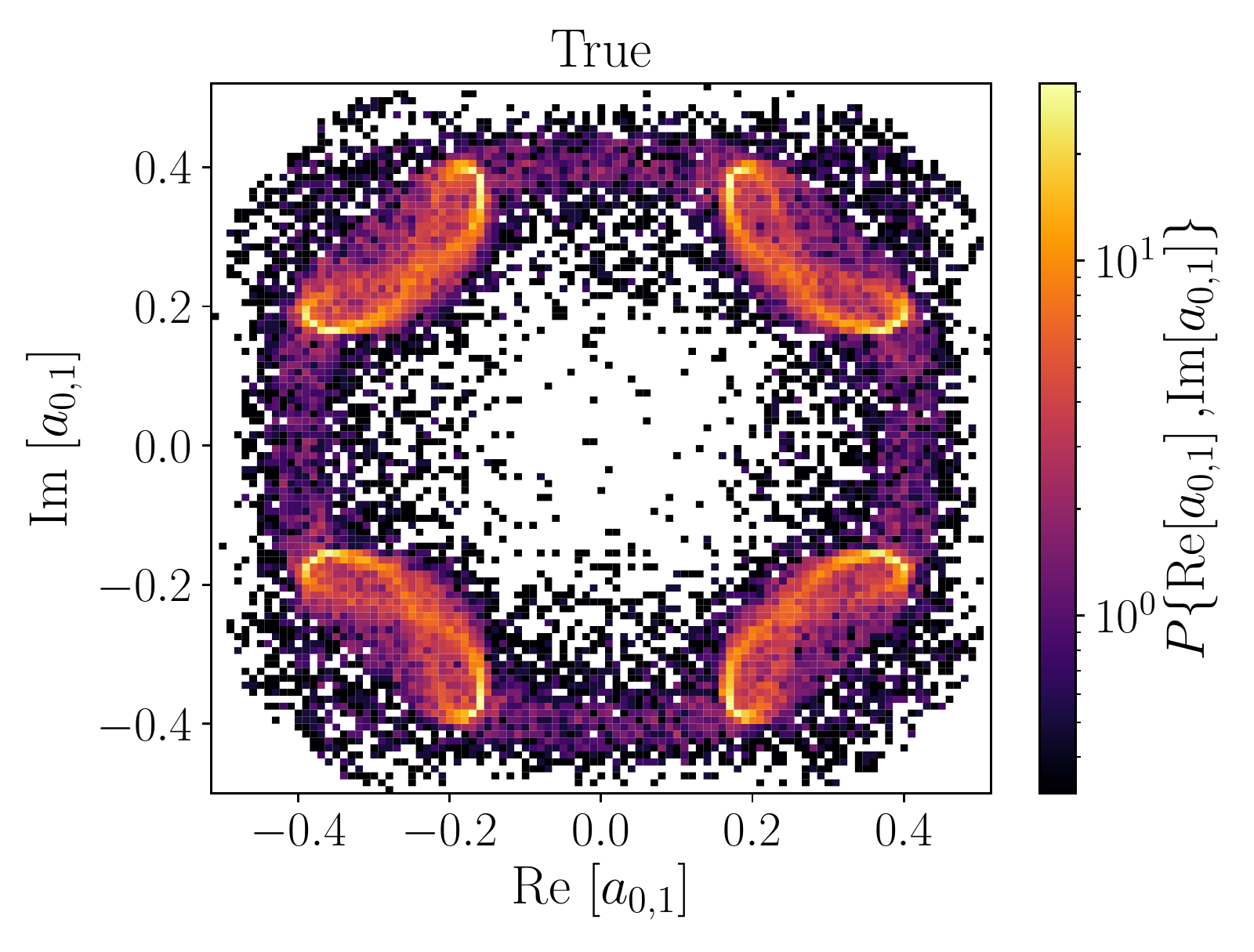}
		\caption{}
		\label{fig:sub01dtrue}
	\end{subfigure}%
	\begin{subfigure}{0.45\textwidth}
	%\centering
	\includegraphics[width=.85\linewidth]{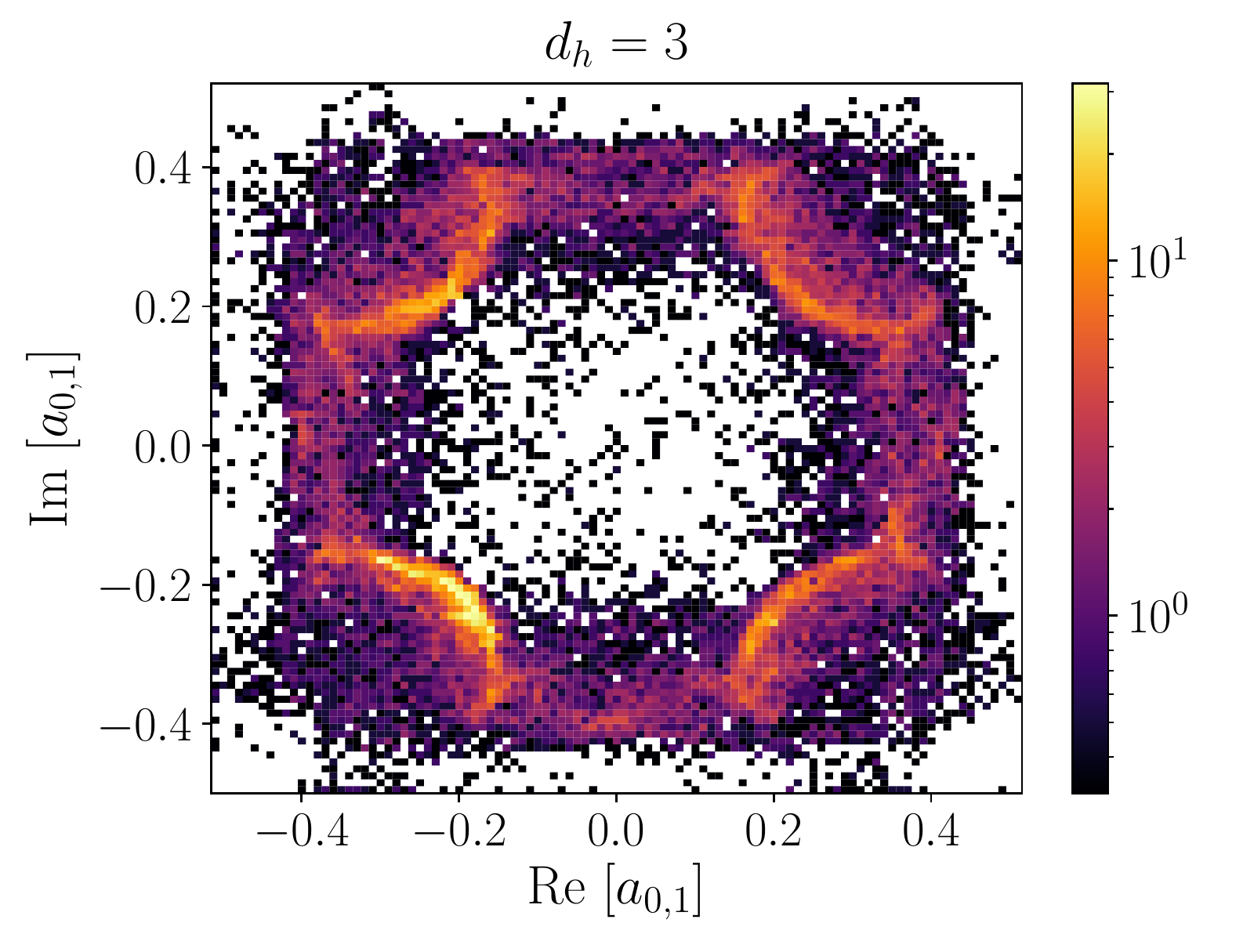}
	\caption{}
	\label{fig:sub01d3}
	\end{subfigure}
	\begin{subfigure}{0.45\textwidth}
	%\centering
	\includegraphics[width=.85\linewidth]{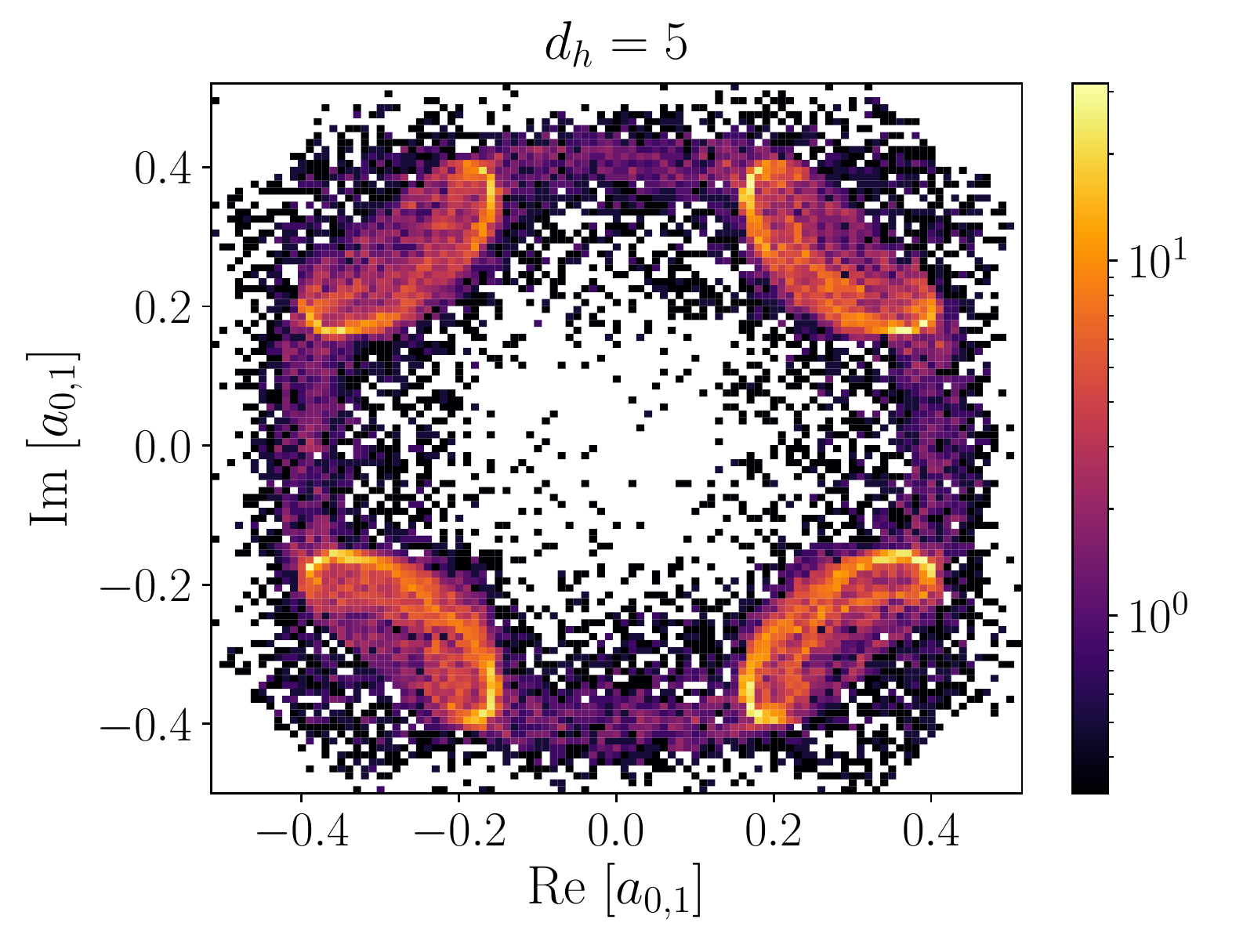}
	\caption{}
	\label{fig:sub01d4}
	\end{subfigure}
	\begin{subfigure}{0.45\textwidth}
		%\centering
		\includegraphics[width=.85\linewidth]{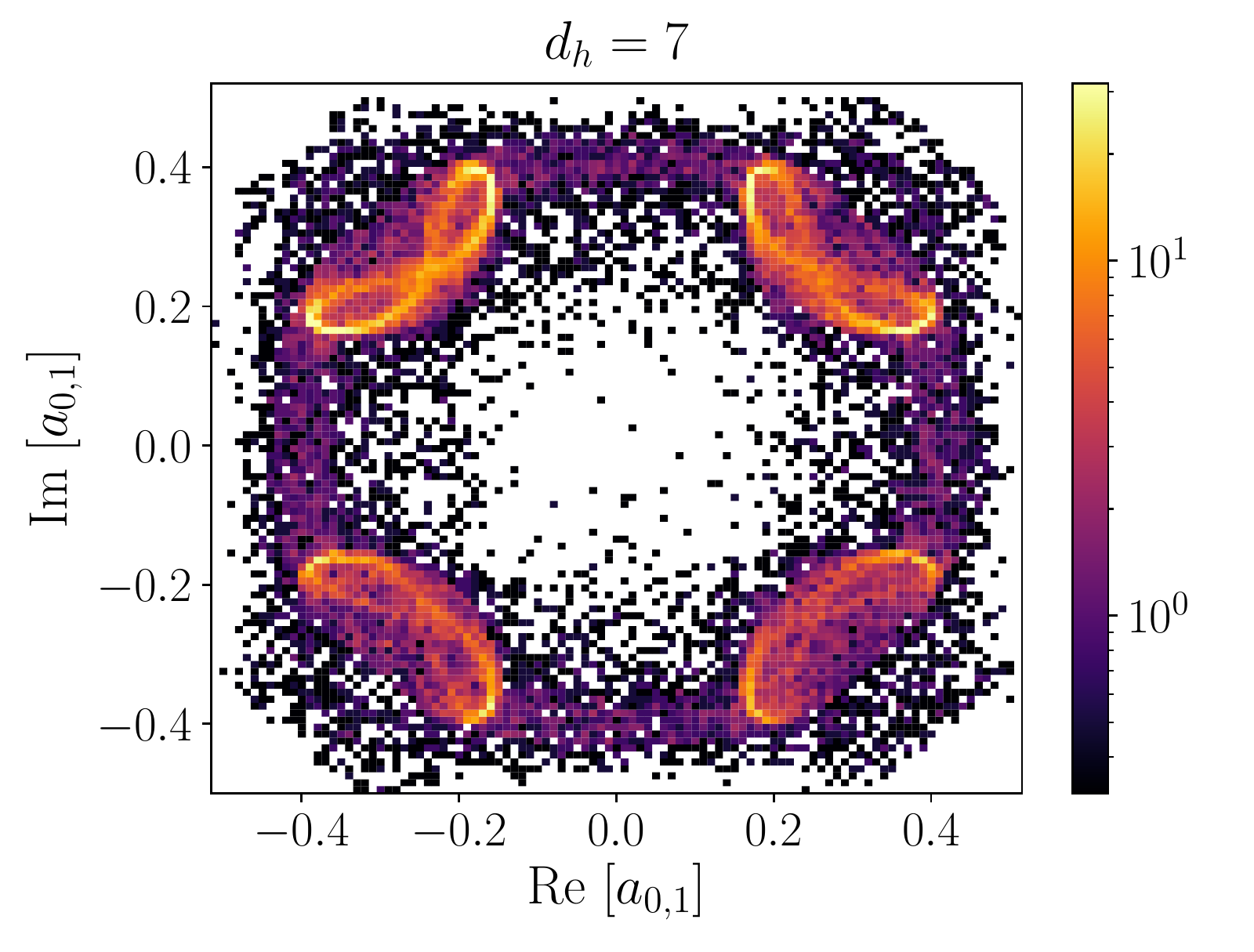}
		\caption{}
		\label{fig:sub01d5}
	\end{subfigure}
	\begin{subfigure}{0.45\textwidth}
	%\centering
	\includegraphics[width=.85\linewidth]{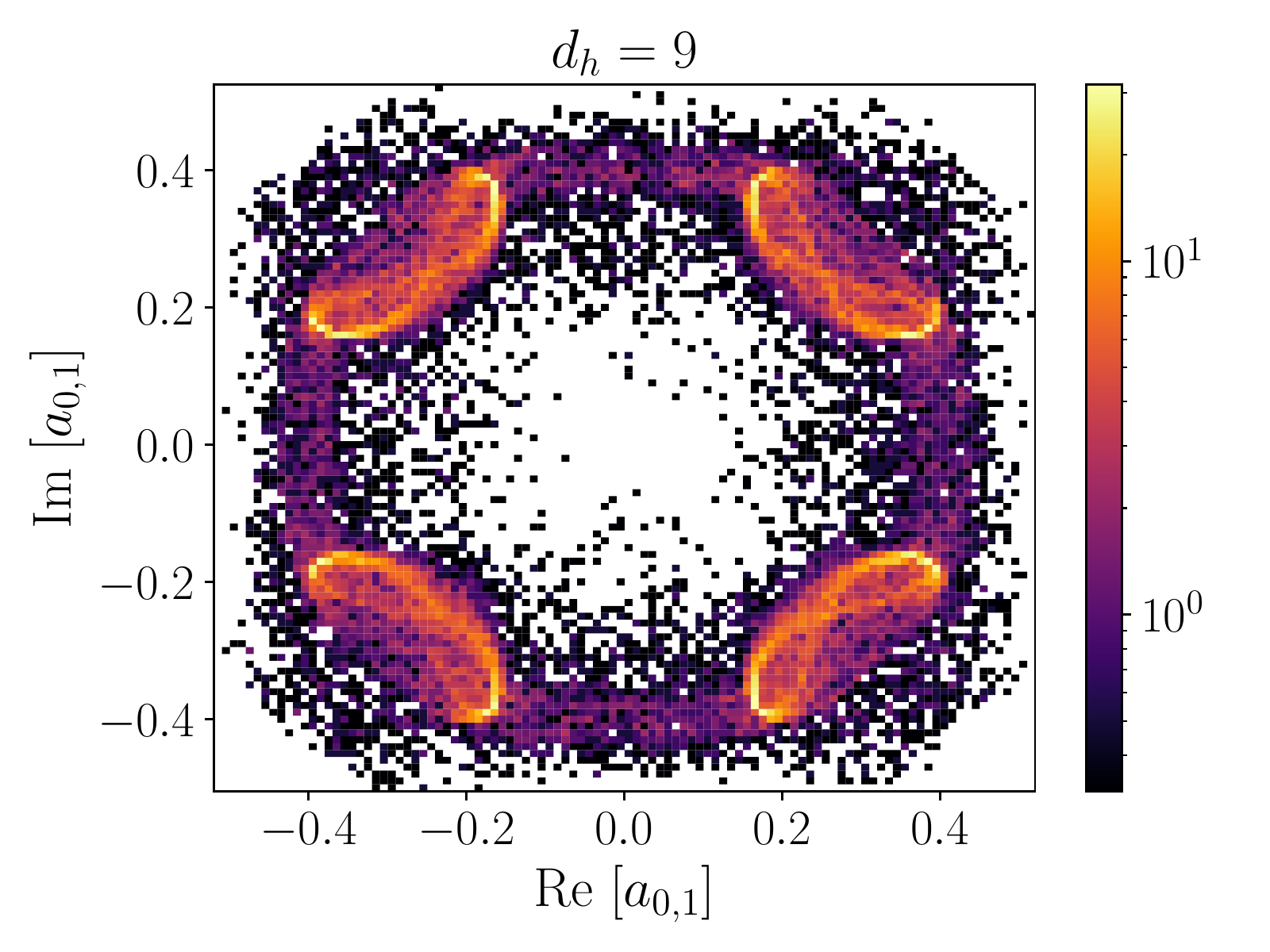}
	\caption{}
	\label{fig:sub01d6}
\end{subfigure}
	\begin{subfigure}{0.45\textwidth}
	%\centering
	\includegraphics[width=.85\linewidth]{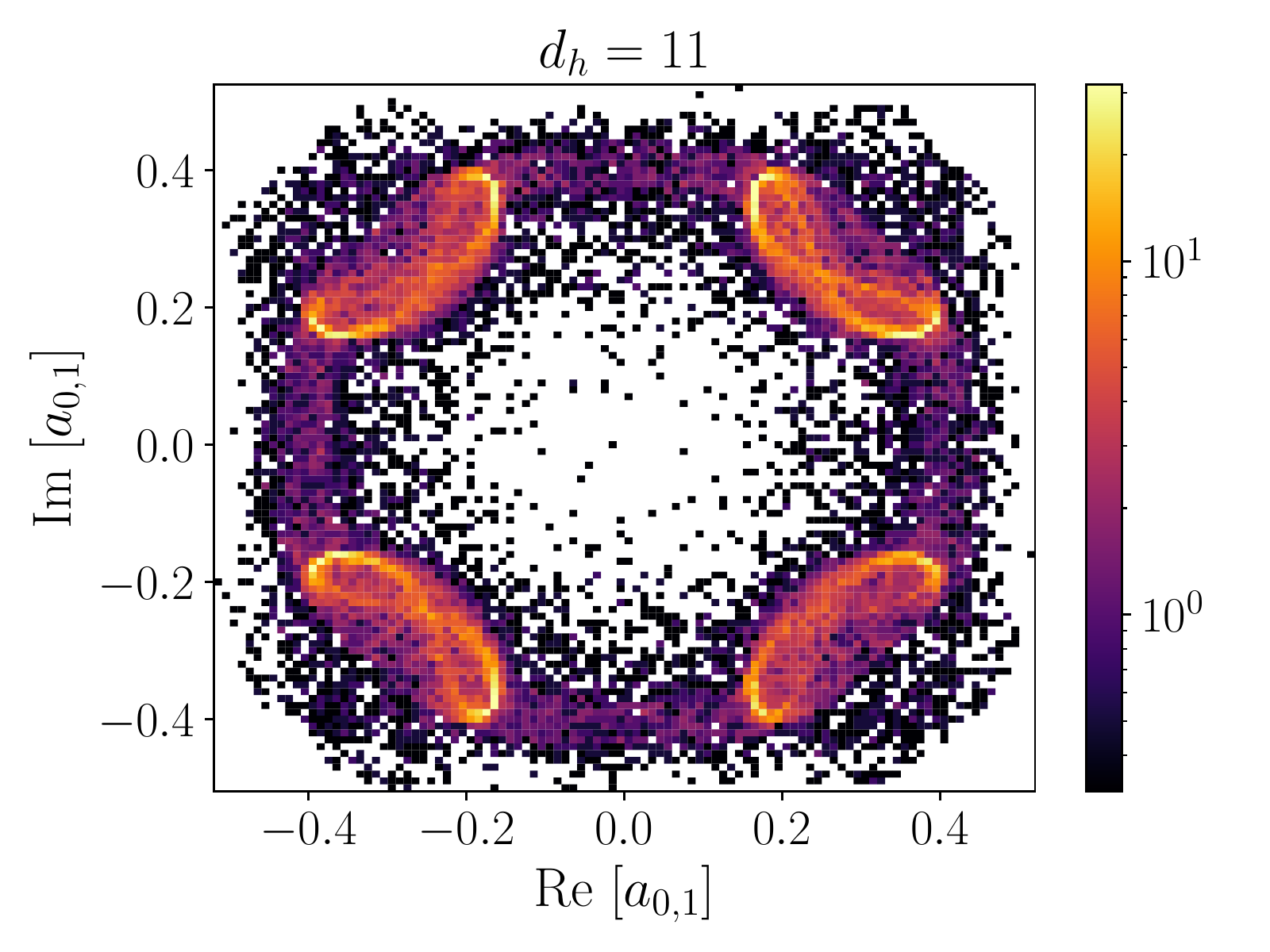}
	\caption{}
	\label{fig:sub01d7}
\end{subfigure}
	
	\caption{$\operatorname{Re}=14.4$: Joint PDFs of $\operatorname{Re}\left[a_{0,1}(t)\right]-\operatorname{Im}\left[a_{0,1}(t)\right]$ corresponding to $\operatorname{Re}=14.4$ for (a) true and  (b)-(f) predicted data  corresponding to dimensions $d_h=3,5,7,9,$ and $11$.}
	\label{re14d4PD01}
	
\end{figure}

\begin{figure}%[H]
	%\vspace{-8mm}
	%\centering
	\begin{subfigure}{0.45\textwidth}
		%\centering
		\includegraphics[width=1\linewidth]{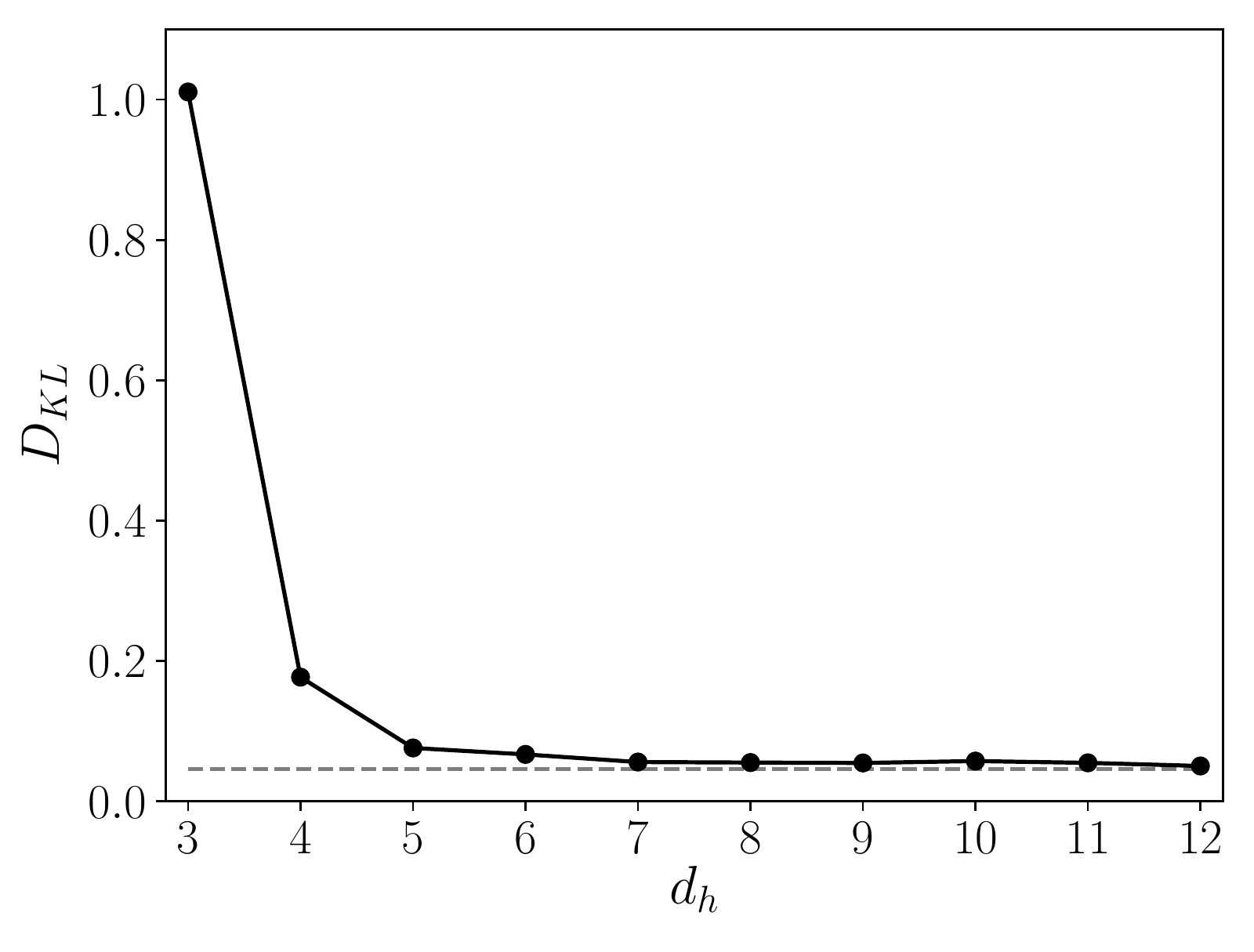} % Figure in KLDiv other bins
		\caption{}
		\label{fig:subKLIPD}
	\end{subfigure}%
	\begin{subfigure}{0.45\textwidth}
		%\centering
		\includegraphics[width=1\linewidth]{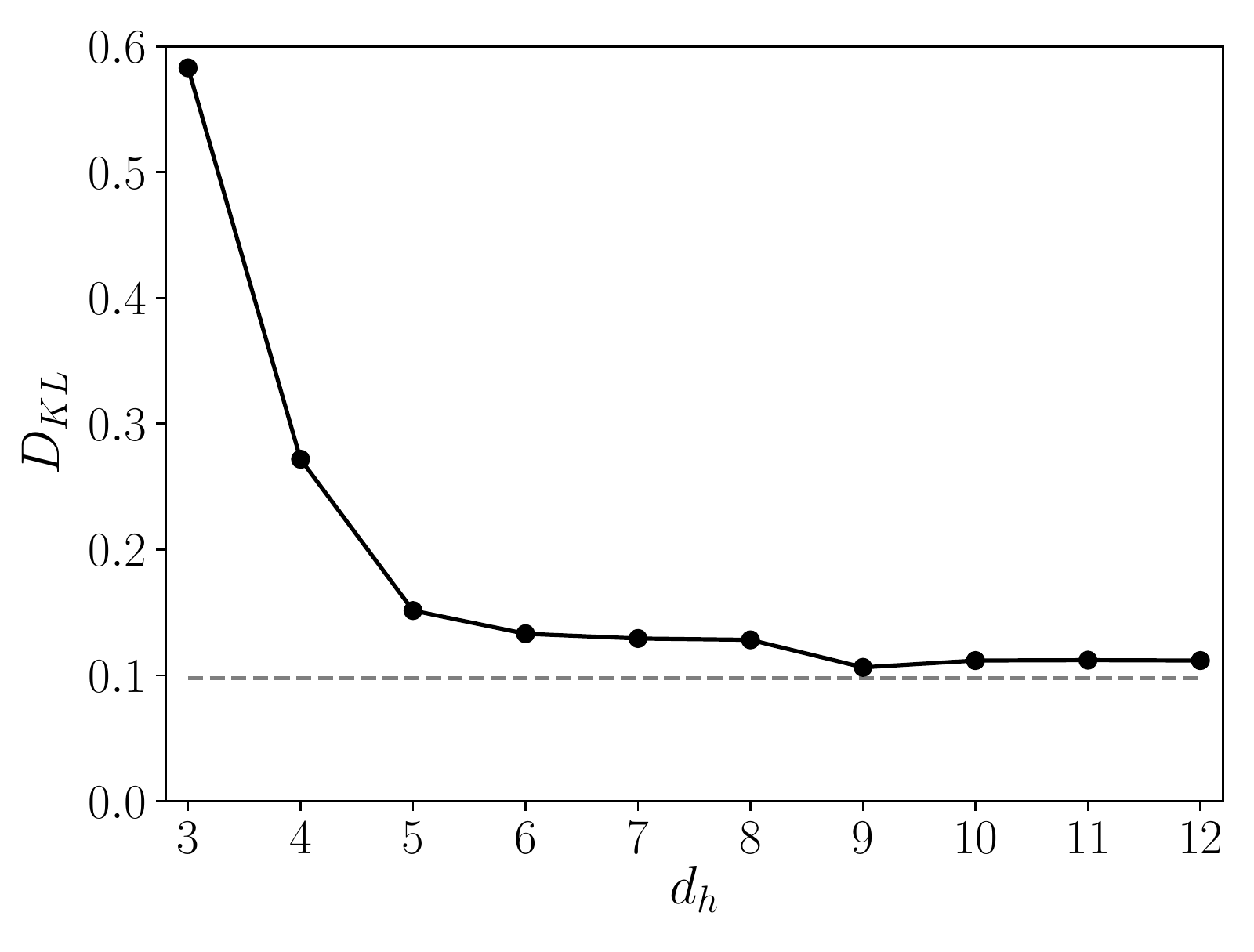}
		\caption{}
		\label{fig:subKLF01}
	\end{subfigure}
	
	\caption{$\operatorname{Re}=14.4$: $D_{KL}$ vs dimension $d_h$ for (a) $I$-$D$ and (b) $\operatorname{Re}\left[a_{0,1}\right]-\operatorname{Im}\left[a_{0,1}\right]$ predicted vs true joint PDFs. Dashed grey line corresponds to $D_{KL}$ calculated over true data sets.}
	\label{re14d4KLDiv}
	
\end{figure}

Figure \ref{re14d4PD} shows the joint probability density function (PDF) of $I$ and $D$ for true and predicted data from models with $d_h=3,5,7,9,$ and $11$ -- note the logarithmic scale, here and below. We notice that at $d_h=3$ the different areas corresponding to quiescent and bursting regions are populated similarly in terms of the probability intensity compared with the true PDF shown, but the shape of the predicted PDF takes a curved form that is not seen in the true PDF. When we get to $d_h=5$ the $D$ and $I$ events are captured better, and similarly for increasing dimensions. We also compute the joint PDF of $\operatorname{Re}\left[a_{0,1}\right]$ and $\operatorname{Im}\left[a_{0,1}\right]$, shown in Figure \ref{re14d4PD01}. From this quantity we can observe the heteroclinic-like connections between the unstable RPOs, which correspond to the four ribbon-like regions of high probability. Here we see similar trends as in the joint PDF for $I$-$D$: $d_h=3$ shows poor qualitative reconstruction compared with higher dimensions, and once $d_h\geq 5$, the joint PDFs from the model prediction are virtually indistinguishable from the true PDFs. To further quantify the relationship of  the PDFs from the models to the true data, we calculate the Kullback-Leibler (KL) divergence,  
\begin{equation}
D_{KL}(\tilde{P}||P)=\int_{-\infty}^\infty \int_{-\infty}^\infty \tilde{P}\{a,b\} \text{ln}\dfrac{\tilde{P}\{a,b\} }{P\{a,b\}}da \; db,
\end{equation}
where $\tilde{P}$ corresponds to the predicted PDF and $P$ to the true PDF. Due to the approximation of the integral to discrete data we ignore areas where either the true or predicted PDFs are zero. Let us first consider the case $a=I$ and $b=D$. Figure \ref{fig:subKLIPD} shows $D_{KL}$ calculated with varying $d_h$. The dashed grey line corresponds to $D_{KL}$ calculated over different true data sets. This serves as a baseline for comparison to the predicted PDFs. A significant decrease happens at $d_h=4$ followed by small decreases at higher dimensions. We see that after $d_h=5$ no significant information is gained, with errors plateauing at approximately $d_h \geq 7$. We can also look at the case where $a= \;$Re $\left[a_{0,1}\right]$ and $b= \;$Im $\left[a_{0,1}\right]$ in Figure \ref{fig:subKLF01}. We notice that errors of the joint PDF in Figure \ref{fig:subKLF01} show a similar trend as Figure \ref{fig:subKLIPD} with errors plateauing at approximately $d_h \geq 9$. We can infer from these results that the embedding dimension of this system lies in the range $d_h=5-9$, and furthermore that the data-driven model can reproduce the long-time statistics with very high fidelity.  

%We will show in further sections that $d_h=5$ does a great job and 

%When considering $\operatorname{Re}=20$ we see that a good reconstruction of the joint PDFs happens at $d_h=10$ as seen in Figure \ref{re20PD}. $D_{KL}$ here seems to stop decreasing at $d_h=10$, which is where we see that the MSE in Figure \ref{fig:subMSEre20} stops decreasing. 

 %However $d_h=5$ shows better qualitative behavior with respect to the true data. Looking at the joint PDF difference in Figure \ref{re14d4PD01error} we can see that at $d_h=5$ is where the MSE plateaus.

%at a dimmension of $d_h=5$ 

%We notice that 

%We now pose the question 

%together with the time stepper NN 
%We now consider the weakly chaotic case of $Re = 14.4$

%\begin{figure}[H]
%	%\vspace{-8mm}
%	\centering
%	\includegraphics[width=0.8\linewidth]{figures/re14d4PD2.pdf}
%	\caption{Joint PDFs corresponding to $Re=14.4$ for true and predicted data of power input and dissipation}
%	\label{re14d4PD}
%\end{figure}

\begin{comment}
\begin{figure}[H]
	%\vspace{-8mm}
	\centering
	\includegraphics[width=0.6\linewidth]{figures/ipdpdferror.pdf}
	\caption{Joint PDFs difference between true and predicted data corresponding to $Re=14.4$ of power input and dissipation.}
	\label{re14d4pdferror}
\end{figure}
\end{comment}

\begin{figure}%[H]
	%\vspace{-8mm}
	%\centering
	\begin{subfigure}{0.45\textwidth}
		%\centering
		\includegraphics[width=0.85\linewidth]{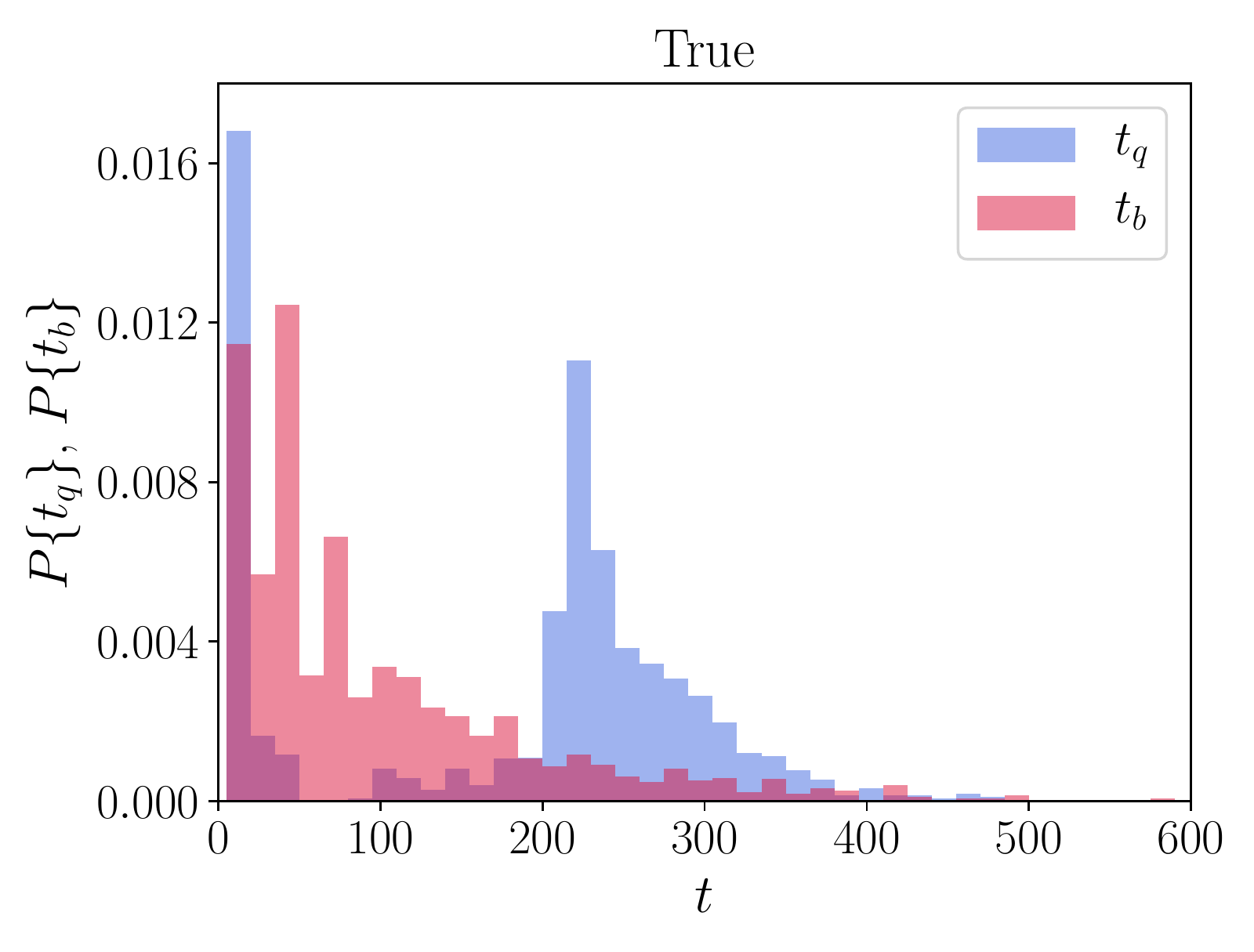} %these are in histcalc.py
		\caption{}
		\label{fig:subhibburtrue}
	\end{subfigure}%
	\begin{subfigure}{0.45\textwidth}
		%\centering
		\includegraphics[width=.85\linewidth]{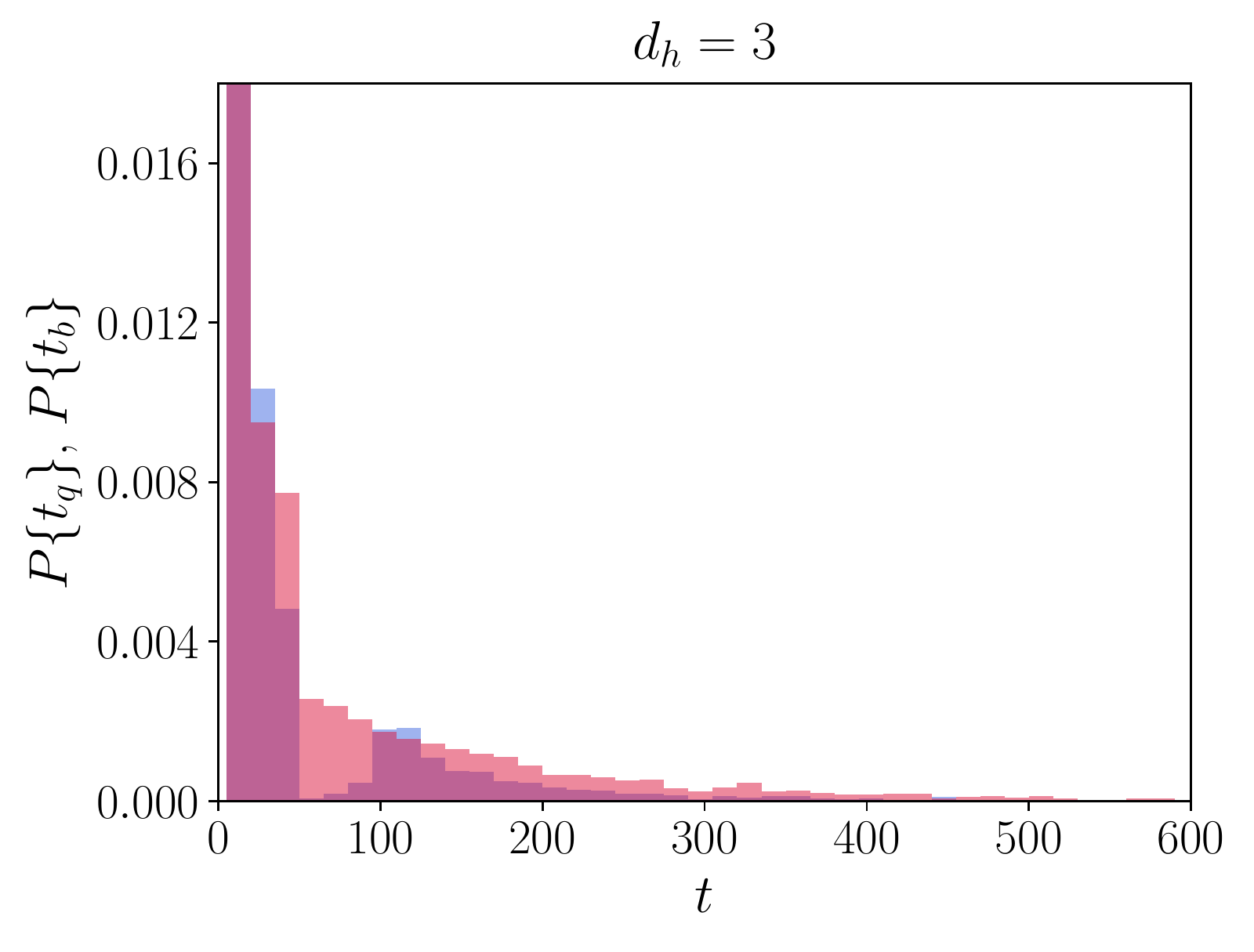}
		\caption{}
		\label{fig:subhibburd3}
	\end{subfigure}
	\begin{subfigure}{0.45\textwidth}
		%\centering
		\includegraphics[width=.85\linewidth]{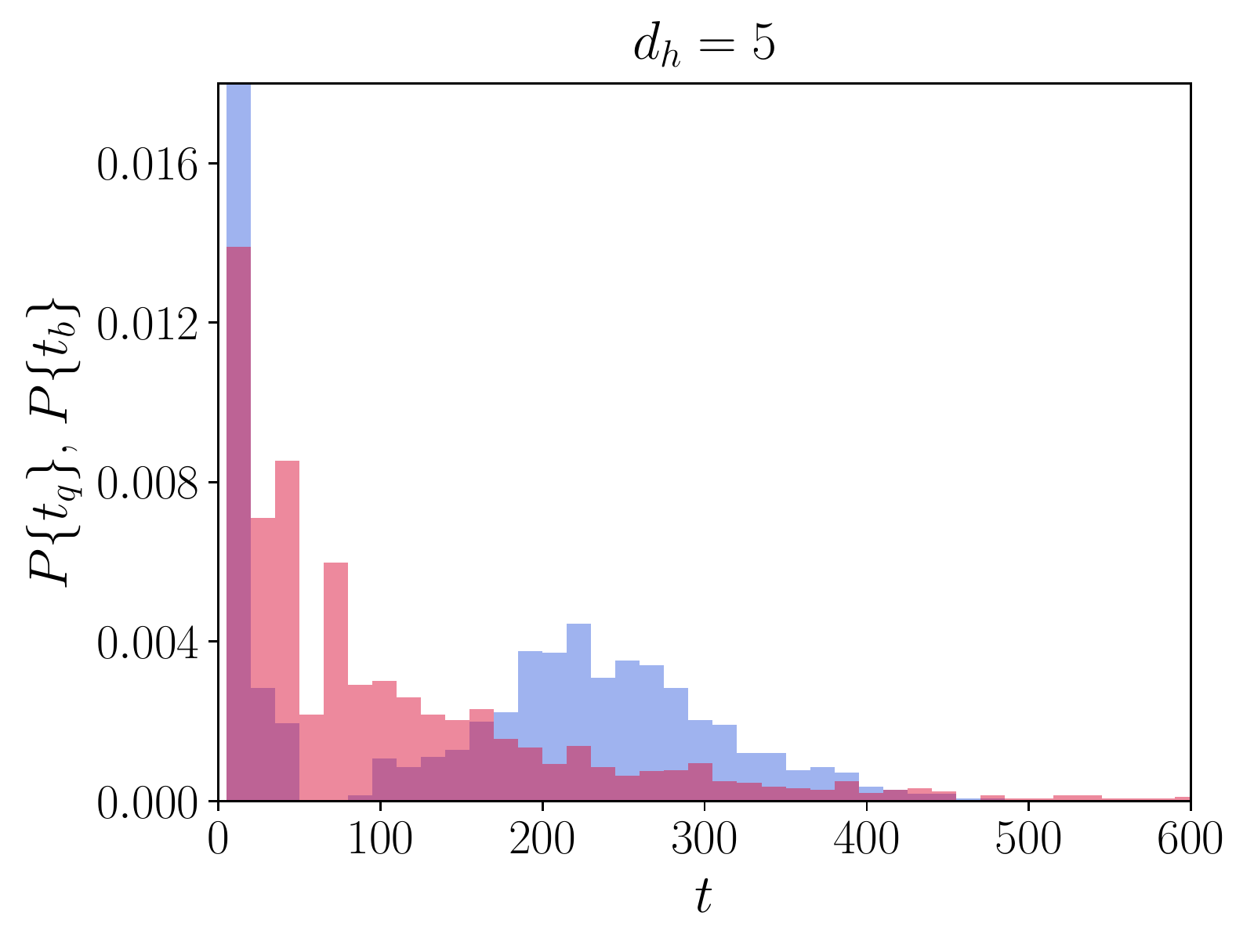}
		\caption{}
		\label{fig:subhibburd4}
	\end{subfigure}
	\begin{subfigure}{0.45\textwidth}
		%\centering
		\includegraphics[width=.85\linewidth]{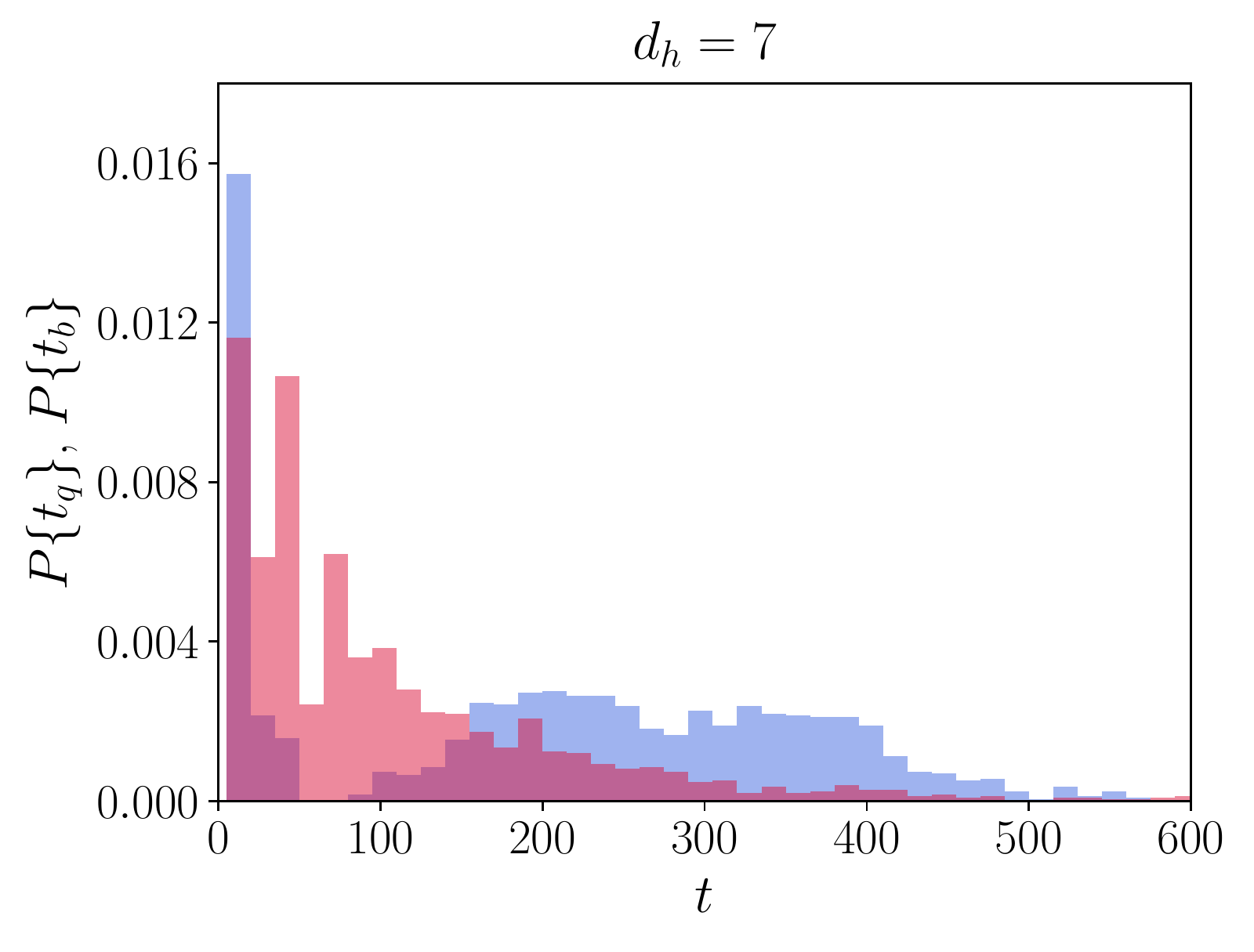}
		\caption{}
		\label{fig:subhibburd5}
	\end{subfigure}
	\begin{subfigure}{0.45\textwidth}
		%\centering
		\includegraphics[width=.85\linewidth]{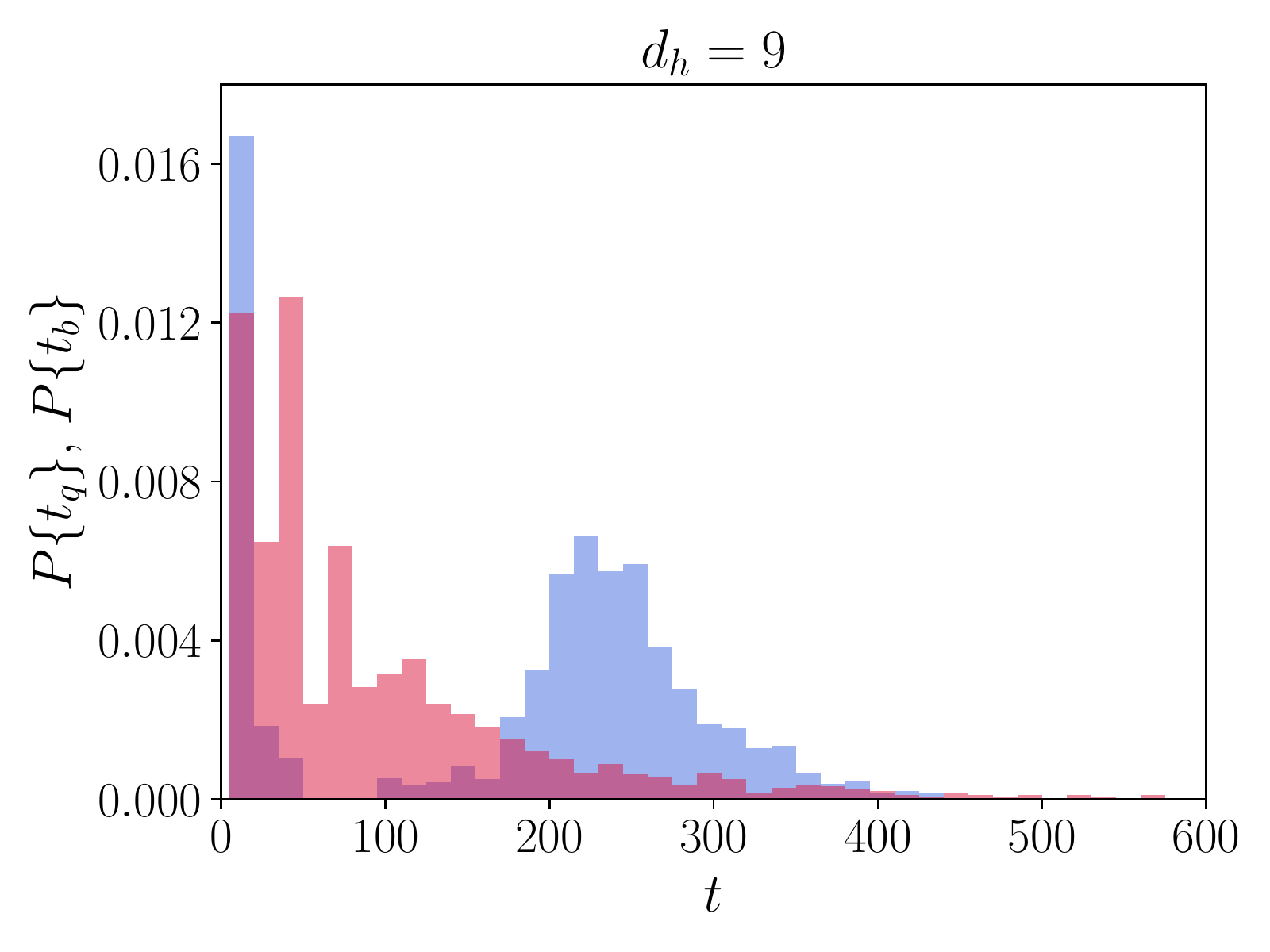}
		\caption{}
		\label{fig:subhibburd6}
	\end{subfigure}
	\begin{subfigure}{0.45\textwidth}
		%\centering
		\includegraphics[width=.85\linewidth]{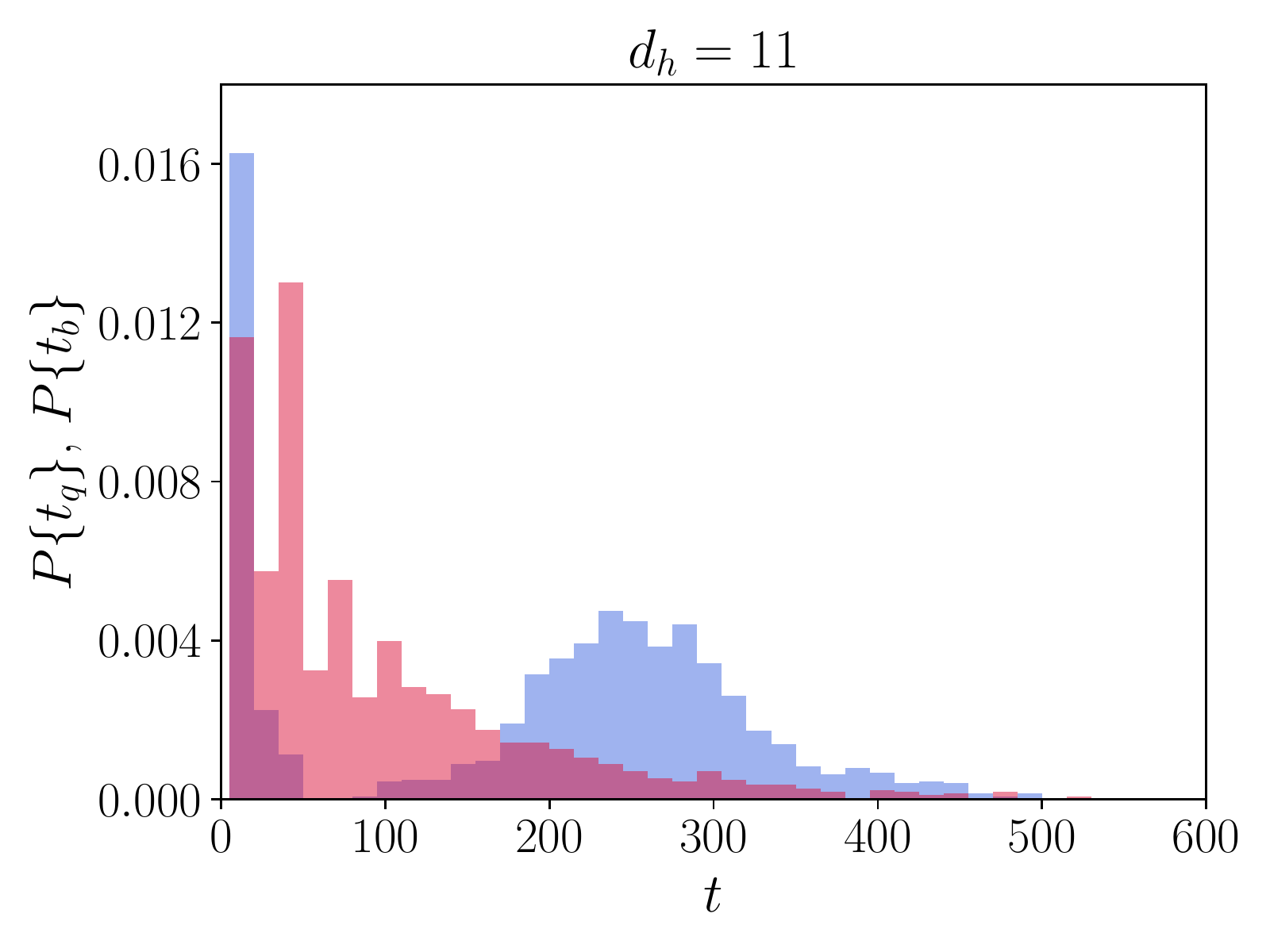}
		\caption{}
		\label{fig:subhibburd7}
	\end{subfigure}
	
	\caption{PDFs of $t_q$ and $t_b$ at $\operatorname{Re}=14.4$ for (a) true and  (b)-(f) predicted data  for dimensions $d_h=3,5,7,9,$ and $11$.}
	\label{re14d4PDhibbur}
	
\end{figure}

\begin{figure}%[H]
	%\vspace{-8mm}
	%\centering
	\begin{subfigure}{0.45\textwidth}
		%\centering
		\includegraphics[width=1\linewidth]{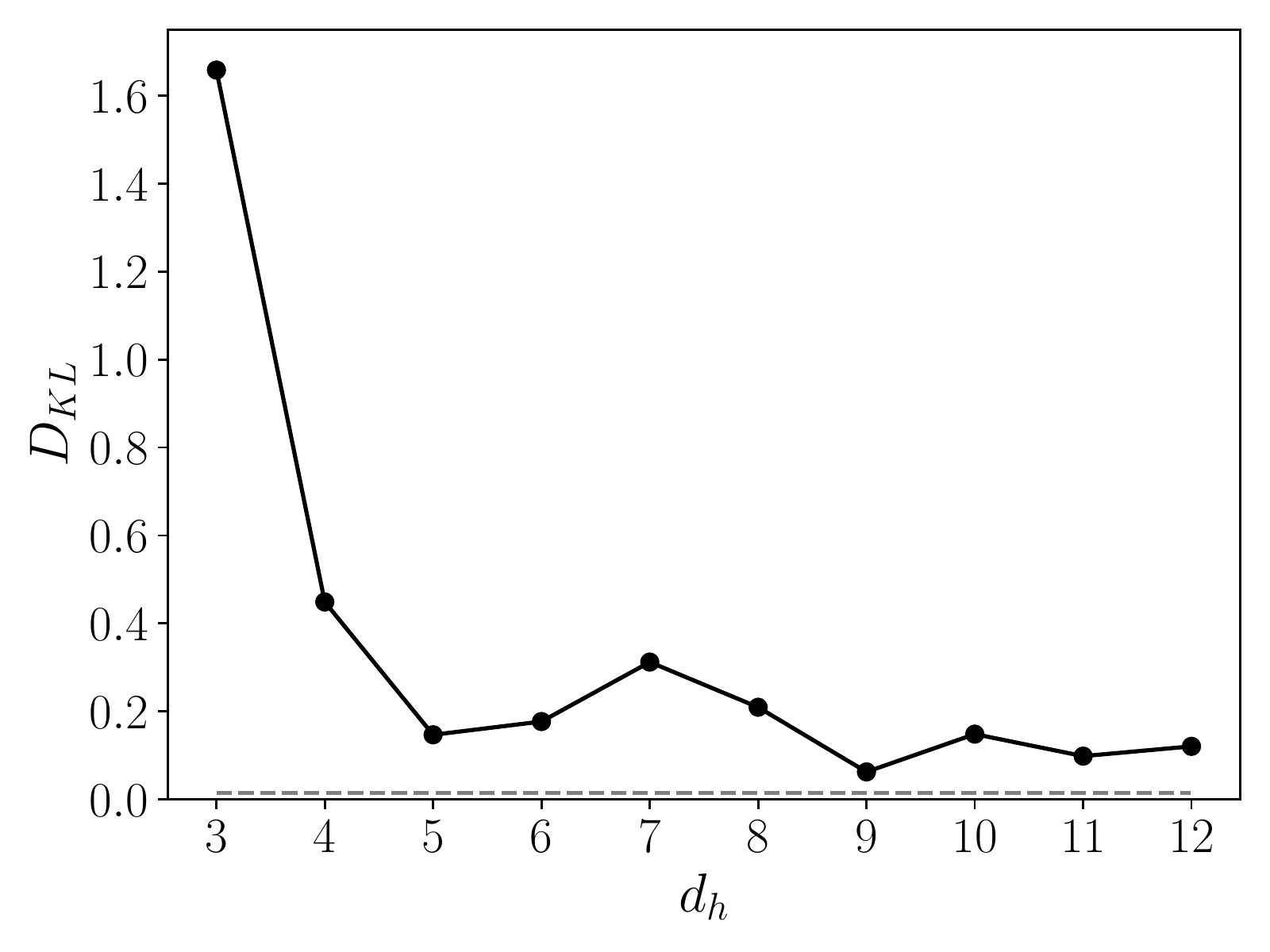}%this plot is from file manyICsProcess_hib_bur_loadindiv in bursting and hibernating folder
		\caption{}
		\label{fig:subtq}
	\end{subfigure}%
	\begin{subfigure}{0.45\textwidth}
		%\centering
		\includegraphics[width=1\linewidth]{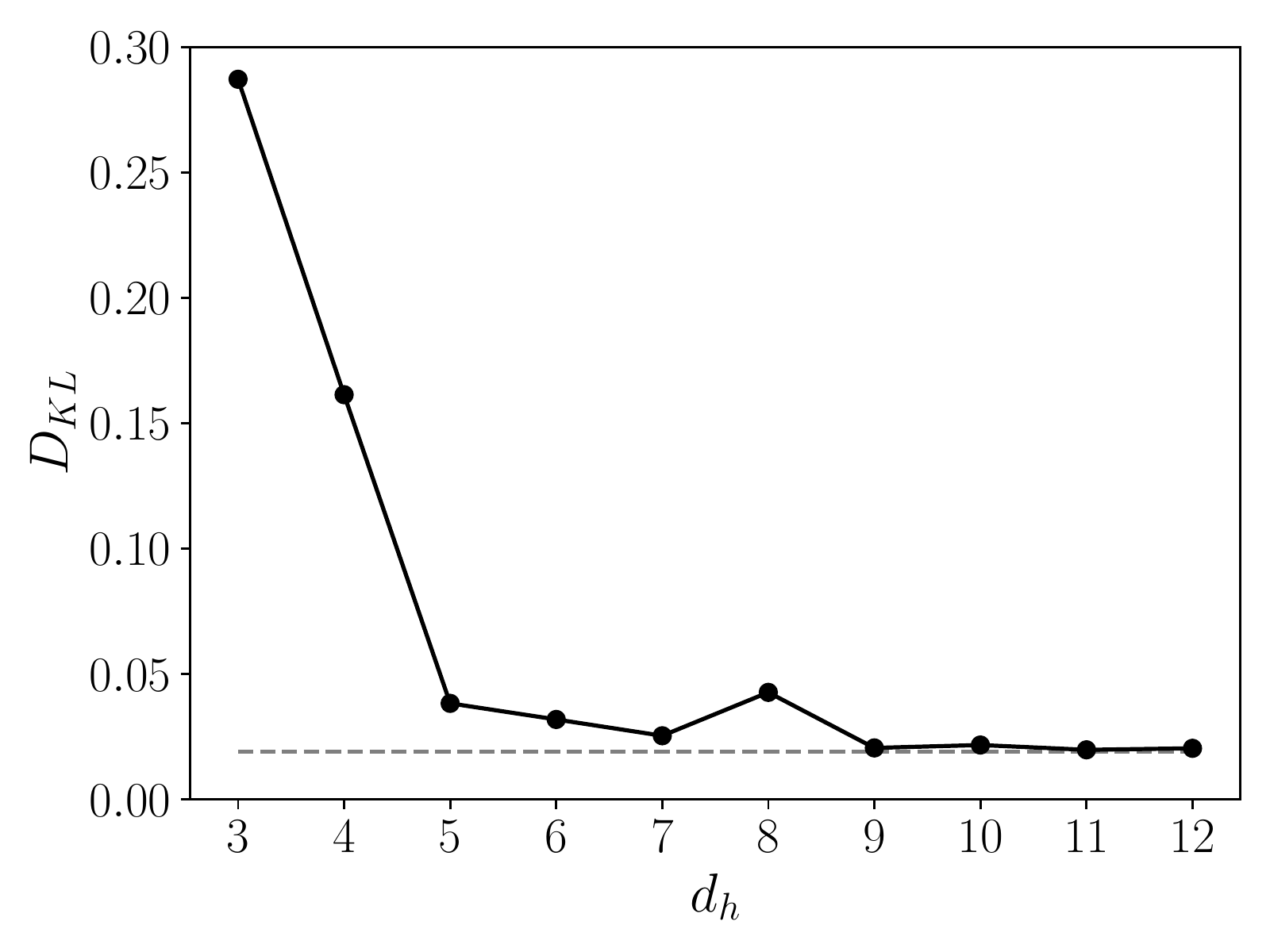}
		\caption{}
		\label{fig:subtb}
	\end{subfigure}
	
	\caption{$\operatorname{Re}=14.4$: $D_{KL}$ vs dimension $d_h$ corresponding to PDFs for (a) $t_q$ (b) $t_b$. Dashed grey line corresponds to $D_{KL}$ calculated over different true data sets.}
	\label{re14d4tqtb}
	
\end{figure}

The above PDFs yield no information about the temporal behavior of the system. One temporal feature of significant interest in problems with intermittency is the probability density of the durations of time intervals with different behavior.  To address this, we consider the PDFs of time spent in bursting ($t_b$) and in quiescent ($t_q$) regions. The labeling method  discussed in the previous section is used. For this calculation we take a trajectory of $10^5$ snapshots from an arbitrary IC. The PDF for the true data is shown in Figure \ref{fig:subhibburtrue} followed by the PDFs that come from the $d_h=3,5,7,9,$ and $11$ models in Figures \ref{fig:subhibburd3} - \ref{fig:subhibburd7}. The true data shows that $t_q$ is mostly concentrated between $t \approx 200-300$ with a high intensity peak shown at $t=5$. We attribute this peak to a small fraction of snapshots in the bursting region that get mislabeled as quiescent due to the weakly chaotic nature of the data. We do not expect for this to drastically change our conclusions because the same labeling system is used for the true data and the models. In the case of $t_b$ we notice that these are mostly concentrated between $t \approx 0-200$. Looking at both the PDFs and averages of the times we see that $d_h=3$ fails to correctly capture the shape of the PDF and also underpredicts $ \langle t_q \rangle$ and $ \langle t_b \rangle$. At $d_h=5$ we start getting better agreement where we see that the PDFs clearly show the two regions where $t_b$ and $t_q$ are concentrated. In the case of $d_h=7$  we can see that the quiescent PDF spreads into regions with higher $t_q$ and for $d_h=9,11$ these seem to agree better with the true PDF. Figure \ref{re14d4tqtb} shows $D_{KL}$ with varying $d_h$ for these PDFs. As expected from observing the PDFs we see that $D_{KL}$ decreases up until $d_h=5$ for both cases. In the case of $t_q$ we see an increase in the error after $d_h=5$ which agrees with the above observation of the PDF at $d_h=7$. For $t_b$, $D_{KL}$ seems to keep slightly decreasing after $d_h=5$. We also notice that for $t_q$, $D_{KL}$ reaches a minimum at $d_h=9$ and for $t_b$ no significant decrease is observed at $d_h \geq 9$. In short, these duration statistics achieve similar agreement at $d_h=9$, and for the case of $t_b$ errors keep decreasing with increasing $d_h$. We also calculate the mean of $t_q$ and $t_b$ for the case of $d_h=9$ and obtain values of $\langle t_q \rangle =174$ and $\langle t_b \rangle =97$ which agree closely with the true values of $\langle t_q \rangle =176$ and $\langle t_b \rangle =97$.

%In short, while the low-dimensional models do not achieve the same agreement with the true results for these duration statistics as we do for the static quantities considered above, they nevertheless capture the key features of the distributions and capture their means with reasonable accuracy, within about $20\%$. We also report the means  of these time durations as well as the standard error of the mean (SE) in Table \ref{tablehibburtime}.
 %which shows that at $d_h=5$ the predicted PDFs assimilate the True data as shown in Figure \ref{re14d4tqtb}. 
 
 %Noticing from the previous results that results do not depend on IC here we select 10 arbitrary ICs that are run to get $10^5$ snapshots

%\begin{figure}%[H]
%	%\vspace{-8mm}
%	%\centering
%	\begin{subfigure}{0.45\textwidth}
%		%\centering
%		\includegraphics[width=1\linewidth]{./tq_plot.pdf} % Figure in KLDiv other bins
%		\caption{}
%		\label{fig:tq_plot}
%	\end{subfigure}%
%	\begin{subfigure}{0.45\textwidth}
%		%\centering
%		\includegraphics[width=1\linewidth]{./tb_plot.pdf}
%		\caption{}
%		\label{fig:tb_plot}
%	\end{subfigure}
%	
%	\caption{\CEPrevise{Average and standard error of the mean of (a) $t_q$ and (b) $t_b$ for true (dashed line) and dimensions $d_h=3-12$. Vertical gray lines and blue shading corresponds to the standard errors of the models and true data respectively.\MDG{remove, just report values at $d_h=11$}}}
%	\label{re14d4_tq_tb_plot}
%	
%\end{figure}

%\begin{table}[h]
%	\caption{\CEPrevise{Average and standard error of the mean of $t_q$ and $t_b$ for true and dimensions $d_h=3-12$.}}
%	$$
%	\begin{array}{lcccc}\hline \hline & \text { $ \langle t_q \rangle$ } \; \; \; & \text { $\langle t_b \rangle$ } \; \; \; & \text { SE($ t_q$) } \; \; \; & \text { SE($ t_b$)  } \; \; \; \\ \hline \text { True } & 176& 97 & 3 & 2 \\ \text { $d_h=3$ } & 37 & 73 & 1 & 2 \\ \text { $d_h=4$ } & 174 & 85 & 5 & 3 \\ \text { $d_h=5$ } & 160 & 105 & 3 & 3\\ \text { $d_h=6$ } & 185 & 106 & 3 & 3\\ \text { $d_h=7$ } & 202 & 101 & 4 & 3 \\ \text { $d_h=8$ } & 165 & 97 & 3 & 2 \\ \text { $d_h=9$ } & 174 & 92 & 3 & 2 \\ \text { $d_h=10$ } & 172 & 97 & 3 & 2 \\ \text { $d_h=11$ } & 185 & 93 & 3 & 2 \\ \text { $d_h=12$ } & 182 & 90 & 3 & 2 \\ \hline \hline\end{array}
%	$$
%	\label{tablehibburtime}
%\end{table}

\subsection{Phase prediction}\label{phasepred}

%For the predicted phase we first calculate $\Delta \tilde{\phi}_x(t + \tau) = G(h(t))$

%To obtain $\Delta \tilde{\phi}_x(t + \tau)$ 

% Figure \ref{re14d4phasevis} shows a short time evolution of $\phi_x(t)$ and $\Delta \phi_x(t + \tau)$ \MDG{what's the difference?} corresponding to the true and predicted data for $d_h=5$ model. 

%\MDG{The following sentence is incorrect -- the regions of smooth increase or decrease are simply the RPO traveling in $x.$ It's moving between RPOs in the bursts, where the phase is fluctuating rapidly. Rewrite.} 

\begin{figure}%[H]
	%\vspace{-8mm}
	%\centering
	\begin{subfigure}{0.45\textwidth}
		%\centering
		\includegraphics[width=1\linewidth]{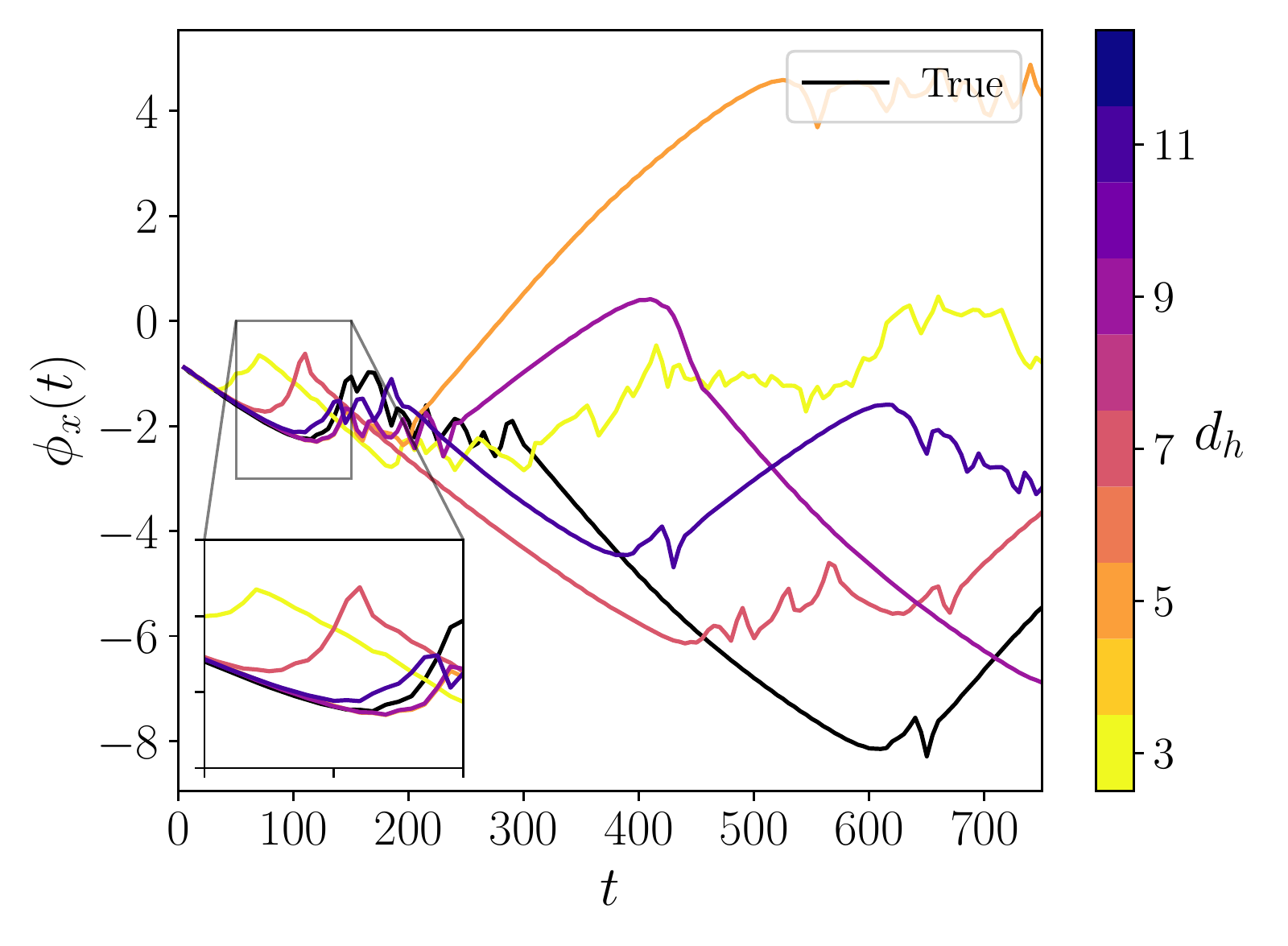}
		\caption{}
		\label{re14d4phasecomparev2}
	\end{subfigure}%
	\begin{subfigure}{0.45\textwidth}
		%\centering
		\includegraphics[width=1\linewidth]{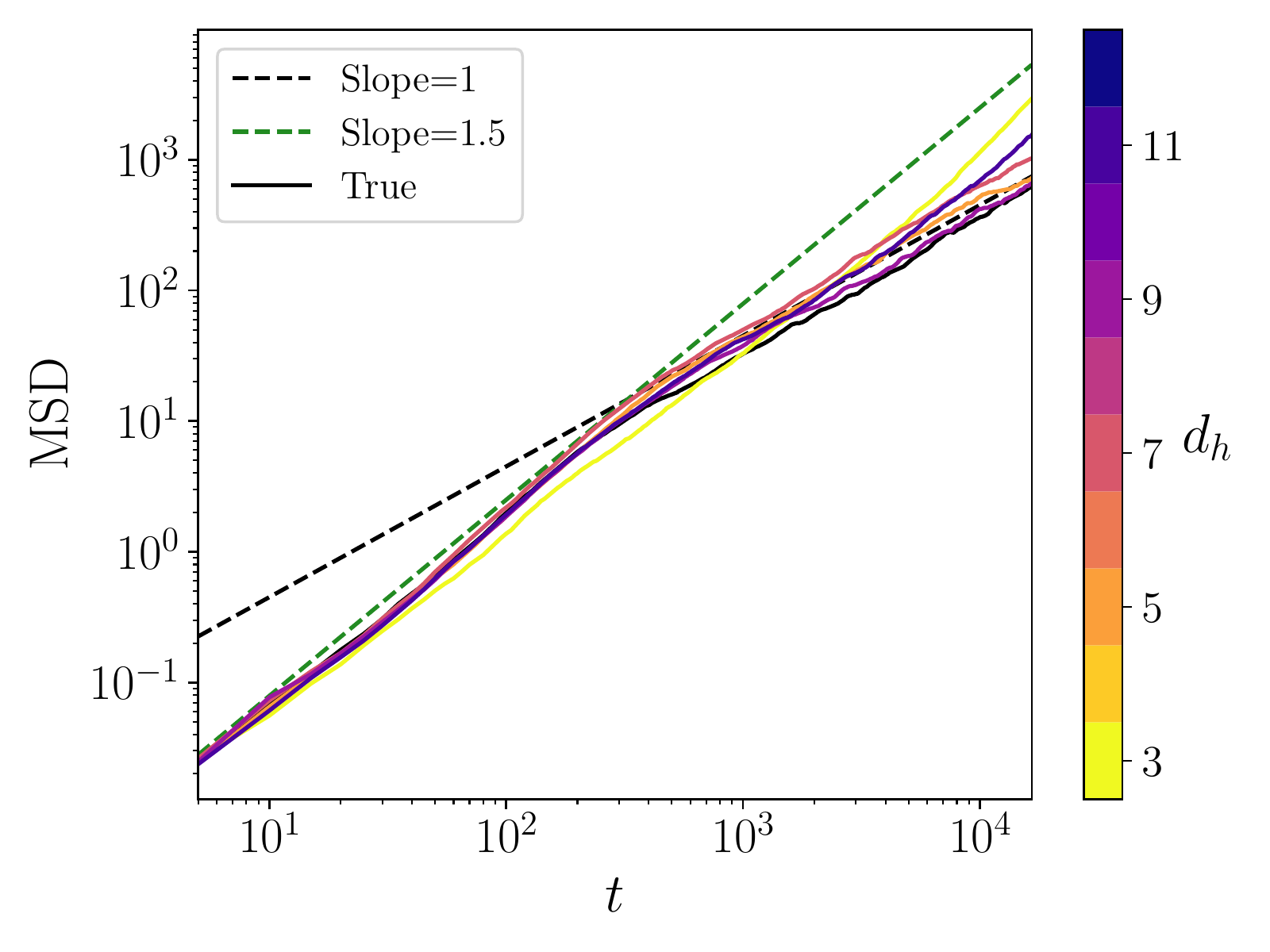}
		\caption{}
		\label{re14d4phaseMSDa}
	\end{subfigure}
	
	\caption{(a) Time evolution of $\phi_x$ corresponding to the true data and models with dimensions $d_h=3,5,7,9,$ and $11$. (b) MSD of $\phi_x (t)$ corresponding to true data and models with dimensions $d_h=3,5,7,9,$ and $11$.}
	\label{re14d4phaseMSD}
	
\end{figure}

Recall that we gain substantial accuracy in dimension reduction by factoring out the spatial phase $\phi_x(t)$ of the data. Here we complete the dynamical picture of the model predictions at $\operatorname{Re}=14.4$ by illustrating the predictions of phase evolution, as given by the learned phase evolution equation \eqref{eq:phaseevolution}. Figure \ref{re14d4phasecomparev2} shows a short time evolution of $\phi_x(t)$ corresponding to the true and predicted data for the $d_h=3,5,7,9,$ and $11$ models. The smooth increases and decreases in Figure \ref{re14d4phasecomparev2} correspond to trajectories during time intervals where they are near an RPO and thus are traveling in the $x$-direction. The intervals where the phase flucuates rapidly are the bursts during which the trajectories are moving between the RPO regions. This behavior is well-captured for all of the dimensions shown except for $d_h=3$. Notice that although the trajectories diverge, for short times we get around two $t_L$ of prediction horizon where the models still capture the correct dynamics, and Figure \ref{re14d4phasecomparev2} provides a clear visual indications that the loss of predictability occurs during the bursts.

We now take an approach to quantify how well the model performs with respect to the true data. Taking a look at the drops and increases for $\phi_x(t)$ we can observe that after every burst the trajectory will either travel, essentially randomly, in the positive (increasing $\phi_x$) or negative (decreasing $\phi_x$) $x$ direction. This behavior is essentially a run and tumble or random walk behavior in the sense that the long periods of positive or negative phase drift correspond to ``runs" that are separated by ``tumbles" that correspond to the bursts, in which the direction of phase motion is reset. Hence, a natural analysis of quantification for this type of dynamics consists of calculating the mean squared displacement (MSD) of the phase: 
\begin{equation}
\mbox{MSD} (t)= \langle (\phi_x (t) - \phi_x (0))^2 \rangle.
\end{equation}
Figure \ref{re14d4phaseMSDa} shows the time evolution of MSD of true and predicted data. The black line corresponds to the true data and the black and green dashed lines serve as references with slopes of 1 and 1.5, respectively. The colored lines correspond to models with various dimensions. Looking at the true curve we notice a change from superdiffusive (slope = 1.5) to diffusive (slope = 1) scaling that happens around $t \approx 200$, which corresponds to the mean duration of the quiescent intervals, as discussed above: i.e., to the average time the trajectories travel along the RPOs before bursting. The trajectory then bursts and reorients which is captured by the long time diffusive trend. Looking at the performance of the models we observe that $d_h=3$ does a good job at capturing the short time scaling, however it is not to able capture the change in slope that is observed in the true data. It is not until $d_h \geq 5$ that the correct behavior at long times is observed -- indeed the predictions agree very well with the data, with a slight upward shift at long times corresponding to the slight overprediction of the mean duration of the quiescent periods.

\subsection{Bursting prediction}\label{burpred}

%Visual inspection for these conditions show a clear threshold between quiescent and bursting events, which is not the case for $\operatorname{Re} = 14.4$, $n=2$. However with the labeling scheme previously discussed we attempt to predict bursting events with our models. 

%Towards this end we consider 

%Finally, the development of ROMs that provide us with the capability of reconstructing the data with high-fidelity and predicting in time motivates us to use the model to try to predict bursting events from data well in advance of the event. 

\begin{figure}%[H]
	%\vspace{-8mm}
	%\centering
	%\begin{subfigure}{0.7\textwidth}
	%\centering
	\includegraphics[width=0.6\linewidth]{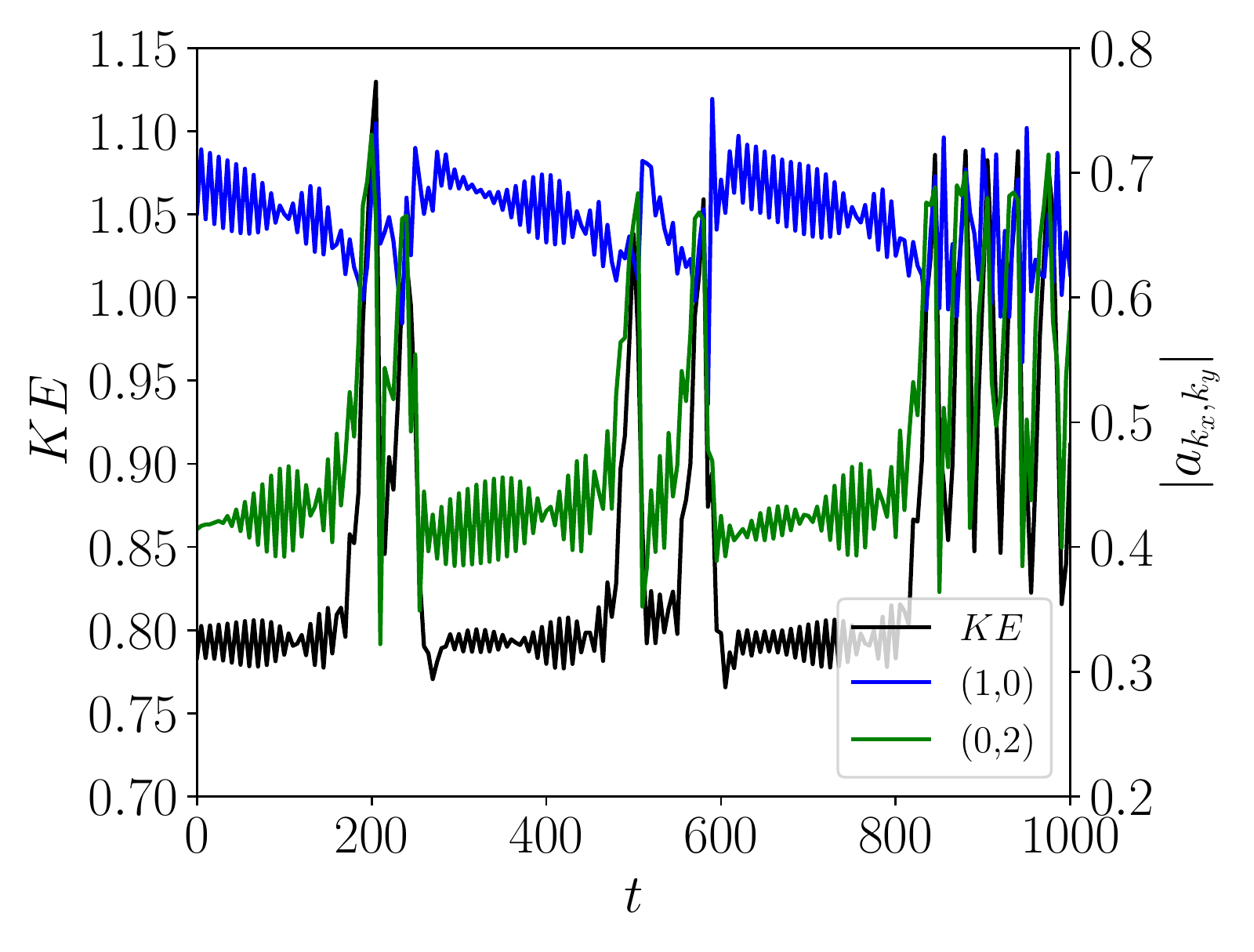}%fouriermodetrayectory.py
	%\caption{}
	%\label{KE0210}
	%\end{subfigure}%
	\begin{comment}
	\begin{subfigure}{0.5\textwidth}
	%\centering
	\includegraphics[width=.9\linewidth]{figures/Indicator_v2.pdf}%burst_different_indicators_dh5.py
	\caption{}
	\label{Indicators}
	\end{subfigure}
	\end{comment}
	
	\caption{Time evolution of $KE$ and amplitudes corresponding to $(1,0)$ and $(0,2)$ Fourier mode for $\operatorname{Re}=14.4$. }
	\label{indandKE}
	
\end{figure}

Previous research has focused on finding indicators that guide predictions of when a burst will occur. It has been shown for the Kolmogorov flow that before a burst there is a depletion of the content in the $(1,0)$ Fourier mode, which then feeds into the forcing mode $(0,n)$ \cite{nicolaenko1990symmetry}. Figure \ref{indandKE} shows how this looks for $\operatorname{Re} = 14.4$, $n=2$.  By considering a variational framework and finding solutions to a constrained optimization problem it was also found that examination of these modes can lead to predictions of when a burst will occur \cite{farazmand2017variational}. % A focus on the mode (1,0) has shown great promise for prediction of these extreme events with control targeted towards removing the excess energy of $(0,n)$ for $n=4$ \cite{farazmand2019closed}.

%\MDG{did any of these papers use a ROM? This is important in terms of comparison between their results and ours}
%\sout{With this finding, research has focused on predicting as well as controlling these extreme events for the case $\operatorname{Re} = 40$, $n=4$  \cite{farazmand2017variational, farazmand2019closed}. }

With our framework,  natural indicators are the latent variables $h$, which we will consider here along with some variations, including the indicators used in previous work.   To predict bursting events based on a given indicator, we will use a simple binary classifier in the form of a support vector machine (SVM) with a radial basis function kernel \cite{boser1992training}. These have shown success in predicting extreme events for problems such as extreme rainfall  \cite{nayak2013prediction}. With this approach, data at time $t$ is used to learn a function that outputs a binary label of bursting/not bursting at time $t+\tau_b$. For all of the cases considered we use the $d_h=5,9$ models, taking a dataset of $5 \times 10^4$ snapshots to train the SVM and another $5 \times 10^4$ as a test set.

\begin{figure}%[H]
	%\vspace{-8mm}
	%\centering
	\begin{subfigure}{0.45\textwidth}
		%\centering
		\includegraphics[width=1\linewidth]{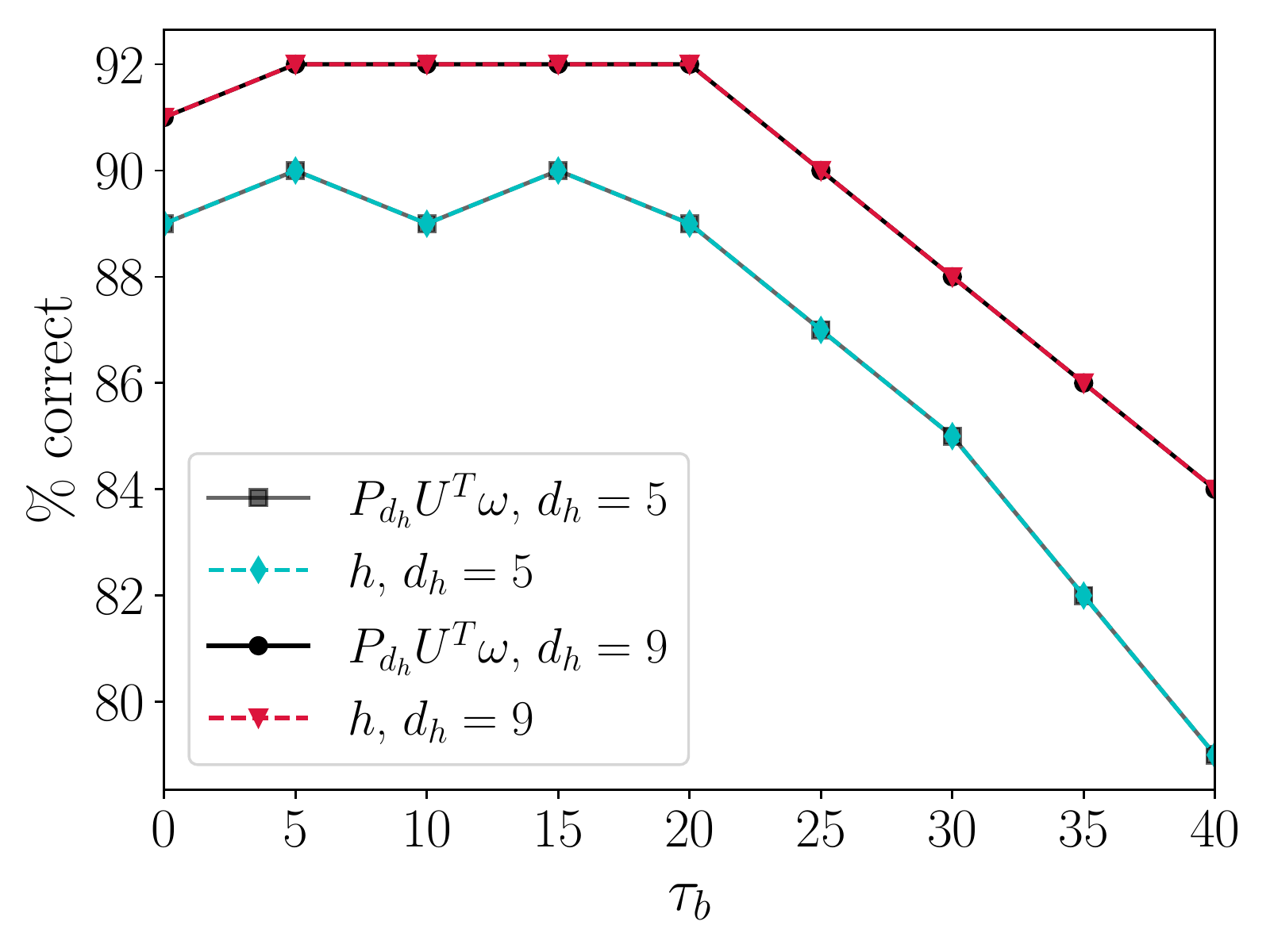} % These figures can be found in file BurstingPrediction/burst_different_indicators_dh5.py
		\caption{}
		\label{Pburs}
	\end{subfigure}%
	\begin{subfigure}{0.45\textwidth}
		%\centering
		\includegraphics[width=1\linewidth]{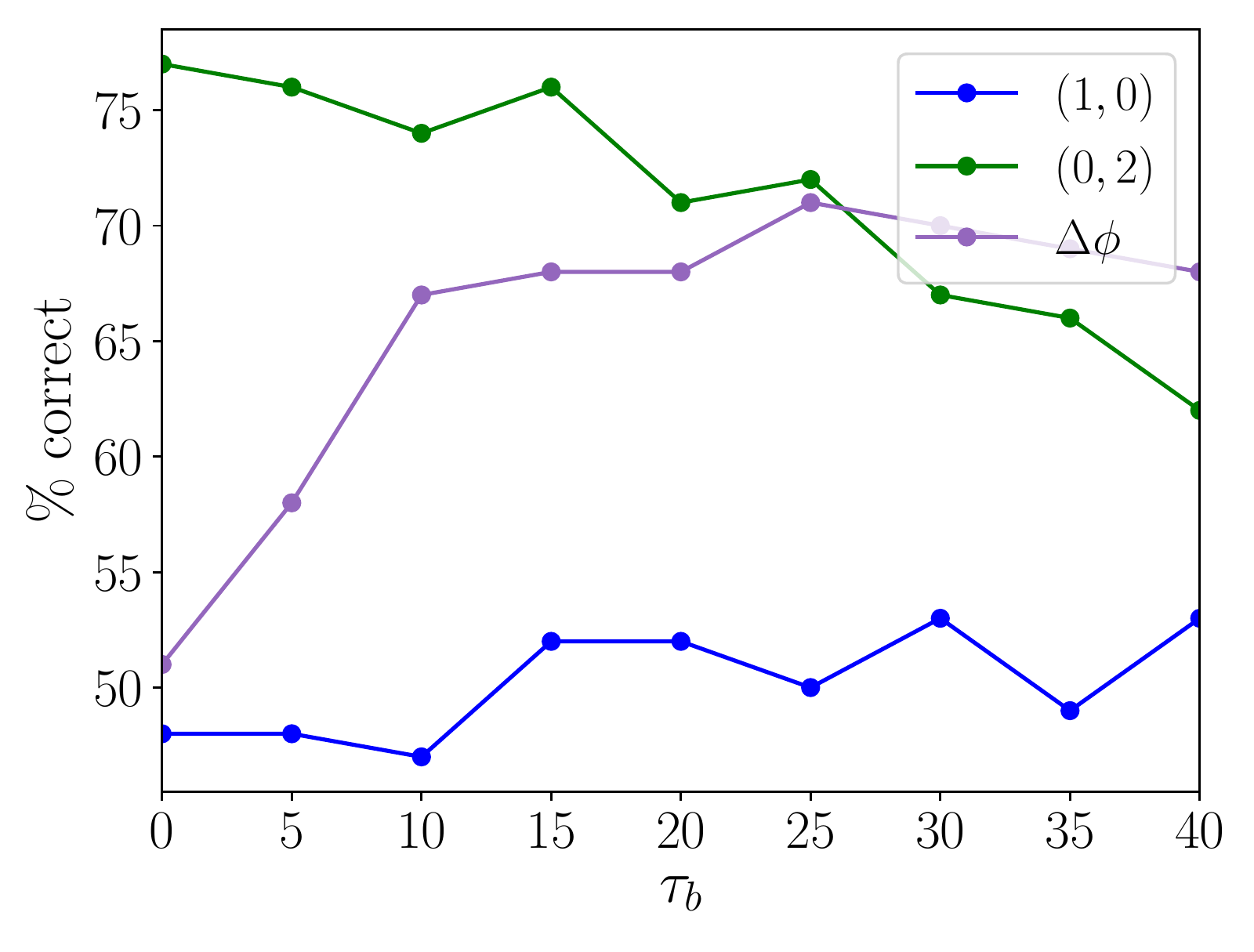}
		\caption{}
		\label{Pbursind}
	\end{subfigure}
	%\begin{subfigure}{0.45\textwidth}
		%\centering
	%	\includegraphics[width=1\linewidth]{figures/Pburs_unit2_v4.pdf}
	%	\caption{}
	%	\label{Pbursunit2}
	%\end{subfigure}
	%\begin{subfigure}{0.45\textwidth}
		%\centering
	%	\includegraphics[width=1\linewidth]{figures/Pburs_dh_v5.pdf}
	%	\caption{}
	%	\label{Pbursdh}
	%\end{subfigure}
	\caption{Percent of correctly classified bursting events at $\tau_b$ forward in time for: (a) $P_{d_{h}} U^{T} \omega$ and $h$ at $d_h=5,9$, (b) and indicators $\Delta \phi$, $(1,0)$, and $(0,2)$. Note that the vertical scales on (a) and (b) are very different.}
	\label{Pbursall}
	
\end{figure}

Figure \ref{Pburs} shows the percent correct classification of bursting events with varying time $\tau_b$ in the future. The black and gray curves corresponds to predicting the events based on the PCA projection of the data, $P_{d_{h}} U^{T} \omega$, into the first $d_h=5$ and $9$ coefficients respectively. The cyan and red curves corresponds to $h$ of dimensions $d_h=5$ and $d_h=9$ respectively.  We notice that the PCA and $h$ curves fall on top of another and have a high probability of correct classification when considering prediction horizons less than one $t_L$. For this purpose we see that PCA is enough to predict bursting events. Figure \ref{Pbursind} shows the percent correct classification of bursting at time $\tau_b$ in the future for the previous discussed indicators. None of these work nearly as well as $P_{d_{h}} U^{T} \omega$ or $h$. The blue curve corresponds to $(1,0)$ amplitude of the original true data, the green curve to the forcing $(0,2)$ amplitude, and we also consider $\Delta \phi$ in the purple curve. In the case of $\Delta \phi$ we see some predictability at times longer than one $t_L$ and less than two.  This also happens for the case of $(1,0)$, however there seems to be no decrease or increase in the probability of correct classification. We can see from Figure \ref{indandKE} that even though there is a depletion in the $(1,0)$ mode preceding bursts, its amplitude does not change dramatically between quiescent and bursting intervals, which may be a reason that it does not provide much predictive power. The amplitude $(0,2)$, which changes more strongly between quiescent and bursting regions, is seen to be the better predictor for bursting events. At small $\tau_b$ its predictions outperform $(1,0)$ and $\Delta \phi$, however at times larger than one $t_L$, $\Delta \phi$ performs better. 

%It is important to emphasize that many of the complications for bursting prediction might be related to the weakly chaotic nature of the $\operatorname{Re}$ considered. 

\begin{comment}
\begin{figure}%[H]
	%\vspace{-8mm}
	\centering
	\includegraphics[width=.5\linewidth]{figures/KE_02_10_v2.pdf}
	\caption{Time evolution of $KE$ and amplitudes corresponding to $(1,0)$ and $(0,2)$ Fourier mode for $\operatorname{Re}=14.4$}
	\label{KE0210}
\end{figure}
\end{comment}

\begin{comment}
\begin{figure}%[H]
	%\vspace{-8mm}
	\centering
	\includegraphics[width=.5\linewidth]{figures/Pburs_v2.pdf}
	\caption{Probability of correctly predicting a burst at $\tau_b$ forward in time}
	\label{Pburs}
\end{figure}
\end{comment}

%take the approach of calculating joint PDF statistics for $\operatorname{Re} = 20$ case. The same 

\section{Conclusion} \label{sec:Conclusion}

%Hence nonlinearity, as oposed to linear, is important when learning a low dimensional coordinate and time map.

The nonlinearity of the NSE poses challenges when using ROMs, where the dynamics are expected to evolve on an invariant manifold that will not lie in a linear subspace. Neural networks have proven to be powerful tools for learning efficient ROMs solely from data, however finding and exploiting a \emph{minimal}-dimensional model has not been emphasized. We present a data-driven methodology to learn an estimate of the embedding dimension of the manifold for chaotic Kolmogorov flow and the time evolution on it. An autoencoder is used to find a nonlinear low-dimensional subspace and a dense neural network to evolve it in time.

Our autoencoders are trained on vorticity data from two cases: a case where the dynamics show a relative periodic orbit solution ($\operatorname{Re}=13.5$), and a case with chaotic dynamics ($\operatorname{Re}=14.4$). The chaotic regime we consider comes with challenges due to the intermittent behavior observed where the trajectory travels in between quiescent intervals and bursting events. We factor out the rich symmetries of Kolmogorov flow before training of the autoencoders, which dramatically improves reconstruction error of the snapshots. This improves training efficiency by not having to learn a compression of the full state. Specifically, factoring out the translation symmetry decreases the mean-squared reconstruction error by an order of magnitude compared to the case where phase is not factored out, and several orders of magnitude compared to PCA. The phase-aligned low-dimensional subspace is then used for time evolution where the RPO dynamics is learned essentially perfectly at $d_h=2$ for $\operatorname{Re}=13.5$ and very good agreement for short and long time statistics is obtained at $d_h=5$ for $\operatorname{Re}=14.4$. Further small improvements in the results occur as dimension is increased to nine, beyond which the statistics of the model and true system are in very good agreement. For comparison, the full state space of the numerical simulation data is $N=1024$.

We also show phase prediction evolution results based on the low-dimensional subspace learned. The time evolution of the true phase exhibits a superdiffusive scaling at short times and a diffusive scaling at long times which we attribute to the traveling near an RPO and the reorientation due to bursting. Finally, using the low-dimensional representation enables accurate prediction of bursting events based on conditions about a Lyapunov time ahead of the event. This work opens new avenues for data-driven ROMs with applications such as control for drag reduction, an example of which is presented for turbulent Couette flow in \cite{linot2023turbulence}. One important challenge that remains is more effective treatment of systems with intermittent dynamics like those described here. A recent study \cite{floryan2021charts} has introduced a method that uses the differential topology formalism of charts and atlases to develop \emph{local} manifold representations and dynamical model that can be stitched together to form a global dynamical model. One attractive feature of that formalism is that it enables use of separate representations for regions of state space with very different dynamics, and has already shown in specific cases to provide dramatically improved results for dynamics with intermittency.   

\begin{acknowledgments}
This work was supported by AFOSR  FA9550-18-1-0174 and ONR N00014-18-1-2865 (Vannevar Bush Faculty Fellowship). We also want to thank the Graduate Engineering Research Scholars (GERS) program and funding through the Advanced Opportunity Fellowship (AOF) as well as the PPG Fellowship.
\end{acknowledgments}

\bibliography{library.bib}

\end{document}